\documentclass[10pt,twocolumn,letterpaper]{article}

\usepackage{cvpr}              
\definecolor{cvprblue}{rgb}{0.21,0.49,0.74}
\usepackage[pagebackref,breaklinks,colorlinks,allcolors=cvprblue]{hyperref}

\usepackage[table]{xcolor}
\usepackage{amsmath}
\usepackage{amsfonts}
\usepackage{bm}
\usepackage{graphicx}
\usepackage{subcaption}
\usepackage{booktabs}
\usepackage{multirow}
\usepackage{wrapfig}
\usepackage{float}
\usepackage{nicefrac}
\usepackage{microtype}
\usepackage{tikz}
\usepackage{url}
\usepackage{hyperref}
\usepackage[accsupp]{axessibility}  

\def\paperID{8356} 
\def\confName{CVPR}
\def\confYear{2026}

\title{Beyond the Ground Truth: Enhanced Supervision for Image Restoration}

\author{Donghun Ryou$^{*2}$ \qquad   Inju Ha$^{*1}$ \qquad Sanghyeok Chu$^{1}$ \qquad Bohyung Han$^{1,2}$ \\
Computer Vision Laboratory, $^1$ECE \& $^2$IPAI, Seoul National University\\
{\tt\small \{dhryou, hij1112, sanghyeok.chu, bhhan\}@snu.ac.kr}\\
\href{https://hij1112.github.io/beyond-the-ground-truth/}{\tt\small https://hij1112.github.io/beyond-the-ground-truth/}
}

\begin{document}
\maketitle

\begingroup
\renewcommand\thefootnote{} 
\footnotetext{\protect\hspace{-0.5em}${}^*$indicates equal contribution.}
\endgroup

\begin{abstract}
Deep learning-based image restoration has achieved significant success. 
However, when addressing real-world degradations, model performance is limited by the quality of ground-truth images in datasets due to practical constraints in data acquisition.
To address this limitation, we propose a novel framework that enhances existing ground truth images to provide higher-quality supervision for real-world restoration.
Our framework generates perceptually enhanced ground truth images using super-resolution by incorporating adaptive frequency masks, which are learned by a conditional frequency mask generator.
These masks guide the optimal fusion of frequency components from the original ground truth and its super-resolved variants, yielding enhanced ground truth images.
This frequency-domain mixup preserves the semantic consistency of the original content while selectively enriching perceptual details, preventing hallucinated artifacts that could compromise fidelity.
The enhanced ground truth images are used to train a lightweight output refinement network that can be seamlessly integrated with existing restoration models. 
Extensive experiments demonstrate that our approach improves the quality of restored images.
We further validate the effectiveness of both supervision enhancement and output refinement through user studies.
\end{abstract}

\vspace{-2mm}
\section{Introduction}

\begin{figure}[t]
    \centering
    \scalebox{0.9}{
    \begin{minipage}{\linewidth}
    \begin{subfigure}[t]{\linewidth}
        \centering
        \begin{tikzpicture}
            \node[anchor=south west,inner sep=0] (image) at (0,0) {
                \scalebox{1}{
                \includegraphics[width=\linewidth]{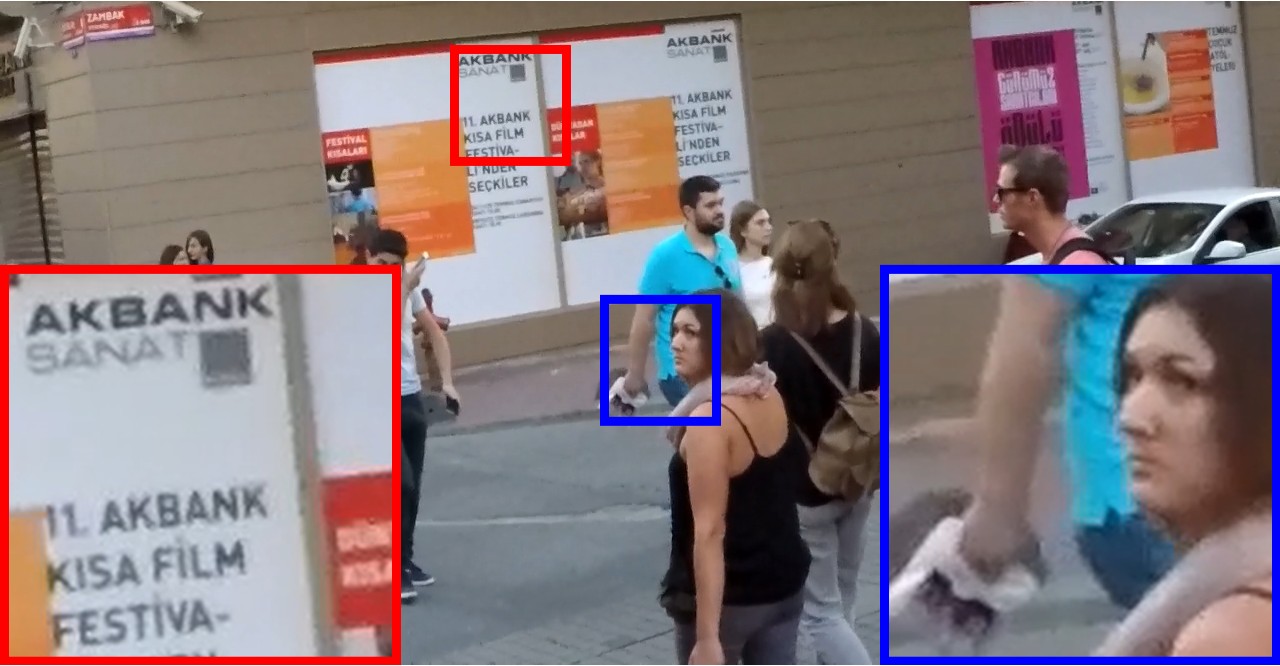}
                }
            };
            \node[anchor=north west, fill=black, fill opacity=0.4, text=white, text opacity=1, inner sep=2pt, xshift=3pt, yshift=-0.5pt] at (image.north west) {\small Original GT};
            \node[anchor=north east, fill=black, fill opacity=0.4, text=white, text opacity=1, inner sep=2pt, xshift=-3pt, yshift=-0.5pt, align=right] at (image.north east) {\small MUSIQ: 47.82 \\ \small VisualQuality-R1: 3.50};
        \end{tikzpicture}
        \vspace{-3mm}
    \end{subfigure}
    \begin{subfigure}[t]{\linewidth}
        \centering
        \begin{tikzpicture}
            \node[anchor=south west,inner sep=0] (image) at (0,0) {
                \scalebox{1}{
                \includegraphics[width=\linewidth]{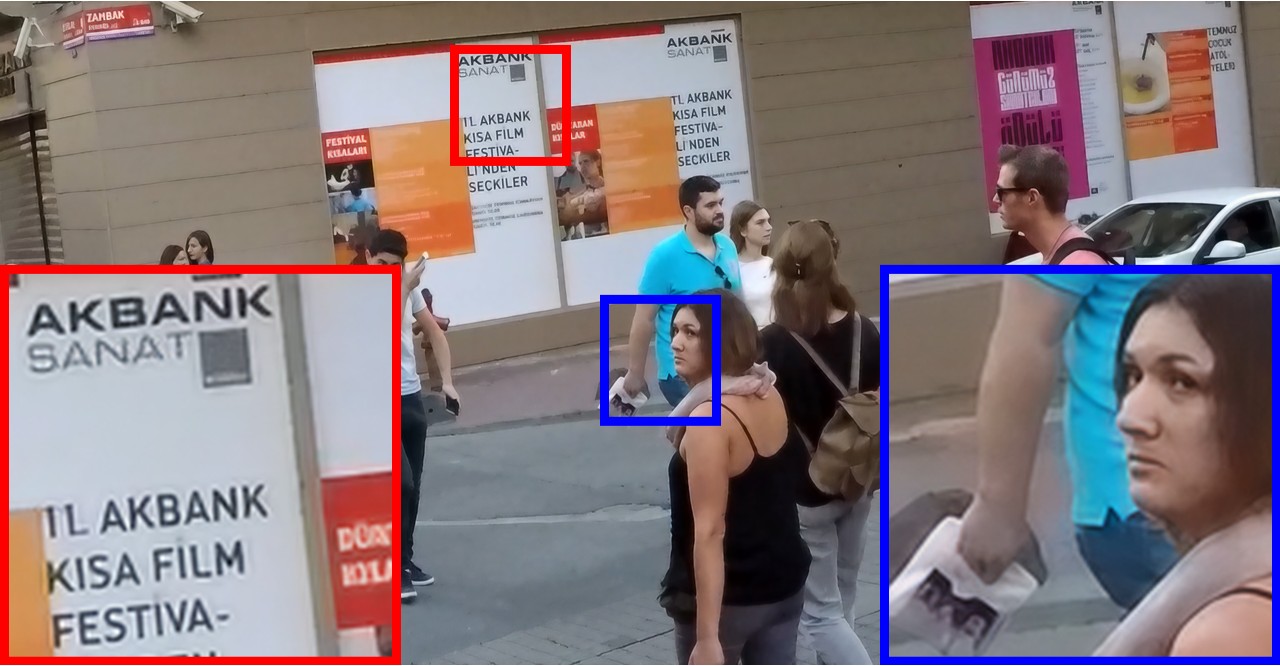}
                }
            };
            \node[anchor=north west, fill=black, fill opacity=0.4, text=white, text opacity=1, inner sep=2pt, xshift=3pt, yshift=-0.5pt] at (image.north west) {\small Enhanced GT};
            \node[anchor=north east, fill=black, fill opacity=0.4, text=white, text opacity=1, inner sep=2pt, xshift=-3pt, yshift=-0.5pt, align=right] at (image.north east) {\small MUSIQ: 71.56 \\ \small VisualQuality-R1: 4.80};
        \end{tikzpicture}
    \end{subfigure}
    \vspace{-6mm}
    \end{minipage}
    }
    \caption{
        Visualization of our enhanced ground truth (GT).
        Our enhanced GT not only demonstrates sharper text and superior perceptual quality but also maintains semantic consistency with respect to the original GT.
        Zoom in for better visualization.
    }
    \label{fig:enhnaced_gt}
    \vspace{-5mm}
\end{figure}
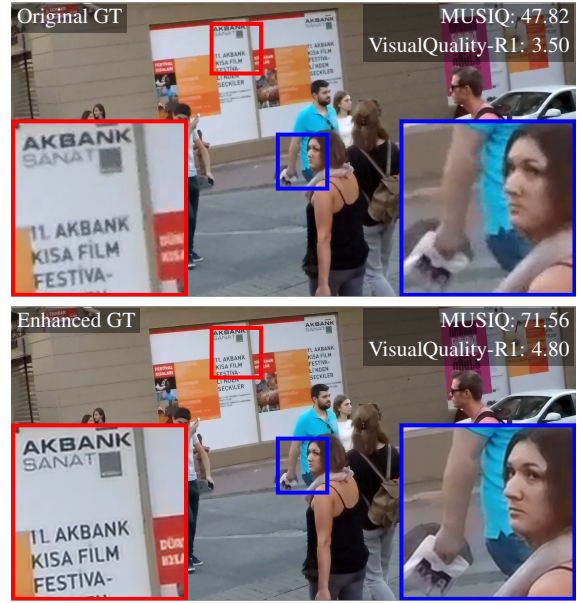
\label{Introduction}
Image restoration has achieved remarkable progress through supervised training on paired low-quality and ground truth images using deep neural networks.
Across various degradation types, a range of architectures~\citep{zhang2017beyond, kupyn2018deblurgan, liang2021swinir, kim2022fine, chen2022simple, zamir2022restormer, guo2024mambair, Sun_2025_CVPR, Chen_2025_CVPR} and learning strategies~\citep{lehtinen2018noise2noise, ulyanov2018deep, yoo2020rethinking, zhang2022self, wu2024id, ryou2024robust, ha2024learning} have been proposed to align restored outputs closely with ground truth images.
Recently, the focus has shifted toward improving perceptual quality of the restored outputs, leveraging advances in generative models to produce visually compelling results~\citep{wang2024exploiting, lin2024diffbir, yu2024scaling, wu2024seesr}.

Despite these advances, improving perceptual quality remains a major challenge in real-world image restoration, where obtaining ideal reference images is inherently difficult due to practical data acquisition constraints.
Many existing datasets rely on indirect ways to construct ground truth images. 
For instance, in deblurring datasets~\citep{nah2017deep, shen2019human, nah2019ntire}, ground truth images are selected from video sequences, which often contain slight camera shake or object movements, limiting the image sharpness.
In denoising datasets~\citep{nam2016holistic, abdelhamed2018high, xu2018real}, ground truth images are constructed by averaging multiple noisy captures, often resulting in blurred references.
As a result, models trained on such suboptimal ground truth images inevitably tend to inherit those imperfections, limiting their ability to achieve high-quality restoration.

To address this limitation, we propose a novel supervision enhancement framework designed to improve the perceptual quality of suboptimal ground truth images. 
The proposed framework consists of two main components: (1) super-resolution using a one-step diffusion model to generate perceptually enhanced ground truth variants, and (2) frequency-domain mixup to produce the final enhanced ground truth images. 
For the frequency-domain mixup, we introduce a conditional frequency mask generator that adaptively produces masks to guide the optimal fusion of frequency components from the original ground truth image and its super-resolved variants.
As illustrated in Figure~\ref{fig:enhnaced_gt}, the resulting enhanced ground truth image provides clearer details and higher perceptual quality than the original.

Building upon the enhanced ground truth images, we design a lightweight output refinement network that can be seamlessly integrated into a wide range of pretrained restoration models without requiring architectural changes or retraining.
Experiments show that the refinement network consistently improves the quality of restored images, benefiting from the enhanced supervision provided by our framework.
Moreover, the network exhibits strong robustness in out-of-distribution scenarios, effectively removing residual degradations that remain after initial restoration.
User studies further confirm the superior quality of both the enhanced ground truth images and the refinement outputs.

In summary, our key contributions are as follows:
\begin{itemize}
\item We identify the limitations of conventional ground truth images as a critical bottleneck in real-world image restoration, and propose a supervision enhancement framework based on frequency-domain mixup of an original ground truth image and its super-resolved variants. This design preserves semantic fidelity while enriching perceptual details, resulting in more reliable supervisory signals.

\item We introduce a lightweight refinement network that is trained solely on the original and enhanced ground truth images, requiring no additional annotations. The module is model-agnostic, seamlessly integrating with arbitrary restoration backbones without architectural modifications or retraining, and is empirically shown to be robust even under out-of-distribution degradations.

\item We validate our approach through extensive experiments and user studies, demonstrating consistent improvements in both enhanced ground truth quality and restored image fidelity. 
\end{itemize}

\section{Related Works}
\label{sec:relatedworks}


With the rise of deep learning, traditional image restoration methods have largely been replaced by data-driven approaches trained on paired low-quality  and ground truth images.
A wide range of architectures has been proposed, including Convolutional Neural Networks~\citep{dong2014learning, zhang2017beyond, chen2022simple}, Transformer-based models~\citep{zamir2020learning, liang2021swinir, zamir2022restormer}, Generative Adversarial Networks~\citep{ledig2017photo, kupyn2018deblurgan}, and more recently, state-space models such as Mamba~\citep{guo2024mambair, guo2024mambairv2}.
These models have typically been trained to maximize metrics such as PSNR and SSIM, which quantify the pixel-wise similarity to the original ground truth images.
While these approaches achieve high performance on standard benchmarks, their outputs often lack perceptual realism of high-quality images.

Recently, there is a growing interest in enhancing the perceptual quality of restored images.
This has been particularly prominent in image super-resolution, demonstrating significant advancements in generating visually plausible high-frequency details~\citep{xia2023diffir, delbracio2023inversion, wang2024exploiting, lin2024diffbir, yu2024scaling, wu2024seesr, wu2024one, kang2025icm}.
This trend extends to broader restoration tasks, such as deblurring and denoising, where diffusion models have been leveraged to enhance perceptual quality~\citep{ohayon2021high, kawar2022denoising, luo2023image, zhu2023denoising, yue2024efficient, liu2025one}.
While these methods effectively enhance perceptual quality, they often incur significant inference overhead and, more critically, risk hallucinating details or textures absent in the ground truth.
In contrast, our approach aims to enhance perceptual quality while preserving the semantics of the original content by our novel frequency mixup strategy.
To assess the perceptual quality of restored images, we employ a combination of deep-learning based image quality assessments~\citep{ke2021musiq, yang2022maniqa, chen2024topiq, zhang2023blind, chen2025toward}, and emerging Vision-Language Model-based methods~\citep{wu2025visualquality,li2025q}.

\section{Supervision Enhancement Framework}
\begin{figure*}[t]
    \centering
    \scalebox{0.93}{ 
        \includegraphics[width=\linewidth]{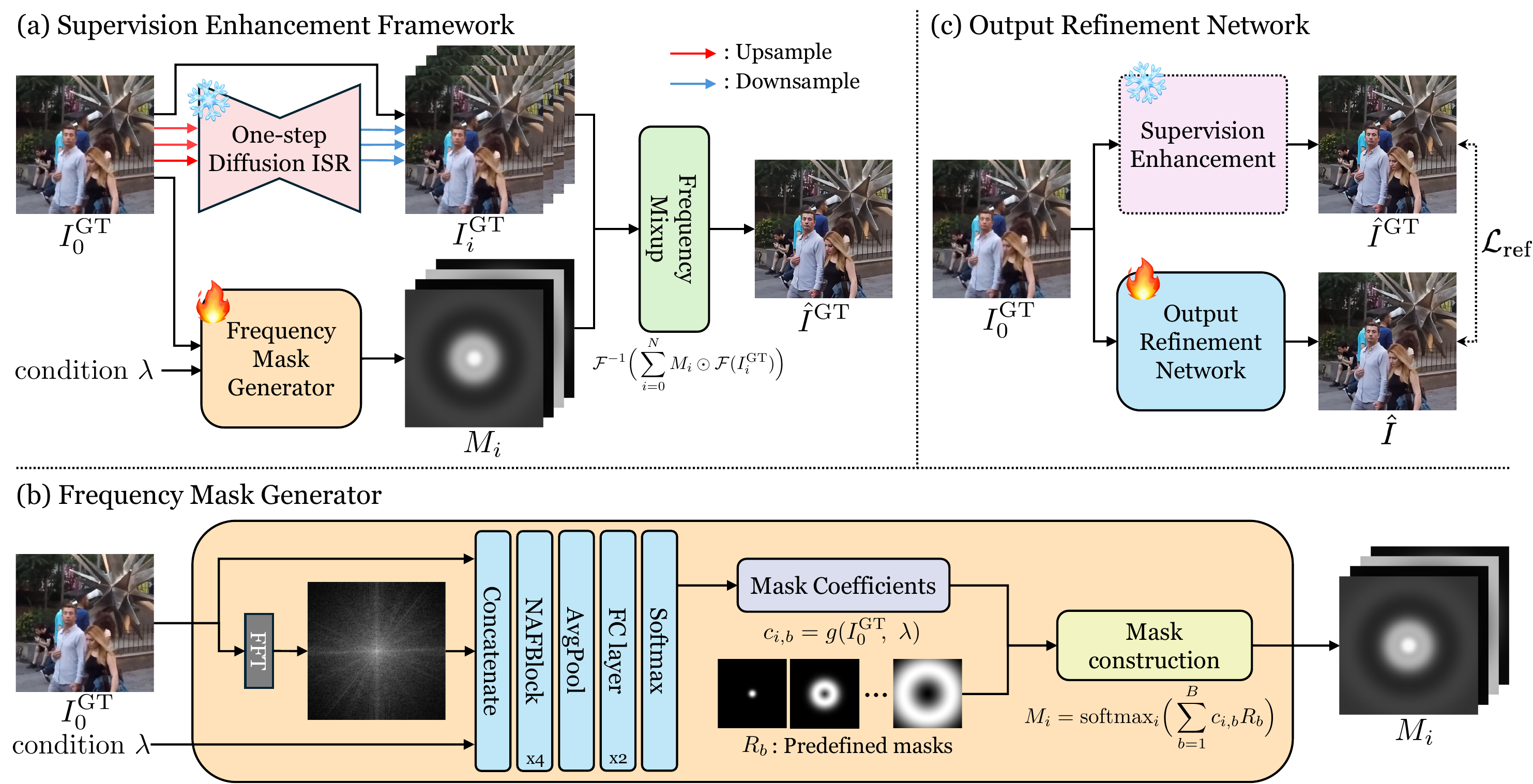}
    }
    \vspace{-1mm}
    \caption{
    Overview of our framework. 
    (a) The supervision enhancement framework produces enhanced ground truth images by fusing frequency components from the original ground truth $I_0^{\text{GT}}$ and its super-resolved variants $I^{\text{GT}}_i$ using adaptive frequency masks $M_i$.
    (b) The conditional frequency mask generator constructs $M_i$ by combining predefined masks $R_b$ weighted with predicted coefficients $c_{i,b}$, followed by a softmax function.
    (c) The output refinement network (ORNet) is trained with this enhanced supervision to improve perceptual quality.
    }    
    \label{fig:overview}
    \vspace{-5mm}
\end{figure*}

In this section, we introduce our supervision enhancement framework, which improves the perceptual quality of ground truth images in existing datasets to provide better supervision for image restoration tasks.
The framework consists of two main components: (1) super-resolution using a one-step diffusion model to generate perceptually enhanced ground truth variants, and (2) combining these variants with the original ground truth image through frequency-domain mixup using masks generated by a conditional frequency mask generator.
Figure~\ref{fig:overview} (a) illustrates an overview of our framework.

\subsection{Image Enhancement with Super-Resolution}

Recent Image super-resolution (ISR) models have shown remarkable capability in improving perceptual quality.
These models are trained using a combination of reconstruction and regularization losses, where the regularization term is crucial in learning natural image distributions and improving output quality.
Typically, ISR models are trained to align the distribution of generated samples $q(\hat{x})$ with the distribution of high-quality real images $p(x_H)$, by minimizing the Kullback-Leibler divergence:
\begin{equation}
  \mathcal{D}_\text{KL}(q(\hat{x})||p(x_H)).
\end{equation}
%
Typically, the distribution $p(x_H)$ is acquired from datasets with genuinely high-quality images, such as DIV2K~\citep{agustsson2017ntire} or LSDIR~\citep{li2023lsdir}, or by leveraging the high-quality image manifold of large-scale pre-trained diffusion models.
As a result, ISR models trained on this regularization effectively generate super-resolved outputs {$\hat{x}$} of high perceptual quality.

Leveraging this capability, we adopt a one-step diffusion ISR model~\citep{wu2024one} to enhance the suboptimal ground truth images. 
Specifically, each original ground truth image $I_0^{\text{GT}}$ is first upsampled using bicubic interpolation with $N$ multiple scale factors.
The diffusion-based ISR model is then applied to these upsampled images, and the outputs are downsampled back to the original resolution, yielding a set of perceptually improved ground truth variants $\{I_i^{\text{GT}}\}_{i=1}^{N}$.

\subsection{Frequency-Domain Mixup}

Although image super-resolution (ISR) can enhance perceptual quality, the generative nature of ISR models often introduces undesirable distortions in both semantic structures and photometric attributes.
To address this, we construct enhanced ground-truth images by integrating the original ground truth with multiple super-resolved variants.

A pixel-wise fusion in the spatial domain is problematic, since it amounts to selecting or averaging pixel intensities across images, making it difficult to preserve high-level semantic structures and frequently introducing unrealistic artifacts.
In contrast, we propose an adaptive frequency mixup, which provides fine-grained control by preserving essential low-frequency components in the original image while selectively incorporating perceptually richer high-frequency details from the super-resolved variants. 
This frequency-domain formulation is particularly suitable for image restoration tasks as it naturally harmonizes images with differing photometric characteristics, yielding more stable and visually coherent results than spatial-domain alternatives.


To facilitate optimal frequency fusion, we introduce a Conditional Frequency Mask Generator. 
As illustrated in Figure~\ref{fig:overview}(b), given a set of input images $\{I_i^{\text{GT}}\}_{i=0}^{N}$, where $i=0$ denotes the original ground truth and $i=1,\dots,N$ denote its super-resolved variants, the mask generator outputs frequency masks $M_i$ by combining a set of predefined ring-shaped Gaussian basis masks $\{R_b\}_{b=1}^{B}$, and predicted coefficients for each basis.

The design of our ring-shaped Gaussian basis masks is crucial for two reasons. 
First, the ring-shape enables precise control from low to high frequencies in a band-wise manner.
Second, the Gaussian shape ensures smooth transitions between frequencies, unlike discrete masks that introduce sharp boundaries, causing training instability and visual artifacts.

Specifically, each basis mask $R_b\in\mathbb{R}^{H\times W}$ is defined as:
\begin{equation}
    (R_b)_{h,w} = \exp\left(-(d(h,w)-\mu_b)^2 / 2\sigma_b^2 \right),
\end{equation}
for $1 \leq h \leq H$ and $1 \leq w \leq W$, where $d(h,w)$ denotes the $\ell_2$-distance from the center (DC component), and $\mu_b, \sigma_b$ represent the Gaussian parameters of the $b$-th mask.
The visualization of these basis masks is in the Appendix.

Given the original ground truth image $I_0^{\text{GT}}$ and the conditional parameter $\lambda$ that adjusts the weight between the original and its variants, a mask coefficient prediction network $g$ predicts coefficients $c_{i,b}\in\mathbb{R}$ as follows:
\begin{equation}
    c_{i,b}=g(I_0^{\text{GT}}\hspace{-1mm},~\lambda).
\end{equation}
Internally, $g$ augments the RGB input with its FFT representation, enabling joint use of spatial and frequency-domain information for mask coefficient prediction.

Then, the adaptive frequency masks $M_i$ are computed by combining these bases using predicted coefficients:
\begin{equation}
    M_i=\text{softmax}_i\Big(\sum_{b=1}^{B}c_{i,b}R_b\Big),
\end{equation}
where the softmax operation ensures masks sum to one, $\sum_{i=0}^{N} M_i(h,w)=1, \forall (h,w)$.

Finally, the enhanced ground truth image $\hat{I}^{\text{GT}}$ is constructed by fusing frequency components of the original ground truth $I_{0}^{\text{GT}}$ and its super-resolved variants $\{I_i^{\text{GT}}\}_{i=1}^{N}$ through frequency-domain mixup:
\begin{equation}
    \hat{I}^{\text{GT}} = \mathcal{F}^{-1}\Big(\sum_{i=0}^{N} M_i \odot \mathcal{F}(I_i^{\text{GT}})\Big),
\end{equation}
where $\mathcal{F}$ and $\mathcal{F}^{-1}$ denote Fourier and inverse Fourier transforms, and $\odot$ represents element-wise multiplication.

\subsection{Optimization}
To predict the mask coefficients for optimal frequency fusion, we train the network $g$, which predicts coefficients for each basis, with a composite loss that balances semantic integrity and perceptual quality.

The reconstruction loss is a $\ell_2$-loss that enforces consistency with the original ground truth $I_0^{\text{GT}}$:
\begin{equation}
    \mathcal{L}_{\text{recon}} = \|\hat{I}^{\text{GT}} - I_0^{\text{GT}}\|_2^2.
\end{equation}

    The perceptual loss is defined by a combination of multiple no-reference IQA metrics that evaluates perceptual quality of images (\eg MUSIQ~\citep{ke2021musiq}, 
MANIQA~\citep{yang2022maniqa}, TOPIQ~\citep{chen2024topiq}), denoted as $\text{IQA}_k(\cdot)$:
\begin{equation}
    \mathcal{L}_{\text{percep}} = -\sum_{k} \text{IQA}_k(\hat{I}^{\text{GT}}).
\end{equation}

The final training loss combines these two terms, with their relative weights controlled by $\lambda \in [0,1]$:
\begin{equation}
    \mathcal{L}=(1-\lambda) \mathcal{L}_{{\text{recon}}} +\lambda \mathcal{L}_{\text{percep}}.
\end{equation}

\section{Output Refinement Network}

We demonstrate the effectiveness of our enhanced ground truth images by training a lightweight Output Refinement Network (ORNet) to improve the outputs of existing restoration models.
While training a full restoration network from scratch using low-quality inputs and enhanced ground truth images is possible, we observe that many state-of-the-art models are already well-optimized for original ground truth images.
Therefore, we propose an efficient strategy that builds on top of a fixed, pre-trained restoration model $R_\phi$.
Specifically, we introduce a modular output refinement network $R_\theta$, which is trained to refine the output of $R_\phi$.
The overall process is formulated as:
\begin{equation}
   \hat{I} = R_\theta(R_\phi(I^\text{LQ}), \lambda),
\end{equation}
where $I^\text{LQ}$ is the low-quality input image, $\lambda$ is a parameter to control the level of perceptual enhancement, and $\hat{I}$ is the final restoration output.

Since the pre-trained restoration model $R_\phi$ produces outputs close to the original ground truth (i.e., $R_\phi(I^{\text{LQ}}) \approx {I_0^{\text{GT}}}$), we train $R_\theta$ to map the $I_0^{\text{GT}}$, which resembles the output of $R_\phi$, toward the enhanced ground truth $\hat{I}^\text{GT}$ generated by our framework.
The training objective for $R_\theta$ is given by:
\begin{equation}
  \mathcal{L}_\text{ref} = \| \hat{I} - \hat{I}^\text{GT} \|_2^2 \approx \| R_\theta(I_0^{\text{GT}}, \lambda) -  \hat{I}^\text{GT} \|_2^2.
\end{equation}
This refinement strategy is model-agnostic, allowing it to be flexibly applied on top of various restoration models without architectural modifications.

\begin{table*}[!t]
    \centering
    \caption{
        Evaluation of restoration performance of our ORNet on GoPro deblurring and SIDD denoising test sets.
    }
    \renewcommand{\arraystretch}{0.9}
    \setlength{\tabcolsep}{12.5pt}
    \vspace{-2mm}
    \scalebox{0.8}{
        \begin{tabular}{l l | cccc | cc | c}
            \toprule
            \rowcolor{gray!15}
                                                                                          &                                             & \multicolumn{4}{c|}{\textit{Perceptual Quality Metrics}} & \multicolumn{2}{c|}{\textit{VLM-based Metrics}} &                                                                                                                    \\
            \rowcolor{gray!15}
            {}                                                                            & {Method}                                    & {MUSIQ}$\uparrow$                                        & {MANIQA}$\uparrow$                              & {TOPIQ}$\uparrow$ & {LIQE}$\uparrow$ & {VisualQuality-R1}$\uparrow$ & {Q-Insight}$\uparrow$ & {A-FINE}$\downarrow$ \\
            \midrule
            \multirow{10.5}{*}{\rotatebox[origin=c]{90}{\large\textit{GoPro Deblurring}}} & \text{Restormer~\citep{zamir2022restormer}} & 45.05                                                    & 0.5265                                          & 0.3346            & 1.5264           & 4.1246                       & 3.4000                & 37.82                \\
                                                                                          & \text{NAFNet~\citep{chen2022simple}}        & 45.33                                                    & 0.5346                                          & 0.3368            & 1.5542           & 4.1539                       & 3.4262                & 36.98                \\
                                                                                          & \text{ResShift~\citep{yue2024efficient}}    & 44.30                                                    & 0.4934                                          & 0.3127            & 1.4090           & 3.9105                       & 3.3480                & 45.27                \\
                                                                                          & \text{IR-SDE~\citep{luo2023image}}          & 46.13                                                    & 0.5336                                          & 0.3410            & 1.6140           & 3.9735                       & 3.3619                & 44.98                \\
                                                                                          & \text{DiffIR~\citep{xia2023diffir}}         & 46.00                                                    & 0.5366                                          & 0.3412            & 1.5820           & 4.1544                       & 3.4269                & 37.14                \\
                                                                                          & \text{HI-Diff~\citep{chen2024hierarchical}} & 45.86                                                    & 0.5337                                          & 0.3398            & 1.5576           & 4.1554                       & 3.4207                & 37.21                \\
            \cmidrule{2-9}
                                                                                          & \text{AdaRevD~\citep{mao2024adarevd}}       & 45.49                                                    & 0.5363                                          & 0.3393            & 1.5660           & 4.1737                       & 3.4386                & 36.19                \\
                                                                                          & \text{+ ORNet \textbf{(Ours)}}              & \textbf{64.25}                                           & \textbf{0.5916}                                 & \textbf{0.4880}   & \textbf{2.4291}  & \textbf{4.1952}              & \textbf{3.5206}       & \textbf{14.38}       \\
            \cmidrule{2-9}
                                                                                          & \text{FFTformer~\citep{kong2023efficient}}  & 46.47                                                    & 0.5420                                          & 0.3456            & 1.6130           & 4.0942                       & 3.4569                & 36.91                \\
                                                                                          & \text{+ ORNet \textbf{(Ours)}}              & \textbf{64.57}                                           & \textbf{0.5949}                                 & \textbf{0.4924}   & \textbf{2.4664}  & \textbf{4.1995}              & \textbf{3.5278}       & \textbf{15.69}       \\
            \midrule
            \multirow{7.5}{*}{\rotatebox[origin=c]{90}{\large\textit{SIDD Denoising}}}    & \text{AP-BSN~\citep{lee2022ap}}             & 20.17                                                    & 0.3613                                          & 0.1977            & 1.0556           & 1.0170                       & 1.5496                & 58.18                \\
                                                                                          & \text{MIRNet-v2~\citep{zamir2020learning}}  & 22.18                                                    & 0.3770                                          & 0.2402            & 1.1855           & 1.0484                       & 1.6197                & 52.81                \\
                                                                                          & \text{Restormer~\citep{zamir2022restormer}} & 22.55                                                    & 0.3839                                          & 0.2439            & 1.2190           & 1.0653                       & 1.6620                & 52.36                \\
            \cmidrule{2-9}
                                                                                          & \text{Xformer~\citep{chen2024hierarchical}} & 22.57                                                    & 0.3828                                          & 0.2472            & 1.2040           & 1.0759                       & 1.6710                & 52.33                \\
                                                                                          & \text{+ ORNet \textbf{(Ours)}}              & \textbf{35.68}                                           & \textbf{0.4310}                                 & \textbf{0.3710}   & \textbf{1.9510}  & \textbf{1.3228}              & \textbf{2.1227}       & \textbf{40.25}       \\
            \cmidrule{2-9}
                                                                                          & \text{NAFNet~\citep{chen2022simple}}        & 22.73                                                    & 0.3937                                          & 0.2458            & 1.2189           & 1.0826                       & 1.7060                & 51.86                \\
                                                                                          & \text{+ ORNet \textbf{(Ours)}}              & \textbf{35.87}                                           & \textbf{0.4380}                                 & \textbf{0.3776}   & \textbf{1.9591}  & \textbf{1.3513}              & \textbf{2.1584}       & \textbf{40.98}       \\
            \bottomrule
        \end{tabular}
    }
    \label{tab:main_exp}
    \vspace{-2mm}
\end{table*}

\begin{table*}[!t]
    \centering
    \caption{
        Evaluation on an OOD environment. Additional Gaussian blur ($\sigma$ = 2.5) is applied to the blurry input images of the GoPro test set.
    }
    \renewcommand{\arraystretch}{0.9}
    \setlength{\tabcolsep}{10pt}
    \vspace{-2mm}
    \scalebox{0.8}{
        \begin{tabular}{l | ccc | ccc | cccc}
            \toprule
            \rowcolor{gray!15}
                                          & \multicolumn{3}{c|}{\textit{Original GT}} & \multicolumn{3}{c|}{\textit{Enhanced GT}} & \multicolumn{4}{c}{\textit{Perceptual Quality Metrics}}                                                                                                                                             \\
            \rowcolor{gray!15}
            {Method}                      & {PSNR}$\uparrow$                          & {SSIM}$\uparrow$                          & {LPIPS}$\downarrow$                                     & {PSNR}$\uparrow$ & {SSIM}$\uparrow$ & {LPIPS}$\downarrow$ & {MUSIQ}$\uparrow$ & {MANIQA}$\uparrow$ & {TOPIQ}$\uparrow$ & {LIQE}$\uparrow$ \\
            \midrule
            \text{FFTFormer~\citep{kong2023efficient}}              & 24.56                                   & 0.7532                                    & 0.4714                                                  & 23.68          & 0.7224           & 0.5441              & 22.3812           & 0.2284             & 0.1832            & 1.0108           \\
            \text{+ORNet \textbf{(Ours)}} & \textbf{24.58}                          & \textbf{0.7670}                           & \textbf{0.3429}                                         & \textbf{23.81} & \textbf{0.7405}  & \textbf{0.3777}     & \textbf{42.9131}  & \textbf{0.2638}    & \textbf{0.2646}   & \textbf{1.0656}  \\
            \bottomrule
        \end{tabular}
    }
    \label{tab:gopro_blur}
    \vspace{-2mm}
\end{table*}
\begin{table*}[!t]
    \centering
    \caption{
        Evaluation on an OOD environment. Additional white noise ($\sigma$ = 9) is applied to the blurry input images of the GoPro test set.
    }
    \renewcommand{\arraystretch}{0.9}
    \setlength{\tabcolsep}{10pt}
    \vspace{-2mm}
    \scalebox{0.8}{
        \begin{tabular}{l | ccc | ccc | cccc}
            \toprule
            \rowcolor{gray!15}
                                          & \multicolumn{3}{c|}{\textit{Original GT}} & \multicolumn{3}{c|}{\textit{Enhanced GT}} & \multicolumn{4}{c}{\textit{Perceptual Quality Metrics}}                                                                                                                                             \\
            \rowcolor{gray!15}
            {Method}                      & {PSNR}$\uparrow$                          & {SSIM}$\uparrow$                          & {LPIPS}$\downarrow$                                     & {PSNR}$\uparrow$ & {SSIM}$\uparrow$ & {LPIPS}$\downarrow$ & {MUSIQ}$\uparrow$ & {MANIQA}$\uparrow$ & {TOPIQ}$\uparrow$ & {LIQE}$\uparrow$ \\
            \midrule
            \text{FFTFormer~\citep{kong2023efficient}}              & 24.35                                   & 0.5867                                    & 0.4463                                                  & {23.83}         & {0.5713}         & {0.4751}            & 30.1430           & 0.4517             & 0.2655            & 1.1819           \\
            \text{+ORNet \textbf{(Ours)}} & \textbf{24.41}                          & \textbf{0.6179}                           & \textbf{0.4070}                                         & \textbf{23.97}  & \textbf{0.6057}  & \textbf{0.4233}     & \textbf{41.8760}  & \textbf{0.4699}    & \textbf{0.3188}   & \textbf{1.4321}  \\
            \bottomrule
        \end{tabular}
    }
    \label{tab:gopro_noise}
    \vspace{-4mm}
\end{table*}
\section{Experiments}
\label{sec:exp}
In this section, we evaluate the perceptual quality of both the enhanced ground truth images, and the outputs of our refinement network, through comprehensive quantitative and qualitative experiments.
In addition, we conduct user studies to assess the perceptual validity of both the enhanced ground truth and the refined outputs.

\subsection{Experimental Settings}

\paragraph{Implementation details}
For the supervision enhancement framework, we adopt OSEDiff~\citep{wu2024one} as the super-resolution network.
We generate three super-resolved ground truth variants using scaling factors of 2, 3, and 4.
The number of predefined masks, B, for constructing the final mask is set to 25.
Both the mask coefficient prediction network $g$ and the output refinement network (ORNet) are built using NAFBlocks, following the architectural design of NAFNet~\citep{chen2022simple}.
Specifically, $g$ consists of 4 NAFBlocks and 2 FC layers, while ORNet is built as a U-Net architecture of 4 encoder blocks, 1 middle block, and 4 decoder blocks.
Finally, the weighting parameter $\lambda$ used in our refinement formulation is set to 0.3 during evaluation.
A detailed description of how $\lambda$ is applied, along with additional evaluation results for different $\lambda$ values, is provided in the Appendix.

\paragraph{Training details}
We train two core networks: the mask coefficient prediction network $g$ and ORNet, using a combined dataset of GoPro~\citep{nah2017deep} and SIDD~\citep{abdelhamed2018high}.
Both $g$ and ORNet are trained for 100$K$ iterations with a batch size of 8, using random $512 \times 512$ crops.
AdamW~\citep{loshchilov2018decoupled} optimizer with cosine annealing learning rate scheduler is used.
The initial learning rate is $1 \times 10^{-4}$ for $g$, and $3 \times 10^{-4}$ for ORNet.
The parameter $\lambda$ is uniformly sampled from $[0, 1]$ during training to support learning of diverse enhancement levels.

\paragraph{Evaluation setup}
We evaluate models under two regimes: in-distribution (ID) and out-of-distribution (OOD).
The ID regime corresponds to standard setups using the provided GT images, whereas the OOD regime is constructed by applying synthetic degradations (\eg blur, noise) to test robustness.

In the ID setting, the provided GT images are suboptimal.
This limits the reliability of standard reference-based metrics such as PSNR, SSIM, and LPIPS~\citep{zhang2018unreasonable}, which measure pixel- or feature-level similarity to these imperfect GTs.
A higher PSNR against such GTs does not necessarily indicate better restoration, and outputs that surpass the GT quality may even receive lower scores.
To mitigate this limitation, ~\citet{chen2025toward} introduced A-FINE, an adaptive fidelity–naturalness evaluator.
Although reference-based, it provides a more comprehensive assessment than traditional similarity metrics; therefore, we measure A-FINE between the original GT and the model output.
Nevertheless, we mainly report no-reference perceptual metrics—MUSIQ~\citep{ke2021musiq}, MANIQA~\citep{yang2022maniqa}, TOPIQ~\citep{chen2024topiq}, LIQE~\citep{zhang2023blind}—as well as two recent VLM-based IQA measures (VisualQuality-R1~\citep{wu2025visualquality}, Q-Insight~\citep{li2025q}), which better align with human perception.

In the OOD setting, restored outputs are often far worse than the original GT due to the additional degradations.
In this case, the GT—though imperfect—still provides a meaningful reference for evaluating fidelity.
Thus, we complement the perceptual metrics with reference-based measures (PSNR, SSIM, LPIPS) computed against both the original and enhanced GTs, providing a comprehensive evaluation of both fidelity and perceptual quality.

\vspace{-1mm}
\subsection{Results}

\begin{figure*}[!t]
    \centering
    \scalebox{0.95}{
    \begin{minipage}{\textwidth}
    \centering
    \begin{subfigure}[b]{0.137\textwidth}\centering
        \includegraphics[width=\linewidth]{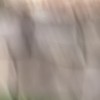}
    \end{subfigure}\hfill
    \begin{subfigure}[b]{0.137\textwidth}\centering
        \includegraphics[width=\linewidth]{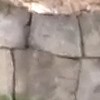}
    \end{subfigure}\hfill
    \begin{subfigure}[b]{0.137\textwidth}\centering
        \includegraphics[width=\linewidth]{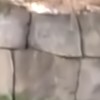}
    \end{subfigure}\hfill
    \begin{subfigure}[b]{0.137\textwidth}\centering
        \includegraphics[width=\linewidth]{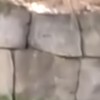}
    \end{subfigure}\hfill
    \begin{subfigure}[b]{0.137\textwidth}\centering
        \includegraphics[width=\linewidth]{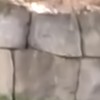}
    \end{subfigure}\hfill
    \begin{subfigure}[b]{0.137\textwidth}\centering
        \includegraphics[width=\linewidth]{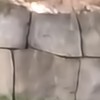}
    \end{subfigure}
    \begin{subfigure}[b]{0.137\textwidth}\centering
        \includegraphics[width=\linewidth]{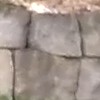}
    \end{subfigure}
    \vfill
    \begin{subfigure}[b]{0.137\textwidth}\centering
        \includegraphics[width=\linewidth]{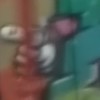}
        \small Blurry Input
    \end{subfigure}\hfill
    \begin{subfigure}[b]{0.137\textwidth}\centering
        \includegraphics[width=\linewidth]{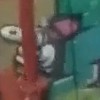}
        \small IR-SDE
    \end{subfigure}\hfill
    \begin{subfigure}[b]{0.137\textwidth}\centering
        \includegraphics[width=\linewidth]{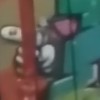}
        \small DiffIR
    \end{subfigure}\hfill
    \begin{subfigure}[b]{0.137\textwidth}\centering
        \includegraphics[width=\linewidth]{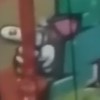}
        \small AdaRevID
    \end{subfigure}\hfill
    \begin{subfigure}[b]{0.137\textwidth}\centering
        \includegraphics[width=\linewidth]{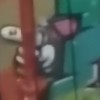}
        \small FFTformer
    \end{subfigure}\hfill
    \begin{subfigure}[b]{0.137\textwidth}\centering
        \includegraphics[width=\linewidth]{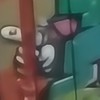}
        \small \bf{Ours}
    \end{subfigure}
    \begin{subfigure}[b]{0.137\textwidth}\centering
        \includegraphics[width=\linewidth]{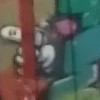}
        \small Original GT
    \end{subfigure}
    \end{minipage}
    }
    \vspace{-3mm}
    \caption{
Qualitative comparison of state-of-the-art deblurring methods, including ours (ORNet applied to FFTformer), on the GoPro dataset. 
Our method significantly improves the visual quality of the deblurred image. Zoom in for better visualization.
    }
    \vspace{-3mm}
    \label{fig:qual_GOPRO}
\end{figure*}


\begin{figure*}[!t]
    \centering
    \scalebox{0.95}{
    \begin{minipage}{\textwidth}
    \centering
    \begin{subfigure}[b]{0.137\textwidth}\centering
        \includegraphics[width=\linewidth]{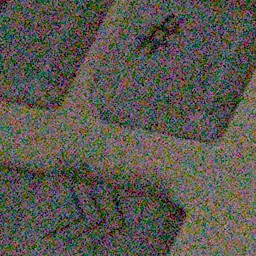}
    \end{subfigure}\hfill
    \begin{subfigure}[b]{0.137\textwidth}\centering
        \includegraphics[width=\linewidth]{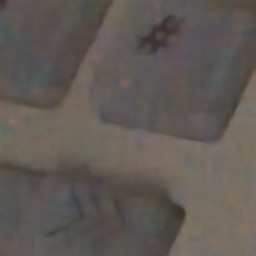}
    \end{subfigure}\hfill
    \begin{subfigure}[b]{0.137\textwidth}\centering
        \includegraphics[width=\linewidth]{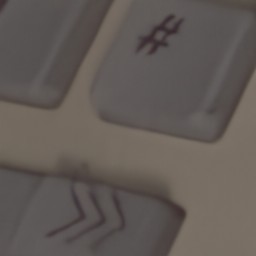}
    \end{subfigure}\hfill
    \begin{subfigure}[b]{0.137\textwidth}\centering
        \includegraphics[width=\linewidth]{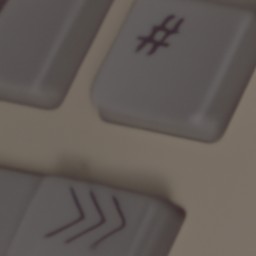}
    \end{subfigure}\hfill
    \begin{subfigure}[b]{0.137\textwidth}\centering
        \includegraphics[width=\linewidth]{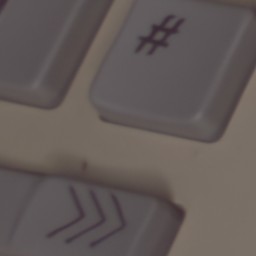}
    \end{subfigure}\hfill
    \begin{subfigure}[b]{0.137\textwidth}\centering
        \includegraphics[width=\linewidth]{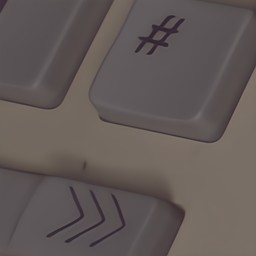}
    \end{subfigure}
    \begin{subfigure}[b]{0.137\textwidth}\centering
        \includegraphics[width=\linewidth]{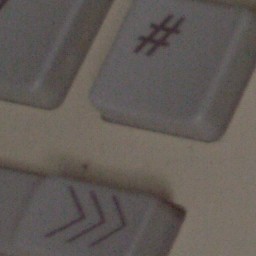}
    \end{subfigure}
    \vfill
    \begin{subfigure}[b]{0.137\textwidth}\centering
        \includegraphics[width=\linewidth]{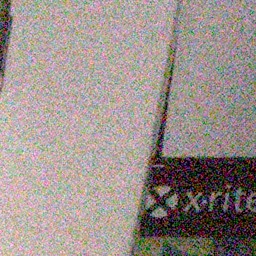}
        \small Noisy Input
    \end{subfigure}\hfill
    \begin{subfigure}[b]{0.137\textwidth}\centering
        \includegraphics[width=\linewidth]{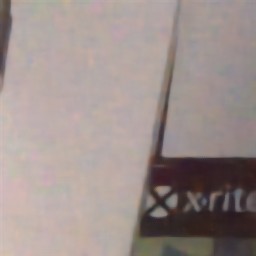}
        \small AP-BSN
    \end{subfigure}\hfill
    \begin{subfigure}[b]{0.137\textwidth}\centering
        \includegraphics[width=\linewidth]{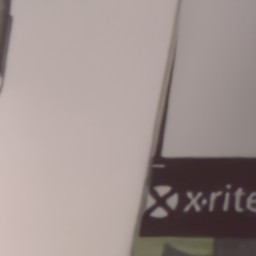}
        \small MIRNet
    \end{subfigure}\hfill
    \begin{subfigure}[b]{0.137\textwidth}\centering
        \includegraphics[width=\linewidth]{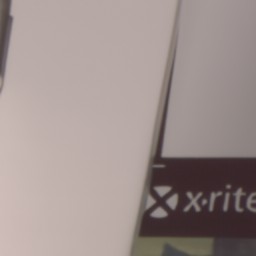}
        \small Xformer
    \end{subfigure}\hfill
    \begin{subfigure}[b]{0.137\textwidth}\centering
        \includegraphics[width=\linewidth]{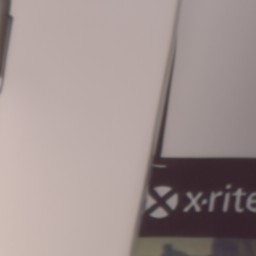}
        \small NAFNet
    \end{subfigure}\hfill
    \begin{subfigure}[b]{0.137\textwidth}\centering
        \includegraphics[width=\linewidth]{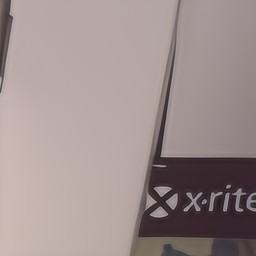}
        \small \bf{Ours}
    \end{subfigure}
    \begin{subfigure}[b]{0.137\textwidth}\centering
        \includegraphics[width=\linewidth]{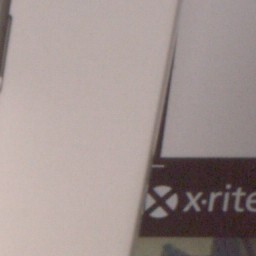}
        \small Original GT
    \end{subfigure}
    \end{minipage}
    }
    \vspace{-3mm}
    \caption{
Qualitative comparison of state-of-the-art denoising methods, including ours (ORNet applied to NAFNet), on the SIDD dataset. 
Our method significantly improves the visual quality of the denoised image. Zoom in for better visualization.
    }
    \vspace{-4mm}
    \label{fig:sidd_qual}
\end{figure*}

\subsubsection{In-distribution Quantitative results}

Table~\ref{tab:main_exp} reports the quantitative results on the GoPro deblurring and SIDD denoising datasets, respectively.
As ORNet is model agnostic, we apply it on top of representative base models from two restoration tasks.
For image deblurring, we integrate ORNet into AdaRevD~\citep{mao2024adarevd} and FFTformer~\citep{kong2023efficient};
for image denoising, we use NAFNet~\citep{chen2022simple} and Xformer~\citep{zhang2023xformer}.
We compare against diverse state-of-the-art methods, including Restormer~\citep{zamir2022restormer}, ResShift~\citep{yue2024efficient}, IR-SDE~\citep{luo2023image}, DiffIR~\citep{xia2023diffir}, HIDiff~\citep{chen2024hierarchical} for deblurring, and AP-BSN~\citep{lee2022ap}, MIRNet-v2~\citep{zamir2020learning}, Restormer for denoising.

Models trained with $\ell_1$ or $\ell_2$-loss and diffusion-based models tend to exhibit low performance.
This suggests that the restoration quality of existing models is upper-bounded by the quality of the original ground truth.
In contrast, our method leverages an enhanced ground truth, thereby achieving a significant improvement in perceptual quality.
\subsubsection{Out-of-distribution Quantitative results}

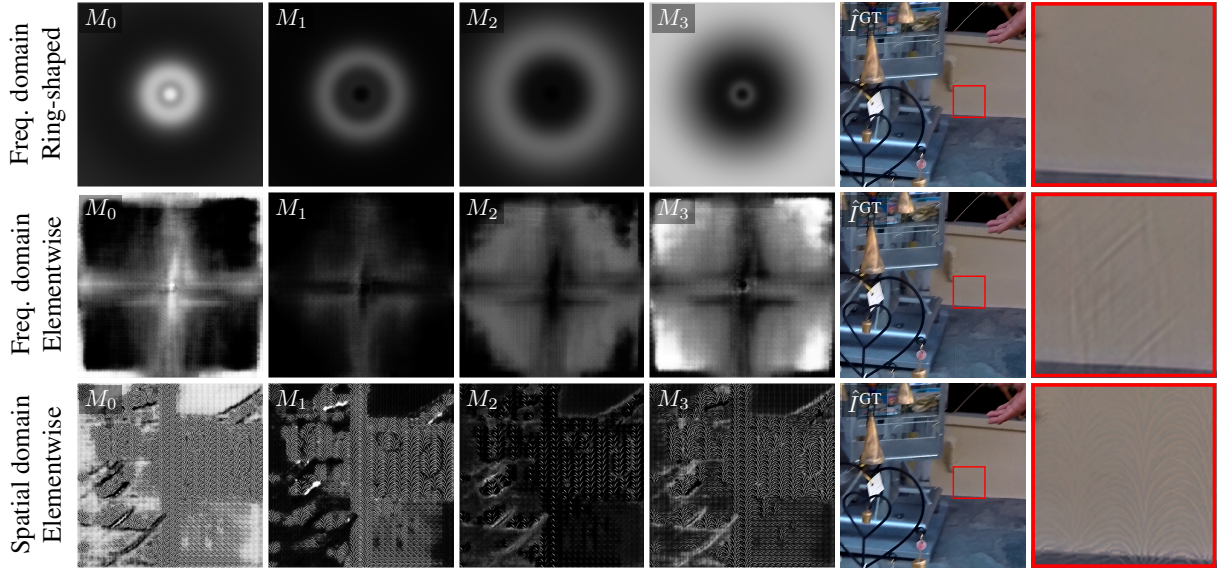
\begin{figure*}[t]
    \centering
    \begin{subfigure}[c]{0.04\textwidth}\centering
        \rotatebox{90}{\parbox{3cm}{\centering Freq. domain\\{Ring-shaped}}}
    \end{subfigure}\hspace{2mm}%
    \scalebox{.88}{%
    \begin{subfigure}[c]{0.16\textwidth}\centering
        \begin{tikzpicture}
            \node[anchor=south west,inner sep=0] (image) at (0,0) {
                \includegraphics[width=\linewidth]{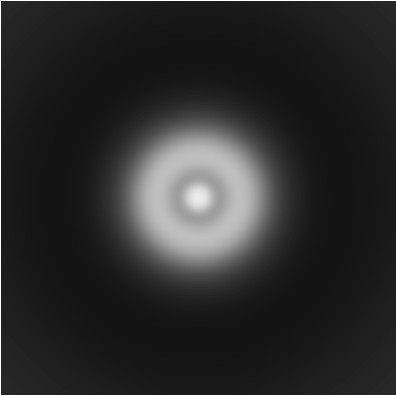}
            };
            \node[anchor=north west, fill=black, fill opacity=0.4, text=white, text opacity=1, inner sep=2pt, xshift=1pt, yshift=-1pt] at (image.north west) {$M_0$};
        \end{tikzpicture}
    \end{subfigure}
    \begin{subfigure}[c]{0.16\textwidth}\centering
        \begin{tikzpicture}
            \node[anchor=south west,inner sep=0] (image) at (0,0) {
                \includegraphics[width=\linewidth]{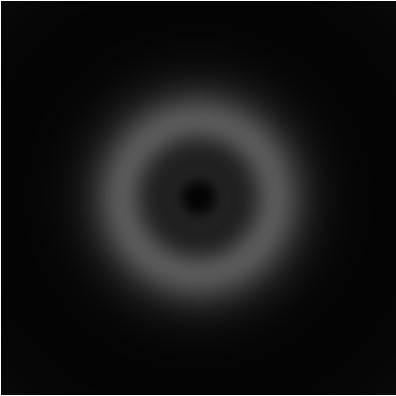}
            };
            \node[anchor=north west, fill=black, fill opacity=0.4, text=white, text opacity=1, inner sep=2pt, xshift=1pt, yshift=-1pt] at (image.north west) {$M_1$};
        \end{tikzpicture}
    \end{subfigure}
    \begin{subfigure}[c]{0.16\textwidth}\centering
        \begin{tikzpicture}
            \node[anchor=south west,inner sep=0] (image) at (0,0) {
                \includegraphics[width=\linewidth]{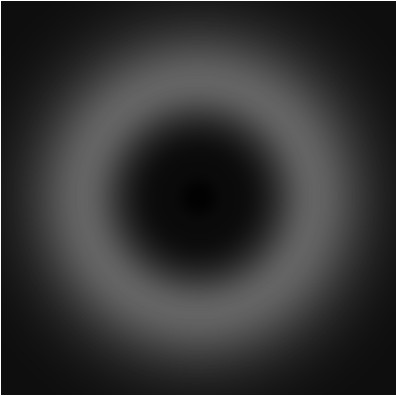}
            };
            \node[anchor=north west, fill=black, fill opacity=0.4, text=white, text opacity=1, inner sep=2pt, xshift=1pt, yshift=-1pt] at (image.north west) {$M_2$};
        \end{tikzpicture}
    \end{subfigure}
    \begin{subfigure}[c]{0.16\textwidth}\centering
        \begin{tikzpicture}
            \node[anchor=south west,inner sep=0] (image) at (0,0) {
                \includegraphics[width=\linewidth]{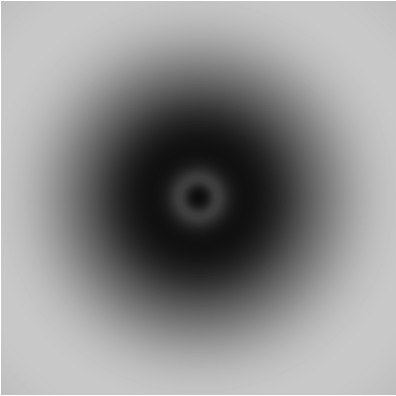}
            };
            \node[anchor=north west, fill=black, fill opacity=0.4, text=white, text opacity=1, inner sep=2pt, xshift=1pt, yshift=-1pt] at (image.north west) {$M_3$};
        \end{tikzpicture}
    \end{subfigure}
    \begin{subfigure}[c]{0.16\textwidth}\centering
        \begin{tikzpicture}
            \node[anchor=south west,inner sep=0] (image) at (0,0) {
                \includegraphics[width=\linewidth]{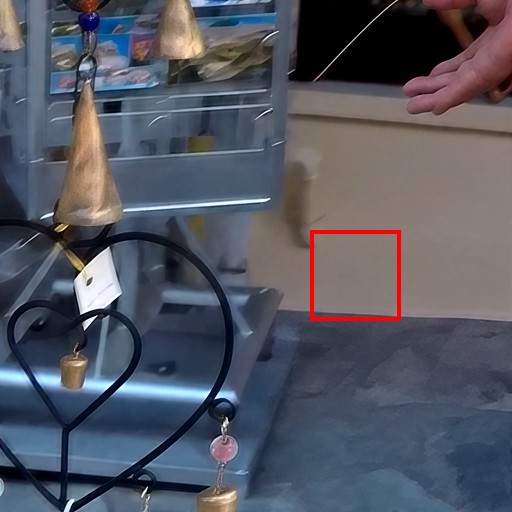}
            };
            \node[anchor=north west, fill=black, fill opacity=0.4, text=white, text opacity=1, inner sep=2pt, xshift=1pt, yshift=-1pt] at (image.north west) {$\hat{I}^{\text{GT}}$};
        \end{tikzpicture}
    \end{subfigure}
    \begin{subfigure}[c]{0.16\textwidth}\centering
        \begin{tikzpicture}
            \node[anchor=south west,inner sep=0] (image) at (0,0) {
                \includegraphics[width=\linewidth]{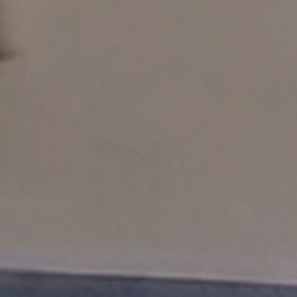}
            };
            \draw[red, line width=1.5pt] ([shift={(1pt,1pt)}]image.south west) rectangle ([shift={(-1pt,-1pt)}]image.north east);
        \end{tikzpicture}
    \end{subfigure}
    }\\[0pt]
    \vspace{-5mm}
    \begin{subfigure}[c]{0.04\textwidth}\centering
        \rotatebox{90}{\parbox{3cm}{\centering Freq. domain\\Elementwise}}
    \end{subfigure}\hspace{2mm}%
    \scalebox{.88}{%
    \begin{subfigure}[c]{0.16\textwidth}\centering
        \begin{tikzpicture}
            \node[anchor=south west,inner sep=0] (image) at (0,0) {
                \includegraphics[width=\linewidth]{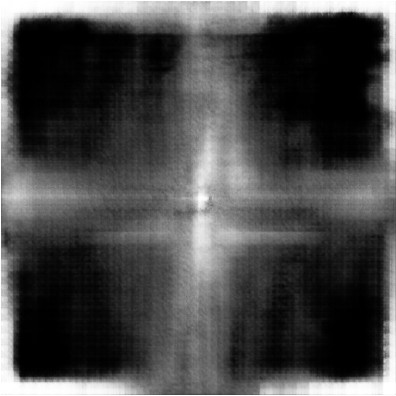}
            };
            \node[anchor=north west, fill=black, fill opacity=0.4, text=white, text opacity=1, inner sep=2pt, xshift=1pt, yshift=-1pt] at (image.north west) {$M_0$};
        \end{tikzpicture}
    \end{subfigure}
    \begin{subfigure}[c]{0.16\textwidth}\centering
        \begin{tikzpicture}
            \node[anchor=south west,inner sep=0] (image) at (0,0) {
                \includegraphics[width=\linewidth]{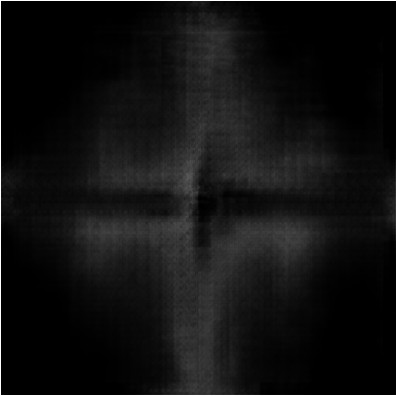}
            };
            \node[anchor=north west, fill=black, fill opacity=0.4, text=white, text opacity=1, inner sep=2pt, xshift=1pt, yshift=-1pt] at (image.north west) {$M_1$};
        \end{tikzpicture}
    \end{subfigure}
    \begin{subfigure}[c]{0.16\textwidth}\centering
        \begin{tikzpicture}
            \node[anchor=south west,inner sep=0] (image) at (0,0) {
                \includegraphics[width=\linewidth]{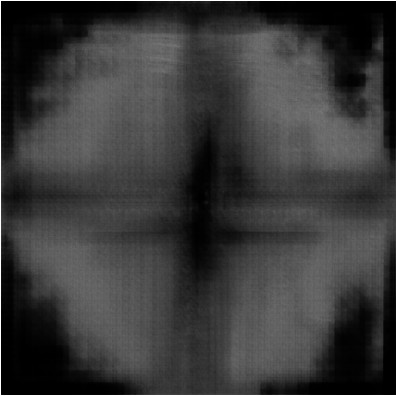}
            };
            \node[anchor=north west, fill=black, fill opacity=0.4, text=white, text opacity=1, inner sep=2pt, xshift=1pt, yshift=-1pt] at (image.north west) {$M_2$};
        \end{tikzpicture}
    \end{subfigure}
    \begin{subfigure}[c]{0.16\textwidth}\centering
        \begin{tikzpicture}
            \node[anchor=south west,inner sep=0] (image) at (0,0) {
                \includegraphics[width=\linewidth]{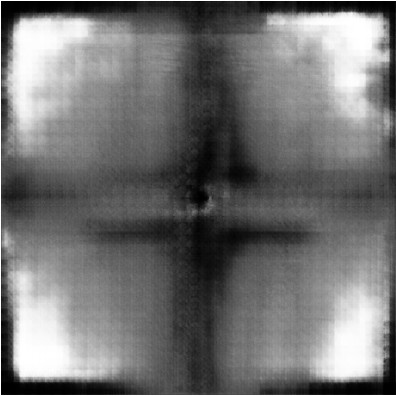}
            };
            \node[anchor=north west, fill=black, fill opacity=0.4, text=white, text opacity=1, inner sep=2pt, xshift=1pt, yshift=-1pt] at (image.north west) {$M_3$};
        \end{tikzpicture}
    \end{subfigure}
    \begin{subfigure}[c]{0.16\textwidth}\centering
        \begin{tikzpicture}
            \node[anchor=south west,inner sep=0] (image) at (0,0) {
                \includegraphics[width=\linewidth]{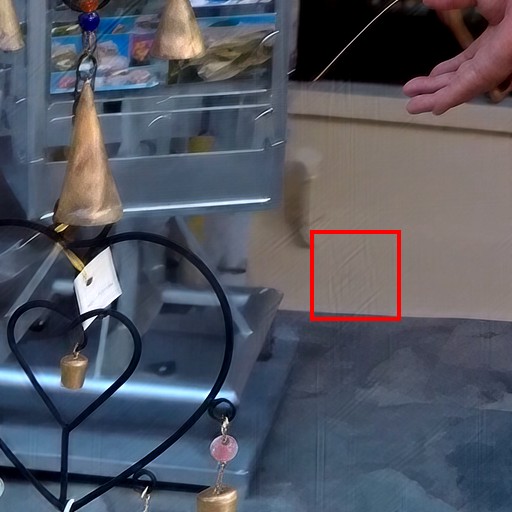}
            };
            \node[anchor=north west, fill=black, fill opacity=0.4, text=white, text opacity=1, inner sep=2pt, xshift=1pt, yshift=-1pt] at (image.north west) {$\hat{I}^{\text{GT}}$};
        \end{tikzpicture}
    \end{subfigure}
    \begin{subfigure}[c]{0.16\textwidth}\centering
        \begin{tikzpicture}
            \node[anchor=south west,inner sep=0] (image) at (0,0) {
                \includegraphics[width=\linewidth]{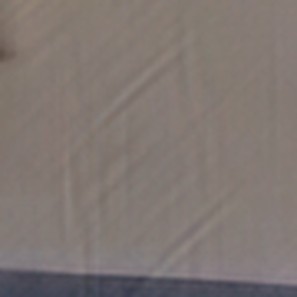}
            };
            \draw[red, line width=1.5pt] ([shift={(1pt,1pt)}]image.south west) rectangle ([shift={(-1pt,-1pt)}]image.north east);
        \end{tikzpicture}
    \end{subfigure}
    }\\[0pt]
    \vspace{-5mm}
    \begin{subfigure}[c]{0.04\textwidth}\centering
        \rotatebox{90}{\parbox{3cm}{\centering Spatial domain\\Elementwise}}
    \end{subfigure}\hspace{2mm}%
    \scalebox{.88}{%
    \begin{subfigure}[c]{0.16\textwidth}\centering
        \begin{tikzpicture}
            \node[anchor=south west,inner sep=0] (image) at (0,0) {
                \includegraphics[width=\linewidth]{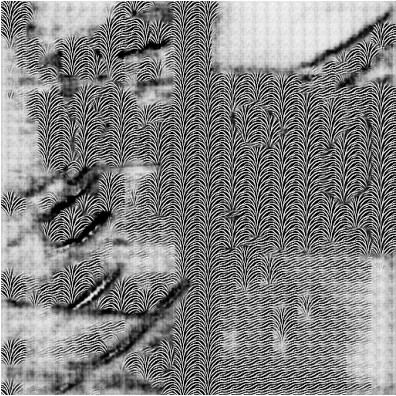}
            };
            \node[anchor=north west, fill=black, fill opacity=0.4, text=white, text opacity=1, inner sep=2pt, xshift=1pt, yshift=-1pt] at (image.north west) {$M_0$};
        \end{tikzpicture}
    \end{subfigure}
    \begin{subfigure}[c]{0.16\textwidth}\centering
        \begin{tikzpicture}
            \node[anchor=south west,inner sep=0] (image) at (0,0) {
                \includegraphics[width=\linewidth]{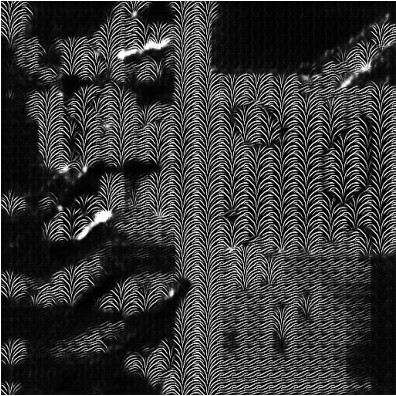}
            };
            \node[anchor=north west, fill=black, fill opacity=0.4, text=white, text opacity=1, inner sep=2pt, xshift=1pt, yshift=-1pt] at (image.north west) {$M_1$};
        \end{tikzpicture}
    \end{subfigure}
    \begin{subfigure}[c]{0.16\textwidth}\centering
        \begin{tikzpicture}
            \node[anchor=south west,inner sep=0] (image) at (0,0) {
                \includegraphics[width=\linewidth]{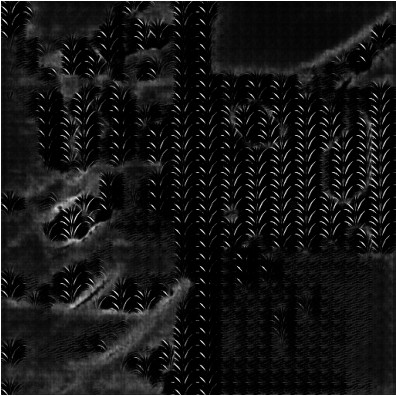}
            };
            \node[anchor=north west, fill=black, fill opacity=0.4, text=white, text opacity=1, inner sep=2pt, xshift=1pt, yshift=-1pt] at (image.north west) {$M_2$};
        \end{tikzpicture}
    \end{subfigure}
    \begin{subfigure}[c]{0.16\textwidth}\centering
        \begin{tikzpicture}
            \node[anchor=south west,inner sep=0] (image) at (0,0) {
                \includegraphics[width=\linewidth]{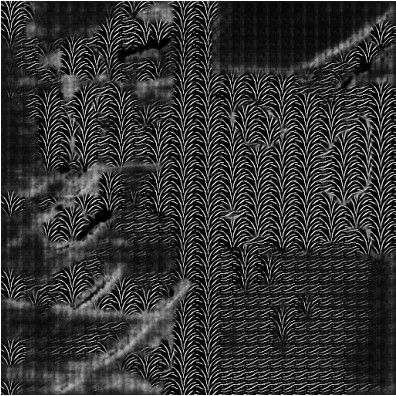}
            };
            \node[anchor=north west, fill=black, fill opacity=0.4, text=white, text opacity=1, inner sep=2pt, xshift=1pt, yshift=-1pt] at (image.north west) {$M_3$};
        \end{tikzpicture}
    \end{subfigure}
    \begin{subfigure}[c]{0.16\textwidth}\centering
        \begin{tikzpicture}
            \node[anchor=south west,inner sep=0] (image) at (0,0) {
                \includegraphics[width=\linewidth]{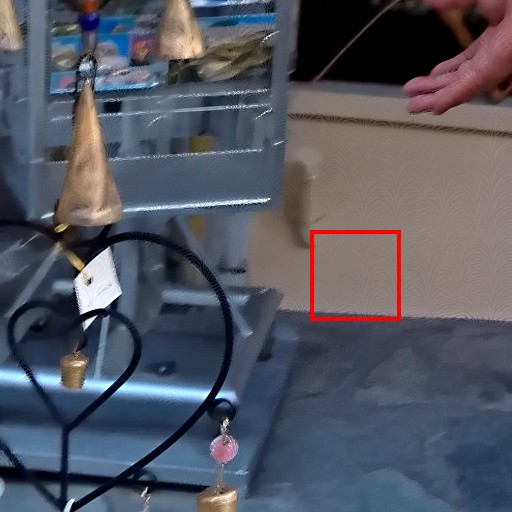}
            };
            \node[anchor=north west, fill=black, fill opacity=0.4, text=white, text opacity=1, inner sep=2pt, xshift=1pt, yshift=-1pt] at (image.north west) {$\hat{I}^{\text{GT}}$};
        \end{tikzpicture}
    \end{subfigure}
    \begin{subfigure}[c]{0.16\textwidth}\centering
        \begin{tikzpicture}
            \node[anchor=south west,inner sep=0] (image) at (0,0) {
                \includegraphics[width=\linewidth]{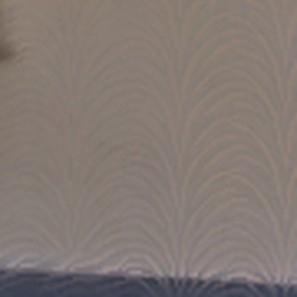}
            };
            \draw[red, line width=1.5pt] ([shift={(1pt,1pt)}]image.south west) rectangle ([shift={(-1pt,-1pt)}]image.north east);
        \end{tikzpicture}
    \end{subfigure}
    }

\vspace{-4.5mm}
    \caption{
        The top row shows the results when the conditional frequency mask generator is trained using our method.
        The second row shows the results when it is trained in an element-wise manner without ring-shaped Gaussian basis in frequency domain.
        The bottom row shows the results when it is trained in an element-wise manner in spatial domain.
        $M_i$ denotes the generated masks, and $\hat{I}^{\text{GT}}$ represents the enhanced ground truth generated using these masks.
        Zoom in for better visualization.
    }
    \vspace{-4mm}
    \label{fig:mask_analysis}
\end{figure*}

To evaluate the generalization performance of our refinement network, we conduct experiments in out-of-distribution (OOD) settings.
These are constructed by augmenting the inputs of the GoPro test set~\citep{nah2017deep} with additional, unseen degradations: one set with Gaussian blur and another with white noise.
We posit that existing state-of-the-art deblurring model FFTformer~\citep{kong2023efficient} overfit to the specific degradation characteristics of their GoPro training data, causing their performance to degrade sharply in such OOD conditions.
In contrast, our ORNet is not trained for a specific degradation; it robustly enhances the output of any given restoration model. As demonstrated in Tables~\ref{tab:gopro_blur} and~\ref{tab:gopro_noise}, applying ORNet leads to a substantial increase in perceptual quality. Simultaneously, reference-based metrics improve against both the original GT and our enhanced GT, which validates that ORNet also effectively preserves semantic details.
Additional analysis of the generalization performance of our ORNet is provided in the Appendix.
\subsubsection{Qualitative results}
We present qualitative results demonstrating the effectiveness of ORNet in refining restoration outputs.
Figures~\ref{fig:qual_GOPRO} and \ref{fig:sidd_qual} demonstrate comparisons on the GoPro and SIDD datasets, respectively.
When applied on top of existing restoration models, ORNet consistently enhances perceptual quality by sharpening fine structures.
Across diverse scenes, ORNet produces outputs with cleaner details and noticeably improved visual sharpness compared to those produced by the original restoration models.

\subsection{Ablation study}
\paragraph{Mask design}

Figure~\ref{fig:mask_analysis} presents a comparative analysis of frequency masks generated by three approaches: our conditional frequency mask generator with ring-shaped Gaussian bases, an element-wise baseline in the frequency domain, and an element-wise baseline in the spatial domain.
Our method constructs frequency masks as a weighted combination of smooth Gaussian bases, promoting spatially coherent mask patterns.
The base mask $M_0$ preserves low-frequency content from the original ground truth, while the additional masks $M_1$, $M_2$, and $M_3$ selectively incorporate high-frequency components from the super-resolved variants.
This structured decomposition enables fine-grained and interpretable frequency control across spatial regions.

In contrast, the element-wise baseline in the frequency domain employs a simple U-Net to directly predict mask values for each spatial coordinate in the frequency domain.
Although it exhibits a similar frequency-selection tendency, the resulting masks lack spatial consistency, which often leads to artifacts in the enhanced ground truth.
Likewise, the element-wise baseline in the spatial domain uses the same network architecture but applies the mixup operation directly in the spatial domain, resulting in similarly unstable and irregular mask patterns.
These inconsistencies arise because element-wise mask prediction provides no structural constraints, making it difficult for perceptual-loss–based optimization to avoid subtle but unnatural distortions.
By contrast, our formulation imposes smoothness and structural coherence, enabling stable frequency control while suppressing artifacts caused by irregular spatial variations.
\begin{table}[t]
    \center
    \caption{Comparison between our output refinement network (ORNet) and directly finetuning the restoration model using our enhanced supervision. FFTFormer* denotes the model finetuned on the GoPro training set where the original ground truth is replaced with our enhanced ground truth.}
    \renewcommand{\arraystretch}{0.9}
    \vspace{-1.5mm}
    \scalebox{0.85}{
        \setlength{\tabcolsep}{5.5pt}
        \begin{tabular}{l  | cc cccc}
            \toprule
            \rowcolor{gray!20}
            Method                                     & MUSIQ$\uparrow$                      & MANIQA$\uparrow$ & TOPIQ$\uparrow$ & LIQE$\uparrow$ \\
            \midrule

            \text{FFTformer}~\citep{kong2023efficient} & 46.47                                & 0.5420           & 0.3456          & 1.6130         \\
            \text{+ ORNet }                            & \bf{64.57}                                & \bf{0.5949}           & \bf{0.4924}          & \bf{2.4664}         \\
            \midrule
            FFTformer*~\citep{kong2023efficient}        & 60.54                                & 0.5854           & 0.4509          & 2.2359         \\
            \bottomrule
        \end{tabular}
    }
    \label{tab:full_finetuning}
    \vspace{-5mm}
\end{table}

\begin{figure}[t]
    \centering
    \begin{subfigure}[t]{0.48\linewidth}
        \hspace{-2mm}
        \includegraphics[width=\linewidth]{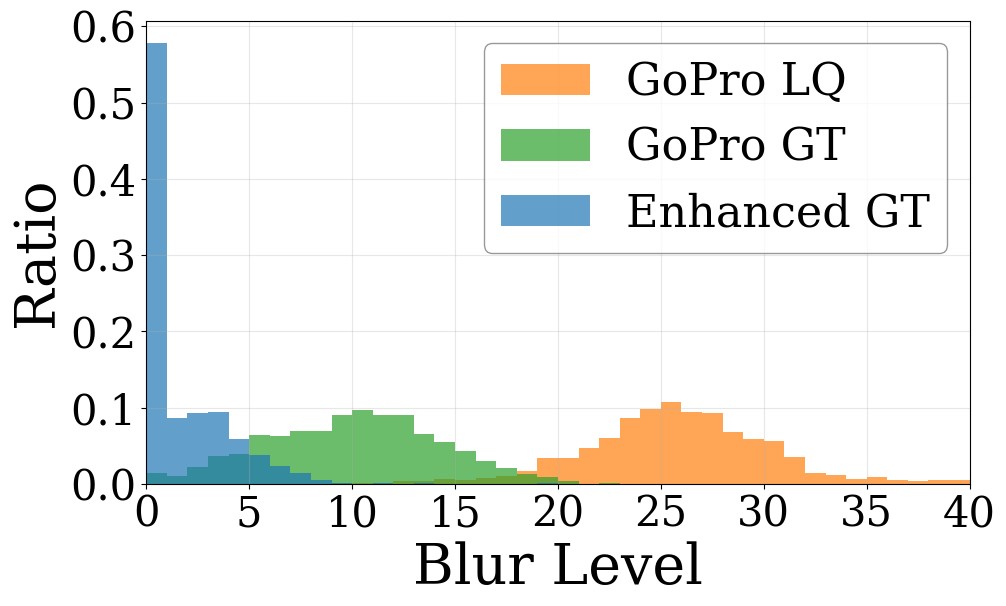}
        \vspace{-4mm}
    \end{subfigure}
    \begin{subfigure}[t]{0.48\linewidth}
        \includegraphics[width=\linewidth]{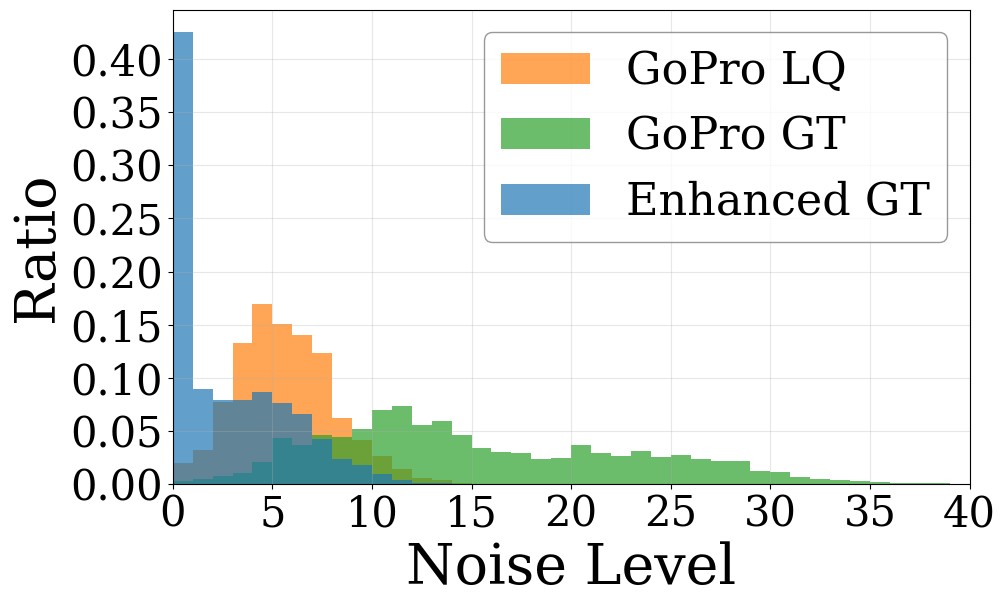}
        \vspace{-4mm}
    \end{subfigure}
    
    
    \begin{subfigure}[t]{0.48\linewidth}
        \hspace{-2mm}
        \includegraphics[width=\linewidth]{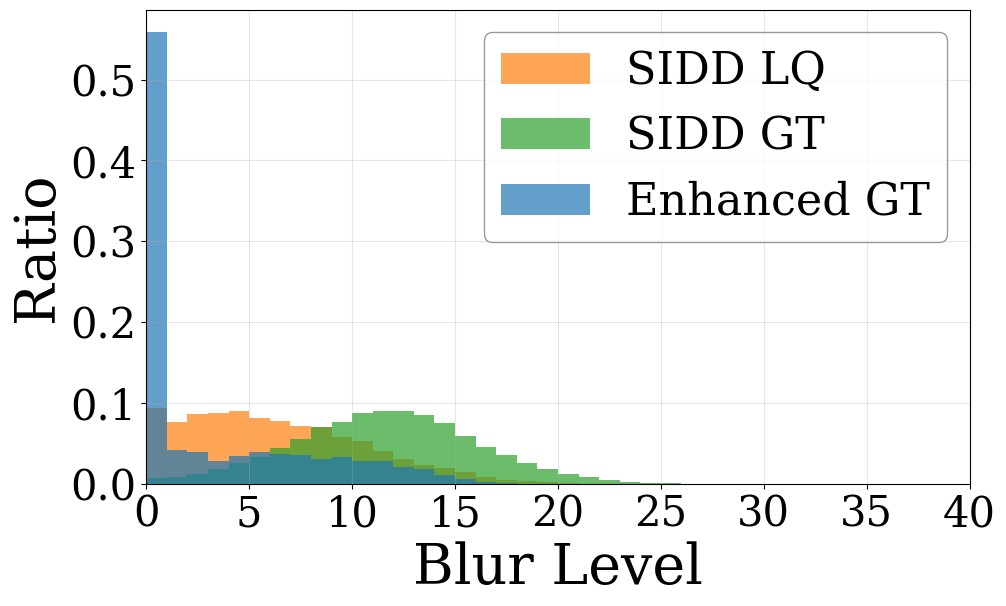}
        \vspace{-4mm}
    \end{subfigure}
    \begin{subfigure}[t]{0.48\linewidth}
        \includegraphics[width=\linewidth]{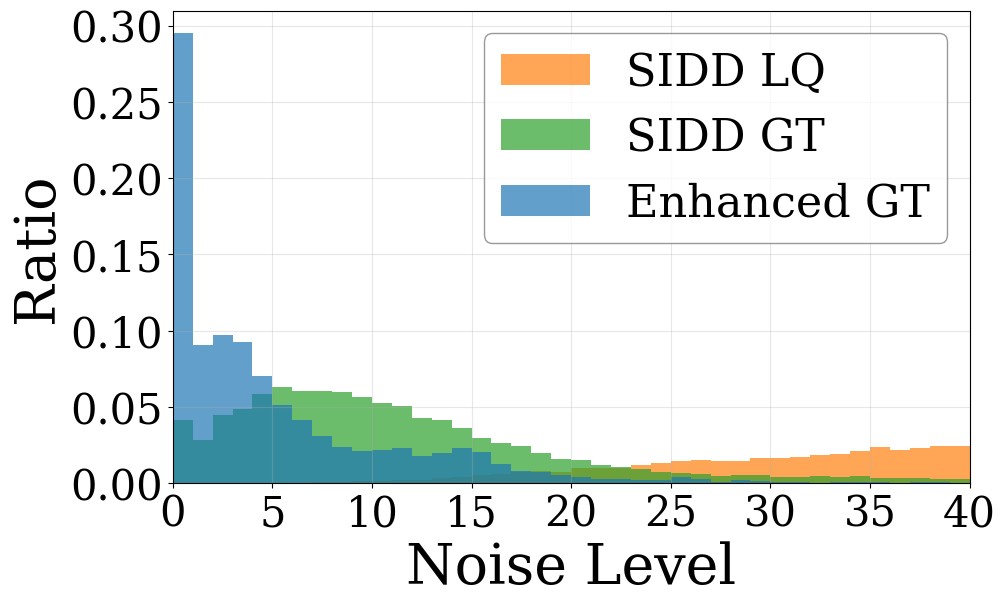}
        \vspace{-4mm}
    \end{subfigure}
    
    \vspace{-3mm}
    \caption{
        Our enhanced GT images demonstrate improved blur and noise levels, assessed with KonIQ++~\citep{su2021koniq++}. Histograms compare low quality (LQ), ground truth (GT), and enhanced GT images from GoPro~\citep{nah2017deep} (Top) and SIDD~\citep{abdelhamed2018high} (Bottom) datasets.
    }
    \label{fig:score_histogram}
    \vspace{-5mm}
\end{figure}
\vspace{-3mm}
\paragraph{Finetuning with enhanced supervision}
Our enhanced supervision can also be applied by directly finetuning existing restoration models.
Table~\ref{tab:full_finetuning} summarizes the results. The first row (FFTformer) reports the performance of the model pretrained on the original GoPro dataset, while the second row shows the effect of applying our refinement network to its outputs.
FFTFormer* corresponds to the pretrained FFTformer finetuned with our enhanced supervision.

We observe that finetuning with our enhanced supervision enables existing restoration models to achieve substantially higher perceptual quality in no-reference metrics.
However, this strategy requires retraining each model with its own enhanced supervision.
In contrast, our modular refinement network serves as a plug-and-play component that can be attached to various restoration models, achieving superior performance without additional per-model finetuning.

\vspace{-1mm}
\subsection{Analysis}
\label{sec:analysis}
\begin{figure}[t]
    \centering
    
    \begin{subfigure}[b]{0.325\linewidth}\centering
        \begin{tikzpicture}
            \node[anchor=south west,inner sep=0] (image) at (0,0) {
                \includegraphics[width=\linewidth]{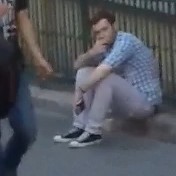}
            };
            \node[anchor=north west, fill=black, fill opacity=0.4, text=white, text opacity=1, inner sep=2pt] at (image.north west) {$I^\text{GT}_0$};
            \node[anchor=south west, fill=black, fill opacity=0.4, text=white, text opacity=1, inner sep=2pt, xshift=0pt, yshift=0pt, align=left] at (image.south west) {\scriptsize CLIP Similarity: -};
        \end{tikzpicture}
        
        Original GT
    \end{subfigure}
    \hfill
    \begin{subfigure}[b]{0.325\linewidth}\centering
        \begin{tikzpicture}
            \node[anchor=south west,inner sep=0] (image) at (0,0) {
                \includegraphics[width=\linewidth]{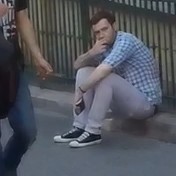}
            };
            \node[anchor=north west, fill=black, fill opacity=0.4, text=white, text opacity=1, inner sep=2pt] at (image.north west) {$\hat{I}^\text{GT}$};
            \node[anchor=south west, fill=black, fill opacity=0.4, text=white, text opacity=1, inner sep=2pt, xshift=0pt, yshift=0pt, align=left] at (image.south west) {\scriptsize CLIP Similarity: 0.9964};
        \end{tikzpicture}
        
        Enhanced GT
    \end{subfigure}
    \hfill
    \begin{subfigure}[b]{0.325\linewidth}\centering
        \begin{tikzpicture}
            \node[anchor=south west,inner sep=0] (image) at (0,0) {
                \includegraphics[width=\linewidth]{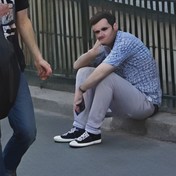}
            };
            \node[anchor=north west, fill=black, fill opacity=0.4, text=white, text opacity=1, inner sep=2pt] at (image.north west) {$I^\text{GT}_i$};
            \node[anchor=south west, fill=black, fill opacity=0.4, text=white, text opacity=1, inner sep=2pt, xshift=0pt, yshift=0pt, align=left] at (image.south west) {\scriptsize CLIP Similarity: 0.8995};
        \end{tikzpicture}
        
        Super-resolved GT
    \end{subfigure}

    \vspace{-2mm}
    \caption{
        Comparison of original, our enhanced, and super-resolved GTs.
        Our enhanced GT obtains high perceptual quality while maintaining semantic consistency.
        Zoom in for better visualization.
    }
    \vspace{-2mm}
    \label{fig:mixed_sample}
\end{figure}

\vspace{-1mm}
\paragraph{KonIQ++ analysis}
Figure~\ref{fig:score_histogram} shows the results of our supervision enhancement framework.
We assess the quality of images using KonIQ++~\citep{su2021koniq++} blur level ($\downarrow$) and noise level ($\downarrow$).
We observe that the ground truth (GT) images in the GoPro~\citep{nah2017deep} dataset, captured using single high-shutter-speed frames from action cameras, tend to be relatively noisy.
In addition, the GT images in the SIDD~\citep{abdelhamed2018high} dataset exhibit high blur scores, indicating that the averaging process used to obtain GT images introduces blurriness.
Our enhancement framework effectively improves the quality of such suboptimal GT images, reducing both blur and noise.
\vspace{-3mm}
\paragraph{Hallucination issue}
Although naive diffusion-based super-resolution methods often introduce undesirable distortions and artifacts, our supervision-enhancement approach is robust to hallucination and effectively preserves semantic consistency with the original GT.
Figure~\ref{fig:mixed_sample} shows that, even though one of the super-resolved GTs contains substantial artifacts on the man's face, hands, and shirt pattern, our enhanced GT through the proposed frequency-domain mixup strategy is almost free of such issues.

\begin{table}[t]
    \centering
    \caption{
Quantitative hallucination analysis on the GoPro dataset using OCR and face verification.
    }
    \vspace{-2mm}
    \setlength{\tabcolsep}{5.2pt}
    \renewcommand{\arraystretch}{1.0}
    \scalebox{0.85}{
        \begin{tabular}{l | cc | cc}
            \toprule
            \rowcolor{gray!20}
                                              & \multicolumn{2}{c|}{\textit{OCR}}          & \multicolumn{2}{c}{\textit{Face Recognition}} \\
            \rowcolor{gray!20}
                                        & Precision$\uparrow$    & Recall $\uparrow$    & Acc.$\uparrow$  & Avg. Cos. Sim.$\uparrow$ \\
            \midrule
            Original GT                                &  0.9130   &      0.9104                  &  -       &       -            \\  
            Enhanced GT                       &  \textbf{0.9287}   &      \textbf{0.9249}                  &  100\%       &       0.9133            \\  
    
            \bottomrule
        \end{tabular}
    }
    \label{tab:ocr_face}
\end{table}


\begin{table}[t]
	\centering
	\renewcommand{\arraystretch}{0.9}
	\setlength{\tabcolsep}{20pt}
	\caption{
		Parameters and MACs of existing restoration networks and our lightweightoutput refinement network.
	}
		\vspace{-2mm}
	\scalebox{0.85}{
		\begin{tabular}{l c c}
			\toprule
			\rowcolor{gray!20}
			\multicolumn{1}{l}{Architecture} & Params. (M) & MACs (G) \\
			\midrule
			AdaRev~\citep{mao2024adarevd}                           & 68.0        & 1386     \\
			FFTformer~\citep{kong2023efficient}                        & 14.9        & 525      \\
			NAFNet~\citep{chen2022simple}                           & 115.9       & 254      \\
			Xformer~\citep{chen2024hierarchical}                          & 25.1        & 571      \\

			\midrule
			\text{ORNet \textbf{(Ours)}}     & 4.5         & 20       \\

			\bottomrule
		\end{tabular}
	}
	\label{tab:params_flops}
	\vspace{-4mm}
\end{table}


For further quantitative validation, we evaluate two semantically sensitive downstream tasks: OCR and face recognition. 
For OCR, we extract corresponding text regions from the original and enhanced GT images across 83 samples, manually label them, and evaluate character-level accuracy using \texttt{EasyOCR}. 
For face recognition, we extract face regions and perform identity verification between the original and enhanced GTs across 566 samples via \texttt{InsightFace}, using a standard cosine similarity threshold of 0.5. 
As Table~\ref{tab:ocr_face} shows, OCR performance on the enhanced GT consistently surpasses that on the original GT and identity verification yields 100\% accuracy.
Since both tasks are highly sensitive to structural integrity and fine-grained semantic details, these results confirm that our approach restores authentic textures without introducing hallucinated artifacts.

\vspace{-4mm}
\paragraph{Computational efficiency}
Table~\ref{tab:params_flops} summarizes the number of parameters and multiply–accumulate operations (MACs) of our refinement network and existing restoration models, computed with an input resolution of 512$\times$512 pixels.
Our refinement network (ORNet) is significantly more lightweight than the baseline restoration models, enabling easy integration into existing architectures with negligible computational overhead.

\vspace{-4mm}
  \begin{figure}[t]
    \centering
    \scalebox{1}{
  \includegraphics[width=\linewidth]{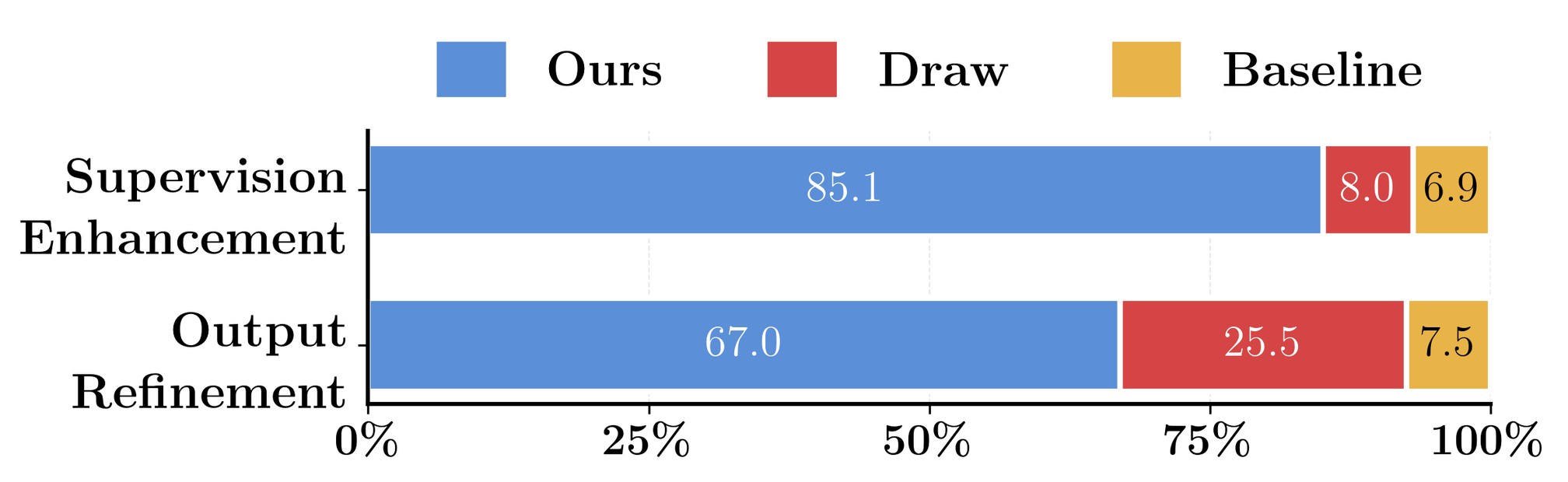}
  }
  \vspace{-7mm}
  \caption{
User preference study results. Participants consistently preferred our enhanced GT and ORNet outputs over the baselines.
}
  \label{fig:user}
  \vspace{-5mm}
\end{figure}
\paragraph{User study}
We conduct two user studies to evaluate the perceptual quality of our supervision enhancement and output refinement network.
Both studies involve 70 participants, each presented with 25 randomly sampled images from the GoPro dataset.
For supervision enhancement, participants are asked to compare the original ground truth (baseline) and enhanced ground truth (ours), based on how well each image appears to restore the low-quality input.
For the output refinement network, participants evaluate which output, FFTformer (baseline) or FFTformer + ORNet (ours), provides a better restoration of the low-quality input.
As shown in Figure~\ref{fig:user}, both our enhanced ground truth images and refinement outputs received significantly higher preference scores.



\section{Conclusion}
\label{sec:conclusion}
We introduce a novel supervision enhancement framework that addresses the fundamental limitation of current image restoration models: suboptimal ground truth images in real-world image restoration. 
By generating perceptually superior GT variants via super-resolution and optimally fusing them with original GT images in the frequency domain using adaptive masks, we achieve enhanced supervision targets.
Comprehensive evaluations, including user studies and diverse metrics, confirm that our method significantly improves the perceptual quality of the restored images.
This enhanced supervision then enables the training of a lightweight, model-agnostic refinement network, which can seamlessly integrate with existing restoration models to further boost their output.
We emphasize that our framework offers a practical path toward higher-fidelity and more visually compelling results in real-world restoration scenarios. 

\paragraph{Acknowledgements}
This work was partly supported by Samsung Electronics Co., Ltd (IO250418-12669-01).
It was also partly supported by the National Research Foundation of Korea (NRF) [RS-2022-NR070855, Trustworthy Artificial Intelligence] and the Institute of Information \& Communications Technology Planning \& Evaluation (IITP) [RS2022-II220959 (No.2022-0-00959), 
(Part 2) Few-Shot Learning of Causal Inference in Vision and Language for Decision Making; 
RS-2025-25442338, AI Star Fellowship Support Program (Seoul National Univ.);
No.RS-2021-II211343, Artificial Intelligence Graduate School Program (Seoul National University)], funded by the Korea government (MSIT).

\par
{
    \small
    \bibliographystyle{ieeenat_fullname}
    \bibliography{main}

@string{bmvc = {BMVC}}

@string{cvpr = {CVPR}}

@string{cvprw = {CVPRW}}

@string{eccv = {ECCV}}

@string{iccv = {ICCV}}

@string{iclr = {ICLR}}

@article{zhang2017beyond,
	author = {Zhang, Kai and Zuo, Wangmeng and Chen, Yunjin and Meng, Deyu and Zhang, Lei},
	journal = {IEEE transactions on image processing},
	number = {7},
	publisher = {IEEE},
	title = {Beyond a gaussian denoiser: Residual learning of deep cnn for image denoising},
	volume = {26},
	year = {2017}}

@inproceedings{chen2022simple,
	author = {Chen, Liangyu and Chu, Xiaojie and Zhang, Xiangyu and Sun, Jian},
	booktitle = {ECCV},
	title = {Simple baselines for image restoration},
	year = {2022}}

@inproceedings{zamir2022restormer,
	author = {Zamir, Syed Waqas and Arora, Aditya and Khan, Salman and Hayat, Munawar and Khan, Fahad Shahbaz and Yang, Ming-Hsuan},
	booktitle = {CVPR},
	title = {Restormer: Efficient transformer for high-resolution image restoration},
	year = {2022}}

@inproceedings{zamir2020learning,
	author = {Zamir, Syed Waqas and Arora, Aditya and Khan, Salman and Hayat, Munawar and Khan, Fahad Shahbaz and Yang, Ming-Hsuan and Shao, Ling},
	booktitle = {ECCV},
	title = {Learning enriched features for real image restoration and enhancement},
	year = {2020}}

@inproceedings{liang2021swinir,
	author = {Liang, Jingyun and Cao, Jiezhang and Sun, Guolei and Zhang, Kai and Van Gool, Luc and Timofte, Radu},
	booktitle = {ICCV},
	title = {Swinir: Image restoration using swin transformer},
	year = {2021}}

@inproceedings{lehtinen2018noise2noise,
	author = {Lehtinen, Jaakko and Munkberg, Jacob and Hasselgren, Jon and Laine, Samuli and Karras, Tero and Aittala, Miika and Aila, Timo},
	booktitle = {ICML},
	title = {Noise2Noise: Learning image restoration without clean data},
	year = {2018}}

@inproceedings{ulyanov2018deep,
	author = {Ulyanov, Dmitry and Vedaldi, Andrea and Lempitsky, Victor},
	booktitle = {CVPR},
	title = {Deep image prior},
	year = {2018}}

@inproceedings{abdelhamed2018high,
	author = {Abdelhamed, Abdelrahman and Lin, Stephen and Brown, Michael S},
	booktitle = {CVPR},
	title = {A high-quality denoising dataset for smartphone cameras},
	year = {2018}}

@inproceedings{lee2022ap,
	author = {Lee, Wooseok and Son, Sanghyun and Lee, Kyoung Mu},
	booktitle = {CVPR},
	title = {Ap-bsn: Self-supervised denoising for real-world images via asymmetric pd and blind-spot network},
	year = {2022}}

@article{xu2018real,
	author = {Xu, Jun and Li, Hui and Liang, Zhetong and Zhang, David and Zhang, Lei},
	journal = {arXiv preprint arXiv:1804.02603},
	title = {Real-world noisy image denoising: A new benchmark},
	year = {2018}}

@inproceedings{nam2016holistic,
	author = {Nam, Seonghyeon and Hwang, Youngbae and Matsushita, Yasuyuki and Kim, Seon Joo},
	booktitle = {CVPR},
	title = {A holistic approach to cross-channel image noise modeling and its application to image denoising},
	year = {2016}}

@inproceedings{loshchilov2018decoupled,
	author = {Ilya Loshchilov and Frank Hutter},
	booktitle = {ICLR},
	title = {Decoupled Weight Decay Regularization},
	year = {2019}}

@inproceedings{zhang2023xformer,
	author = {Zhang, Jiale and Zhang, Yulun and Gu, Jinjin and Dong, Jiahua and Kong, Linghe and Yang, Xiaokang},
	booktitle = {ICLR},
	title = {Xformer: Hybrid x-shaped transformer for image denoising},
	year = {2024}}

@inproceedings{ryou2024robust,
	author = {Ryou, Donghun and Ha, Inju and Yoo, Hyewon and Kim, Dongwan and Han, Bohyung},
	booktitle = {CVPR},
	title = {Robust Image Denoising through Adversarial Frequency Mixup},
	year = {2024}}

@inproceedings{nah2017deep,
	author = {Nah, Seungjun and Hyun Kim, Tae and Mu Lee, Kyoung},
	booktitle = {CVPR},
	title = {Deep multi-scale convolutional neural network for dynamic scene deblurring},
	year = {2017}}

@inproceedings{yu2024scaling,
	author = {Yu, Fanghua and Gu, Jinjin and Li, Zheyuan and Hu, Jinfan and Kong, Xiangtao and Wang, Xintao and He, Jingwen and Qiao, Yu and Dong, Chao},
	booktitle = {CVPR},
	title = {Scaling up to excellence: Practicing model scaling for photo-realistic image restoration in the wild},
	year = {2024}}

@inproceedings{wu2024seesr,
	author = {Wu, Rongyuan and Yang, Tao and Sun, Lingchen and Zhang, Zhengqiang and Li, Shuai and Zhang, Lei},
	booktitle = {CVPR},
	title = {Seesr: Towards semantics-aware real-world image super-resolution},
	year = {2024}}

@inproceedings{wu2024one,
	author = {Wu, Rongyuan and Sun, Lingchen and Ma, Zhiyuan and Zhang, Lei},
	booktitle = {NeurIPS},
	title = {One-step effective diffusion network for real-world image super-resolution},
	year = {2024}}

@inproceedings{zhang2018unreasonable,
	author = {Zhang, Richard and Isola, Phillip and Efros, Alexei A and Shechtman, Eli and Wang, Oliver},
	booktitle = {CVPR},
	title = {The unreasonable effectiveness of deep features as a perceptual metric},
	year = {2018}}

@inproceedings{ke2021musiq,
	author = {Ke, Junjie and Wang, Qifei and Wang, Yilin and Milanfar, Peyman and Yang, Feng},
	booktitle = {ICCV},
	title = {Musiq: Multi-scale image quality transformer},
	year = {2021}}

@inproceedings{yang2022maniqa,
	author = {Yang, Sidi and Wu, Tianhe and Shi, Shuwei and Lao, Shanshan and Gong, Yuan and Cao, Mingdeng and Wang, Jiahao and Yang, Yujiu},
	booktitle = {CVPR},
	title = {Maniqa: Multi-dimension attention network for no-reference image quality assessment},
	year = {2022}}

@article{chen2024topiq,
	author = {Chen, Chaofeng and Mo, Jiadi and Hou, Jingwen and Wu, Haoning and Liao, Liang and Sun, Wenxiu and Yan, Qiong and Lin, Weisi},
	journal = {IEEE Transactions on Image Processing},
	publisher = {IEEE},
	title = {Topiq: A top-down approach from semantics to distortions for image quality assessment},
	year = {2024}}

@inproceedings{zhang2023blind,
	author = {Zhang, Weixia and Zhai, Guangtao and Wei, Ying and Yang, Xiaokang and Ma, Kede},
	booktitle = {CVPR},
	title = {Blind image quality assessment via vision-language correspondence: A multitask learning perspective},
	year = {2023}}

@inproceedings{luo2023image,
	author = {Luo, Ziwei and Gustafsson, Fredrik K and Zhao, Zheng and Sj{\"o}lund, Jens and Sch{\"o}n, Thomas B},
	booktitle = {ICML},
	title = {Image Restoration with Mean-Reverting Stochastic Differential Equations},
	year = {2023}}

@inproceedings{xia2023diffir,
	author = {Xia, Bin and Zhang, Yulun and Wang, Shiyin and Wang, Yitong and Wu, Xinglong and Tian, Yapeng and Yang, Wenming and Van Gool, Luc},
	booktitle = {ICCV},
	title = {Diffir: Efficient diffusion model for image restoration},
	year = {2023}}

@inproceedings{chen2024hierarchical,
	author = {Chen, Zheng and Zhang, Yulun and Liu, Ding and Gu, Jinjin and Kong, Linghe and Yuan, Xin and others},
	booktitle = {NeurIPS},
	title = {Hierarchical integration diffusion model for realistic image deblurring},
	year = {2024}}

@article{liu2025one,
	author = {Liu, Xiaoyang and Wang, Yuquan and Chen, Zheng and Cao, Jiezhang and Zhang, He and Zhang, Yulun and Yang, Xiaokang},
	journal = {arXiv preprint arXiv:2503.06537},
	title = {One-Step Diffusion Model for Image Motion-Deblurring},
	year = {2025}}

@inproceedings{kong2023efficient,
	author = {Kong, Lingshun and Dong, Jiangxin and Ge, Jianjun and Li, Mingqiang and Pan, Jinshan},
	booktitle = {CVPR},
	title = {Efficient frequency domain-based transformers for high-quality image deblurring},
	year = {2023}}

@inproceedings{mao2024adarevd,
	author = {Mao, Xintian and Li, Qingli and Wang, Yan},
	booktitle = {CVPR},
	title = {Adarevd: Adaptive patch exiting reversible decoder pushes the limit of image deblurring},
	year = {2024}}

@inproceedings{dong2014learning,
	author = {Dong, Chao and Loy, Chen Change and He, Kaiming and Tang, Xiaoou},
	booktitle = {ECCV},
	title = {Learning a deep convolutional network for image super-resolution},
	year = {2014}}

@inproceedings{ledig2017photo,
	author = {Ledig, Christian and Theis, Lucas and Husz{\'a}r, Ferenc and Caballero, Jose and Cunningham, Andrew and Acosta, Alejandro and Aitken, Andrew and Tejani, Alykhan and Totz, Johannes and Wang, Zehan and others},
	booktitle = {CVPR},
	title = {Photo-realistic single image super-resolution using a generative adversarial network},
	year = {2017}}

@inproceedings{kupyn2018deblurgan,
	author = {Kupyn, Orest and Budzan, Volodymyr and Mykhailych, Mykola and Mishkin, Dmytro and Matas, Ji{\v{r}}{\'\i}},
	booktitle = {CVPR},
	title = {Deblurgan: Blind motion deblurring using conditional adversarial networks},
	year = {2018}}

@inproceedings{yoo2020rethinking,
	author = {Yoo, Jaejun and Ahn, Namhyuk and Sohn, Kyung-Ah},
	booktitle = {CVPR},
	title = {Rethinking data augmentation for image super-resolution: A comprehensive analysis and a new strategy},
	year = {2020}}

@inproceedings{wu2024id,
	author = {Wu, Jia-Hao and Tsai, Fu-Jen and Peng, Yan-Tsung and Tsai, Chung-Chi and Lin, Chia-Wen and Lin, Yen-Yu},
	booktitle = {CVPR},
	title = {Id-blau: Image deblurring by implicit diffusion-based reblurring augmentation},
	year = {2024}}

@inproceedings{ha2024learning,
	author = {Ha, Inju and Ryou, Donghun and Seo, Seonguk and Han, Bohyung},
	booktitle = {CVPR Findings},
	title = {Learning to Translate Noise for Robust Image Denoising},
	year = {2026}}

@inproceedings{zhang2022self,
	author = {Zhang, Zhilu and Xu, RongJian and Liu, Ming and Yan, Zifei and Zuo, Wangmeng},
	booktitle = {NeurIPS},
	title = {Self-supervised image restoration with blurry and noisy pairs},
	year = {2022}}

@article{wang2024exploiting,
	author = {Wang, Jianyi and Yue, Zongsheng and Zhou, Shangchen and Chan, Kelvin CK and Loy, Chen Change},
	journal = {International Journal of Computer Vision},
	number = {12},
	pages = {5929--5949},
	publisher = {Springer},
	title = {Exploiting diffusion prior for real-world image super-resolution},
	volume = {132},
	year = {2024}}

@inproceedings{lin2024diffbir,
	author = {Lin, Xinqi and He, Jingwen and Chen, Ziyan and Lyu, Zhaoyang and Dai, Bo and Yu, Fanghua and Qiao, Yu and Ouyang, Wanli and Dong, Chao},
	booktitle = {ECCV},
	title = {Diffbir: Toward blind image restoration with generative diffusion prior},
	year = {2024}}

@inproceedings{guo2024mambair,
	author = {Guo, Hang and Li, Jinmin and Dai, Tao and Ouyang, Zhihao and Ren, Xudong and Xia, Shu-Tao},
	booktitle = {ECCV},
	title = {Mambair: A simple baseline for image restoration with state-space model},
	year = {2024}}

@article{guo2024mambairv2,
	author = {Guo, Hang and Guo, Yong and Zha, Yaohua and Zhang, Yulun and Li, Wenbo and Dai, Tao and Xia, Shu-Tao and Li, Yawei},
	journal = {arXiv preprint arXiv:2411.15269},
	title = {MambaIRv2: Attentive State Space Restoration},
	year = {2024}}

@inproceedings{shen2019human,
	author = {Shen, Ziyi and Wang, Wenguan and Lu, Xiankai and Shen, Jianbing and Ling, Haibin and Xu, Tingfa and Shao, Ling},
	booktitle = {ICCV},
	title = {Human-aware motion deblurring},
	year = {2019}}

@inproceedings{nah2019ntire,
	author = {Nah, Seungjun and Baik, Sungyong and Hong, Seokil and Moon, Gyeongsik and Son, Sanghyun and Timofte, Radu and Mu Lee, Kyoung},
	booktitle = {CVPRW},
	title = {Ntire 2019 challenge on video deblurring and super-resolution: Dataset and study},
	year = {2019}}

@inproceedings{su2021koniq++,
	author = {Su, Shaolin and Hosu, Vlad and Lin, Hanhe and Zhang, Yanning and Saupe, Dietmar},
	booktitle = {BMVC},
	title = {Koniq++: Boosting no-reference image quality assessment in the wild by jointly predicting image quality and defects},
	year = {2021}}

@article{delbracio2023inversion,
	author = {Delbracio, Mauricio and Milanfar, Peyman},
	journal = {arXiv preprint arXiv:2303.11435},
	title = {Inversion by direct iteration: An alternative to denoising diffusion for image restoration},
	year = {2023}}

@inproceedings{cai2023retinexformer,
	author = {Cai, Yuanhao and Bian, Hao and Lin, Jing and Wang, Haoqian and Timofte, Radu and Zhang, Yulun},
	booktitle = {ICCV},
	title = {Retinexformer: One-stage retinex-based transformer for low-light image enhancement},
	year = {2023}}

@article{yan2025hvi,
	author = {Yan, Qingsen and Feng, Yixu and Zhang, Cheng and Pang, Guansong and Shi, Kangbiao and Wu, Peng and Dong, Wei and Sun, Jinqiu and Zhang, Yanning},
	journal = {arXiv preprint arXiv:2502.20272},
	title = {HVI: A New color space for Low-light Image Enhancement},
	year = {2025}}

@article{wei2018deep,
	author = {Wei, Chen and Wang, Wenjing and Yang, Wenhan and Liu, Jiaying},
	journal = {arXiv preprint arXiv:1808.04560},
	title = {Deep retinex decomposition for low-light enhancement},
	year = {2018}}

@inproceedings{agustsson2017ntire,
	author = {Agustsson, Eirikur and Timofte, Radu},
	booktitle = {CVPRW},
	title = {Ntire 2017 challenge on single image super-resolution: Dataset and study},
	year = {2017}}

@inproceedings{li2023lsdir,
	author = {Li, Yawei and Zhang, Kai and Liang, Jingyun and Cao, Jiezhang and Liu, Ce and Gong, Rui and Zhang, Yulun and Tang, Hao and Liu, Yun and Demandolx, Denis and others},
	booktitle = {CVPR},
	title = {Lsdir: A large scale dataset for image restoration},
	year = {2023}}

@article{yue2024efficient,
	author = {Yue, Zongsheng and Wang, Jianyi and Loy, Chen Change},
	journal = {IEEE Transactions on Pattern Analysis and Machine Intelligence},
	publisher = {IEEE},
	title = {Efficient diffusion model for image restoration by residual shifting},
	year = {2024}}

@inproceedings{ohayon2021high,
	author = {Ohayon, Guy and Adrai, Theo and Vaksman, Gregory and Elad, Michael and Milanfar, Peyman},
	booktitle = {ICCV},
	title = {High perceptual quality image denoising with a posterior sampling cgan},
	year = {2021}}

@inproceedings{kawar2022denoising,
	author = {Kawar, Bahjat and Elad, Michael and Ermon, Stefano and Song, Jiaming},
	booktitle = {NeurIPS},
	title = {Denoising diffusion restoration models},
	year = {2022}}

@inproceedings{zhu2023denoising,
	author = {Zhu, Yuanzhi and Zhang, Kai and Liang, Jingyun and Cao, Jiezhang and Wen, Bihan and Timofte, Radu and Van Gool, Luc},
	booktitle = {CVPR},
	title = {Denoising diffusion models for plug-and-play image restoration},
	year = {2023}}

@inproceedings{chen2025toward,
	author = {Chen, Du and Wu, Tianhe and Ma, Kede and Zhang, Lei},
	booktitle = {CVPR},
	title = {Toward generalized image quality assessment: Relaxing the perfect reference quality assumption},
	year = {2025}}

@article{wu2025visualquality,
	author = {Wu, Tianhe and Zou, Jian and Liang, Jie and Zhang, Lei and Ma, Kede},
	journal = {arXiv preprint arXiv:2505.14460},
	title = {VisualQuality-R1: Reasoning-Induced Image Quality Assessment via Reinforcement Learning to Rank},
	year = {2025}}

@article{li2025q,
	author = {Li, Weiqi and Zhang, Xuanyu and Zhao, Shijie and Zhang, Yabin and Li, Junlin and Zhang, Li and Zhang, Jian},
	journal = {arXiv preprint arXiv:2503.22679},
	title = {Q-insight: Understanding image quality via visual reinforcement learning},
	year = {2025}}

@inproceedings{yang2025gyro,
	author = {Yang, Heemin and Rim, Jaesung and Lee, Seungyong and Baek, Seung-Hwan and Cho, Sunghyun},
	booktitle = {CVPR},
	title = {Gyro-based neural single image deblurring},
	year = {2025}}

@article{kang2025icm,
  title={ICM-SR: Image-Conditioned Manifold Regularization for Image Super-Resolution},
  author={Kang, Junoh and Ryou, Donghun and Han, Bohyung},
  journal={arXiv preprint arXiv:2511.22048},
  year={2025}
}

@inproceedings{Sun_2025_CVPR,
    author    = {Sun, Lei and Guo, Hang and Ren, Bin and Van Gool, Luc and Timofte, Radu and Li, Yawei},
    title     = {The Tenth NTIRE 2025 Image Denoising Challenge Report},
    booktitle = {CVPRW},
    year      = {2025},
}

@inproceedings{Chen_2025_CVPR,
    author    = {Chen, Zheng and Liu, Kai and Gong, Jue and Wang, Jingkai and Sun, Lei and Wu, Zongwei and Timofte, Radu and others},
    title     = {NTIRE 2025 Challenge on Image Super-Resolution (x4): Methods and Results},
    booktitle = {CVPRW},
    year      = {2025},
}

@article{kim2022fine,
  title={Fine-grained neural architecture search for image super-resolution},
  author={Kim, Heewon and Hong, Seokil and Han, Bohyung and Myeong, Heesoo and Lee, Kyoung Mu},
  journal={Journal of Visual Communication and Image Representation},
  volume={89},
  pages={103654},
  year={2022},
  publisher={Elsevier}
}
}

\makeatletter
\renewcommand{\maketitlesupplementary}{%
   \newpage
   \onecolumn
   \begin{center}
        \Large
        \textbf{\thetitle}\\
        \vspace{0.5em}Supplementary Material \\
        \vspace{0.5em}
        \normalsize\@author
        \vspace{1.0em}
   \end{center}
}
\makeatother

\maketitlesupplementary

\begingroup
\renewcommand\thefootnote{} 
\footnotetext{\protect\hspace{-0.5em}${}^*$indicates equal contribution.}
\endgroup

\section{Additional implementation details}
\subsection{Basis masks}
\vspace{-2mm}
\begin{figure*}[h]
  \centering
  \begin{subfigure}[b]{0.074\textwidth}\centering
    \includegraphics[width=\linewidth]{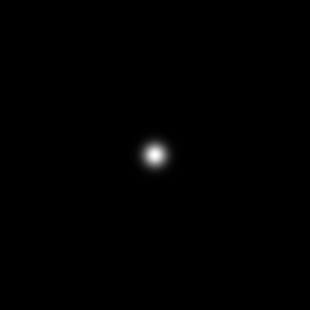}
    $R_1$
  \end{subfigure}\hfill
  \begin{subfigure}[b]{0.074\textwidth}\centering
    \includegraphics[width=\linewidth]{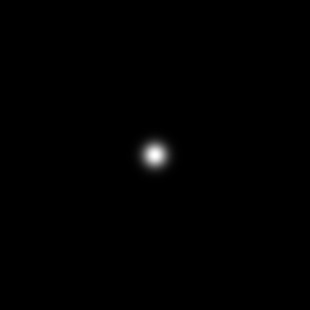}
    $R_2$
  \end{subfigure}\hfill
  \begin{subfigure}[b]{0.074\textwidth}\centering
    \includegraphics[width=\linewidth]{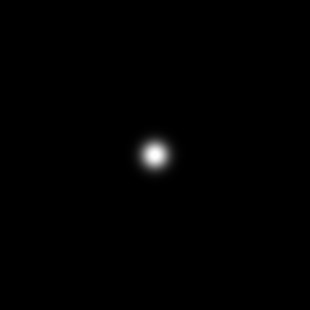}
    $R_3$
  \end{subfigure}\hfill
  \begin{subfigure}[b]{0.074\textwidth}\centering
    \includegraphics[width=\linewidth]{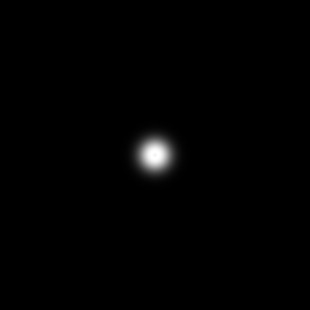}
    $R_4$
  \end{subfigure}\hfill
  \begin{subfigure}[b]{0.074\textwidth}\centering
    \includegraphics[width=\linewidth]{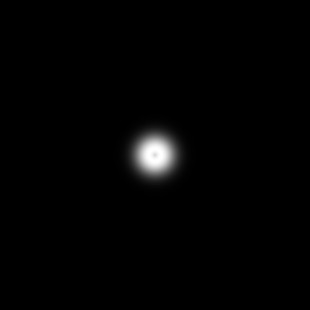}
    $R_5$
  \end{subfigure}\hfill
  \begin{subfigure}[b]{0.074\textwidth}\centering
    \includegraphics[width=\linewidth]{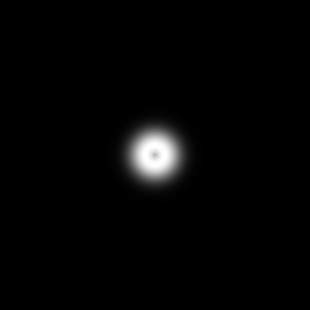}
    $R_6$
  \end{subfigure}\hfill
  \begin{subfigure}[b]{0.074\textwidth}\centering
    \includegraphics[width=\linewidth]{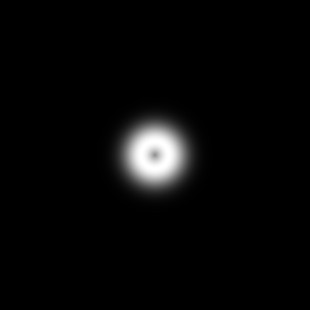}
    $R_7$
  \end{subfigure}\hfill
  \begin{subfigure}[b]{0.074\textwidth}\centering
    \includegraphics[width=\linewidth]{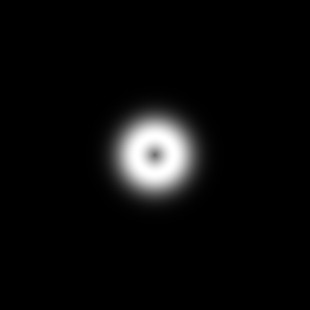}
    $R_8$
  \end{subfigure}\hfill
  \begin{subfigure}[b]{0.074\textwidth}\centering
    \includegraphics[width=\linewidth]{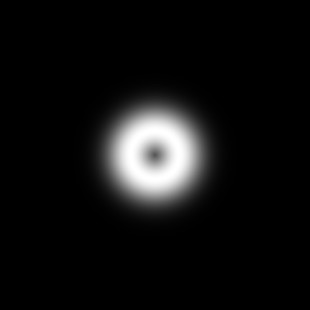}
    $R_9$
  \end{subfigure}\hfill
  \begin{subfigure}[b]{0.074\textwidth}\centering
    \includegraphics[width=\linewidth]{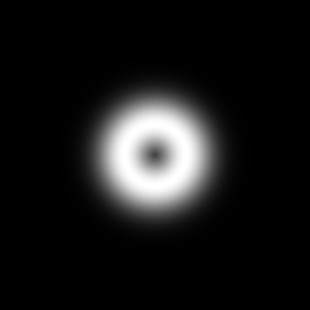}
    $R_{10}$
  \end{subfigure}\hfill
  \begin{subfigure}[b]{0.074\textwidth}\centering
    \includegraphics[width=\linewidth]{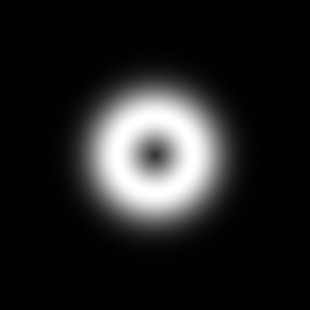}
    $R_{11}$
  \end{subfigure}\hfill
  \begin{subfigure}[b]{0.074\textwidth}\centering
    \includegraphics[width=\linewidth]{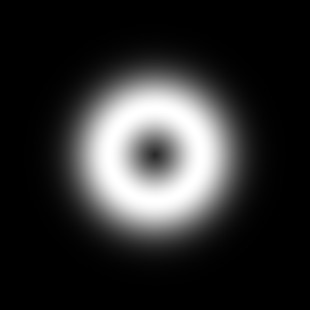}
    $R_{12}$
  \end{subfigure}\hfill
  \begin{subfigure}[b]{0.074\textwidth}\centering
    \includegraphics[width=\linewidth]{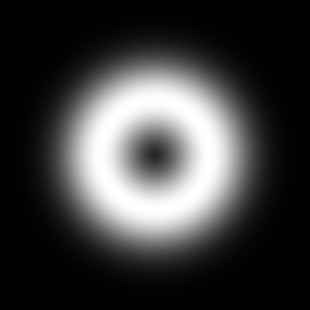}
    $R_{13}$
  \end{subfigure}
  \vfill
  \vspace{1mm}
  \begin{subfigure}[b]{0.074\textwidth}\centering
    \includegraphics[width=\linewidth]{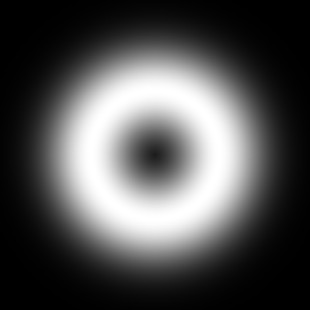}
    $R_{14}$
  \end{subfigure}\hfill
  \begin{subfigure}[b]{0.074\textwidth}\centering
    \includegraphics[width=\linewidth]{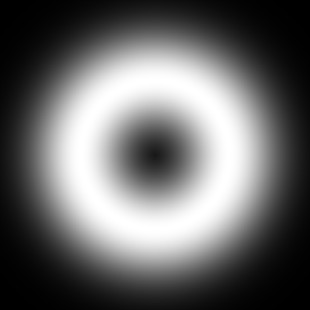}
    $R_{15}$
  \end{subfigure}\hfill
  \begin{subfigure}[b]{0.074\textwidth}\centering
    \includegraphics[width=\linewidth]{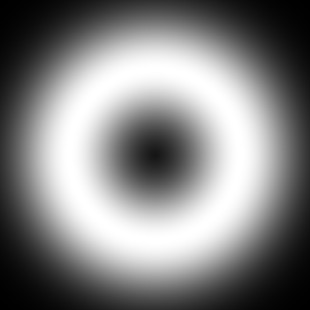}
    $R_{16}$
  \end{subfigure}\hfill
  \begin{subfigure}[b]{0.074\textwidth}\centering
    \includegraphics[width=\linewidth]{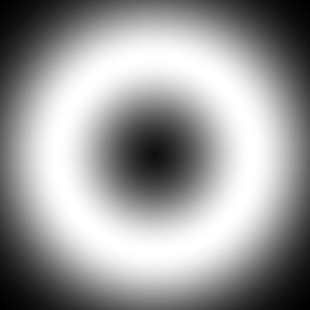}
    $R_{17}$
  \end{subfigure}\hfill
  \begin{subfigure}[b]{0.074\textwidth}\centering
    \includegraphics[width=\linewidth]{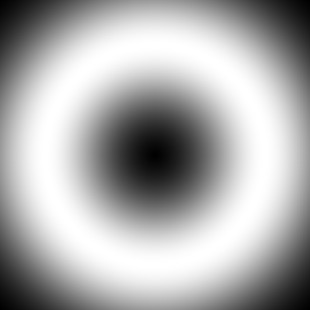}
    $R_{18}$
  \end{subfigure}\hfill
  \begin{subfigure}[b]{0.074\textwidth}\centering
    \includegraphics[width=\linewidth]{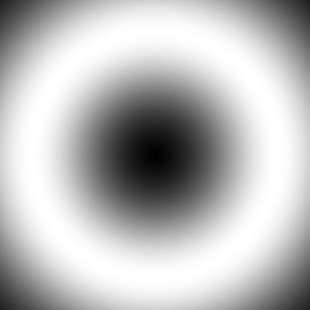}
    $R_{19}$
  \end{subfigure}\hfill
  \begin{subfigure}[b]{0.074\textwidth}\centering
    \includegraphics[width=\linewidth]{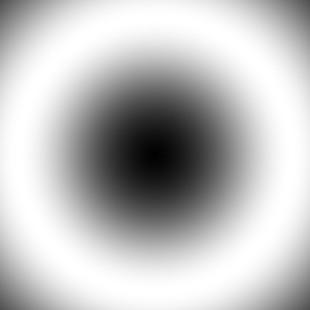}
    $R_{20}$
  \end{subfigure}\hfill
  \begin{subfigure}[b]{0.074\textwidth}\centering
    \includegraphics[width=\linewidth]{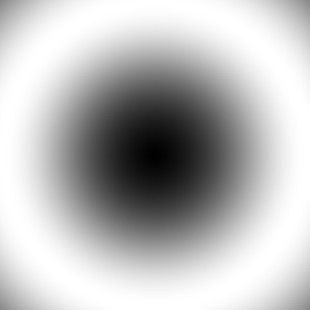}
    $R_{21}$
  \end{subfigure}\hfill
  \begin{subfigure}[b]{0.074\textwidth}\centering
    \includegraphics[width=\linewidth]{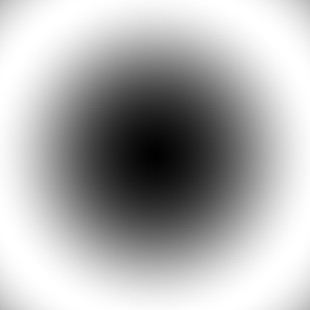}
    $R_{22}$
  \end{subfigure}\hfill
  \begin{subfigure}[b]{0.074\textwidth}\centering
    \includegraphics[width=\linewidth]{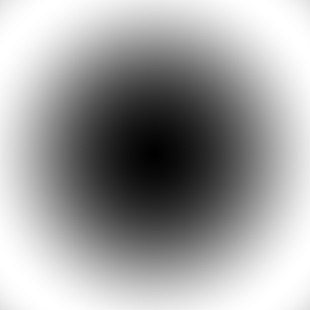}
    $R_{23}$
  \end{subfigure}\hfill
  \begin{subfigure}[b]{0.074\textwidth}\centering
    \includegraphics[width=\linewidth]{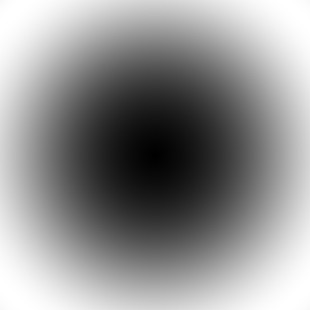}
    $R_{24}$
  \end{subfigure}\hfill
  \begin{subfigure}[b]{0.074\textwidth}\centering
    \includegraphics[width=\linewidth]{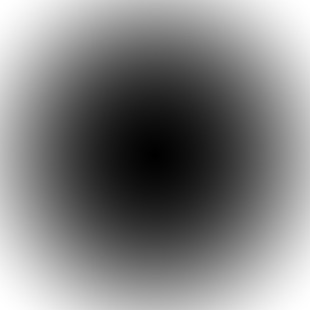}
    $R_{25}$
  \end{subfigure}\hfill
  \begin{subfigure}[b]{0.074\textwidth}\centering
    \hspace{0pt}
  \end{subfigure}
  \vfill

  \caption{Visualization of the predefined masks $R_1$-$R_{25}$.
    It demonstrates denser partitioning in the low-frequency domain and broader partitioning in the high-frequency domain.}
  \label{fig:mask_basis}
\end{figure*}
\vspace{-5mm}

\label{sec:basis_mask}
In our main paper, Equation (2) defines the $b$-th ring-shaped Gaussian basis mask $R_b$ using parameters $\mu_b$ and $\sigma_b$.
Here, $\mu_b$ represents the radial distance from the frequency-domain center where the mask has its peak, and $\sigma_b$ indicates the spread of the mask.
We construct a total of $B = 25$ Gaussian basis masks.
The first mask is centered at $\mu_1 = 0$ with a standard deviation of $\sigma_1 = 0.05$.
For $b = 1, \dots, B$, the peak positions $\mu_b$ are arranged by quadratically spacing values between \(0\) and $\sqrt{H^2 + W^2}/2$, where $H$ and $W$ are height and width of the image, yielding denser coverage near the DC component and sparser placement at higher frequencies.
Simultaneously, the spreads \(\sigma_b\) increase quadratically from \(0.05\) up to \(0.55\), providing narrower rings at low frequencies and broader ones at high frequencies.
This design ensures fine control around the low-frequency region and efficient coverage of the full frequency range.
The all predefined masks are visualized in Figure~\ref{fig:mask_basis}.

\subsection{Architecture details}
\begin{figure*}[h]
    \centering
    \vspace{2mm}
    \scalebox{0.7}{ 
        \includegraphics[width=\linewidth]{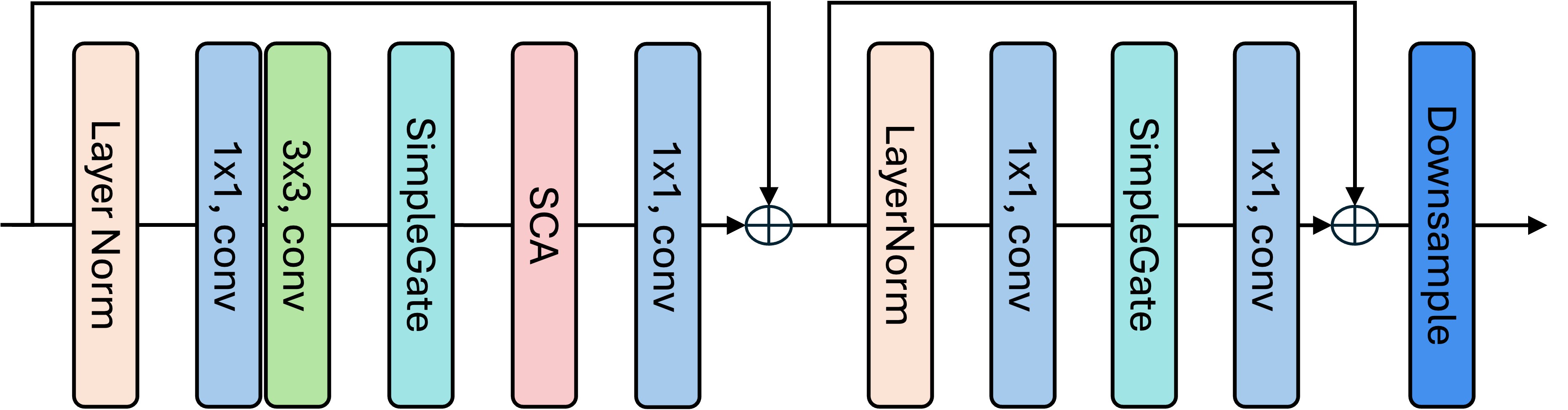}
    }
    \caption{The details of the NAFBlock.}    
    \label{fig:nafblock}
    \vspace{-2mm}
\end{figure*}

Figure~\ref{fig:nafblock} shows the details of the NAFBlock, utilized within the frequency mask generator illustrated in Figure~\ref{fig:overview} (b).
The foundational block structures, including the Simple Gate and Simplified Channel Attention (SCA), are adopted from the NAFNet architecture~\citep{chen2022simple}.
An additional Downsample operation, composed of a convolution with a kernel size of $2 \times 2$ and a stride of 2, is incorporated into this NAFBlock variant.
And the proposed output refinement network (ORNet) consists of the 4 encoder, 1 middle, and 4 decoder blocks, employing the NAFBlock as their fundamental building unit.
The encoder block is same with figure~\ref{fig:nafblock}.
The middle block does not incorporate last downsampling operation.
The decoder block implements an upsampling instead of downsampling: it first doubles the channel dimensionality using a $1 \times 1$ convolution, followed by a pixel shuffle module that doubles both the height and width of the feature maps, following the NAFNet architecture~\citep{chen2022simple}.

To control the perceptual enhancement intensity, we incorporate $\lambda$ as a conditional input for both the frequency mask generator and the output refinement network.
Specifically, we spatially expand the scalar $\lambda$ into a tensor of size $H \times W$ to match the input resolution and concatenate it with the input image along the channel dimension.
This conditioning allows the network to recognize the target enhancement level directly.

\section{Additional experiments}

\subsection{Visualization of Enhanced GT}

\begin{figure*}[t]
    \centering
    \begin{subfigure}[t]{0.49\textwidth}
        \includegraphics[width=\linewidth]{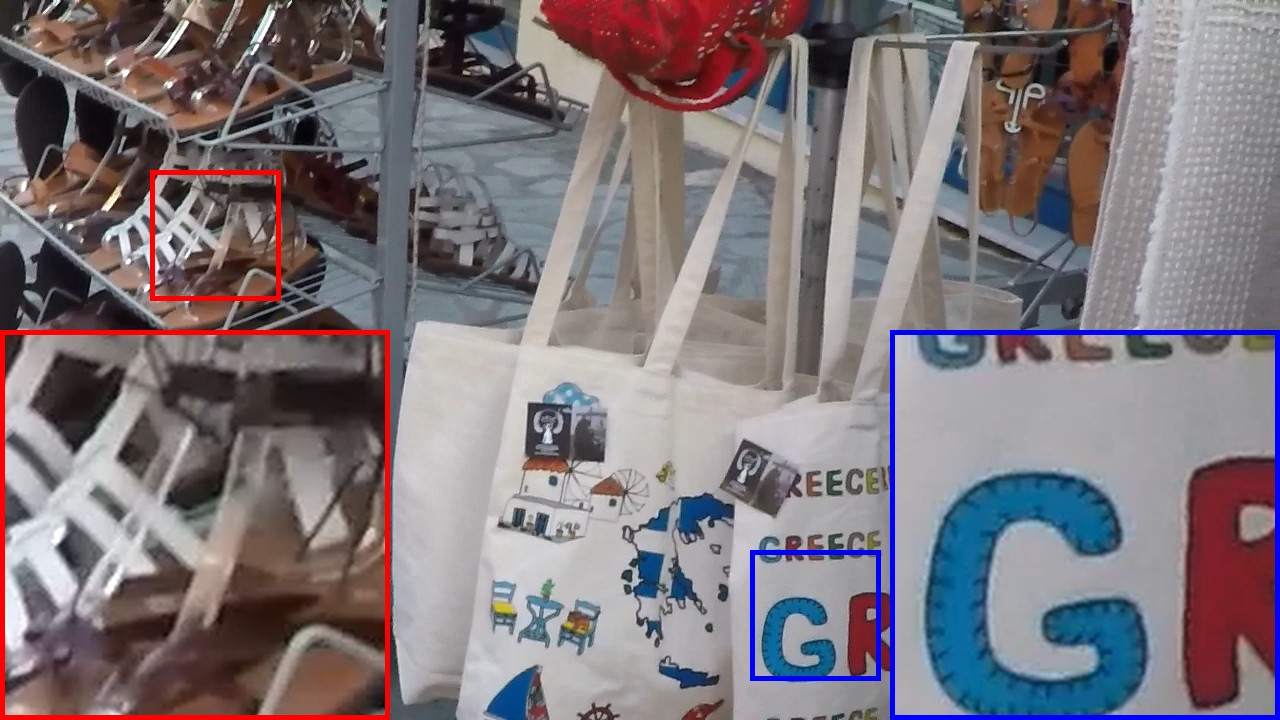}
        \centering GoPro GT
        \vspace{-3mm}
    \end{subfigure}
    \begin{subfigure}[t]{0.49\textwidth}
        \includegraphics[width=\linewidth]{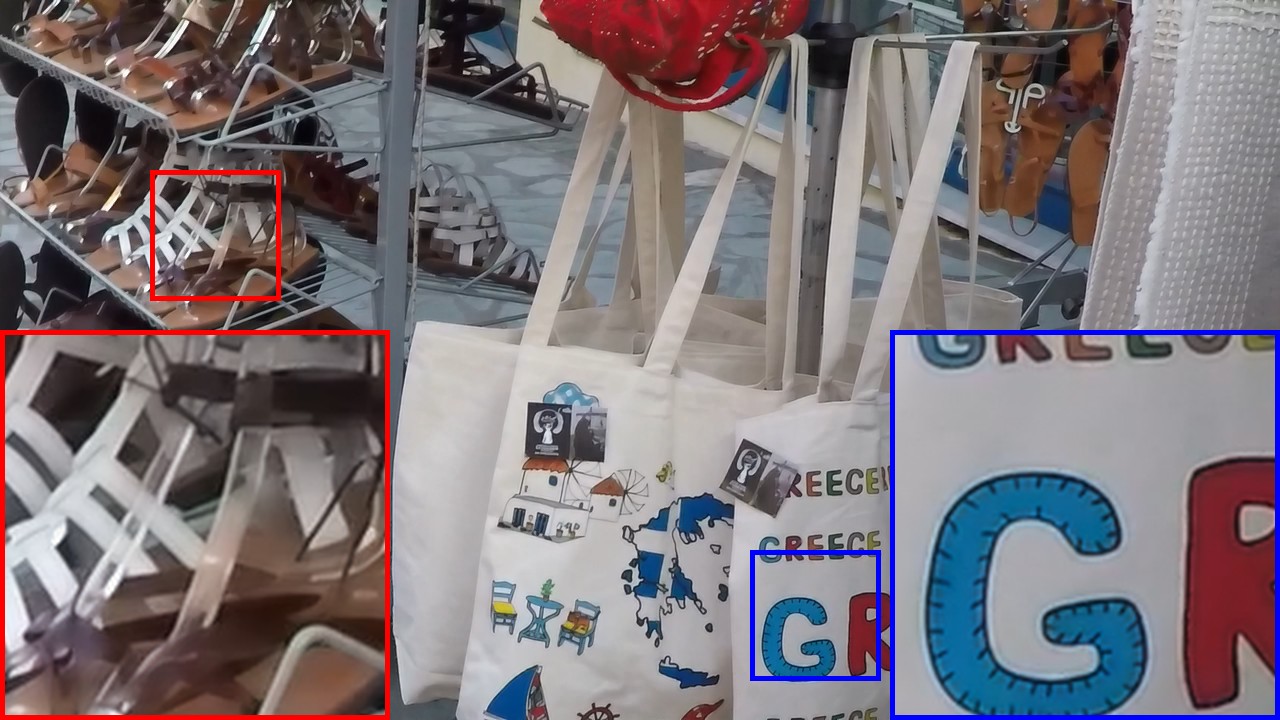}
        \centering Enhanced GT
        \vspace{-3mm}
    \end{subfigure}
    \\
    \begin{subfigure}[t]{0.49\textwidth}
        \includegraphics[width=\linewidth]{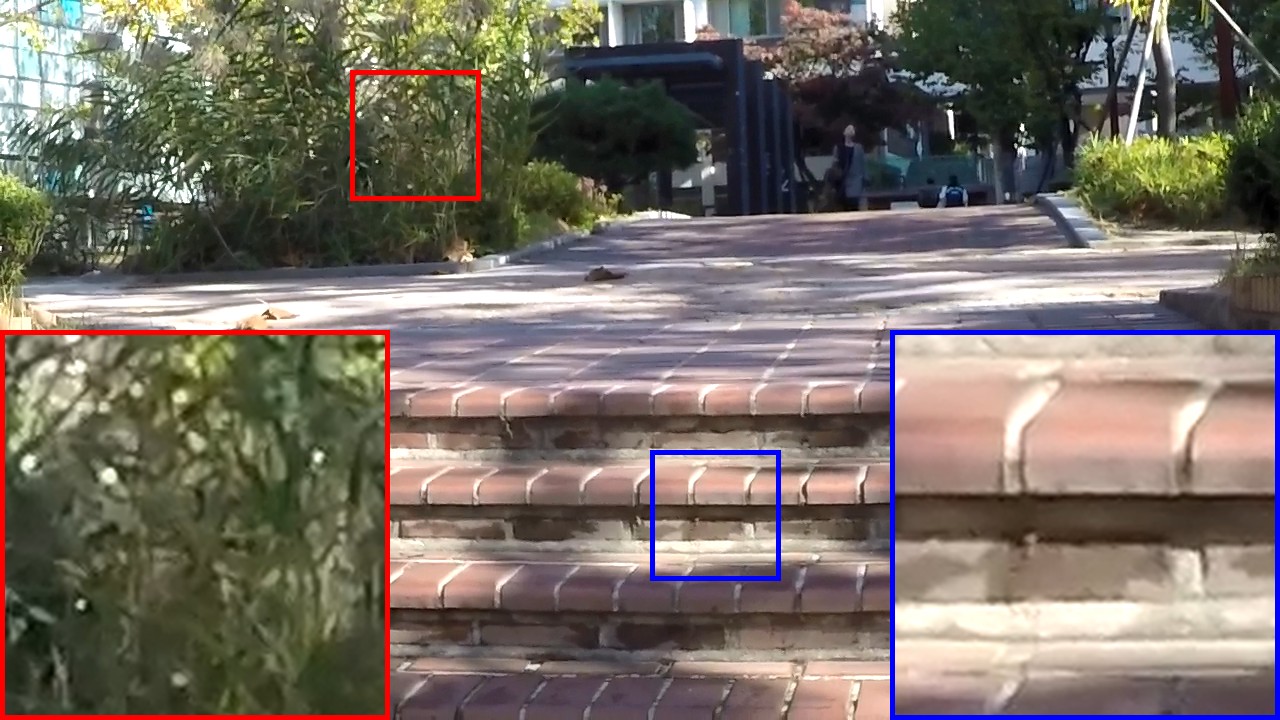}
        \centering GoPro GT
        \vspace{-3mm}
    \end{subfigure}
    \begin{subfigure}[t]{0.49\textwidth}
        \includegraphics[width=\linewidth]{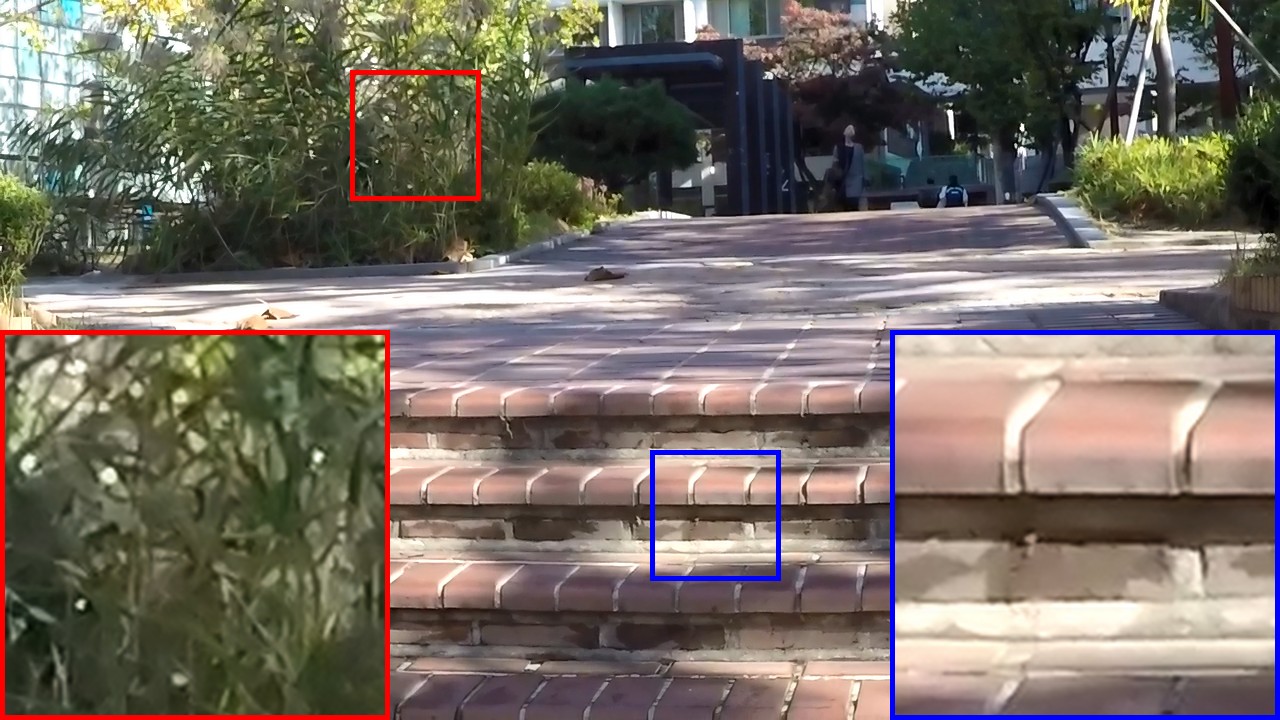}
        \centering Enhanced GT
        \vspace{-3mm}
    \end{subfigure}
    \\
    \hspace{0.03em}
    \begin{subfigure}[t]{0.49\textwidth}
        \includegraphics[width=\linewidth]{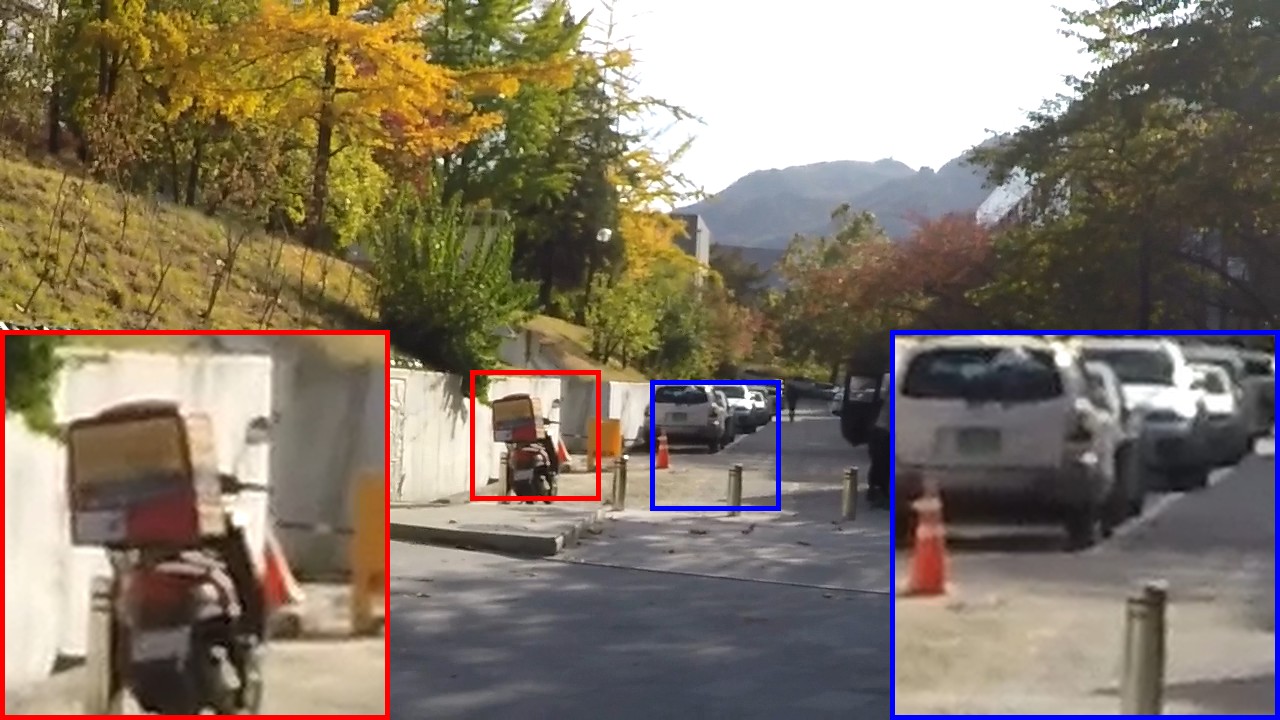}
        \centering GoPro GT
        \vspace{-3mm}
    \end{subfigure}
    \begin{subfigure}[t]{0.49\textwidth}
        \includegraphics[width=\linewidth]{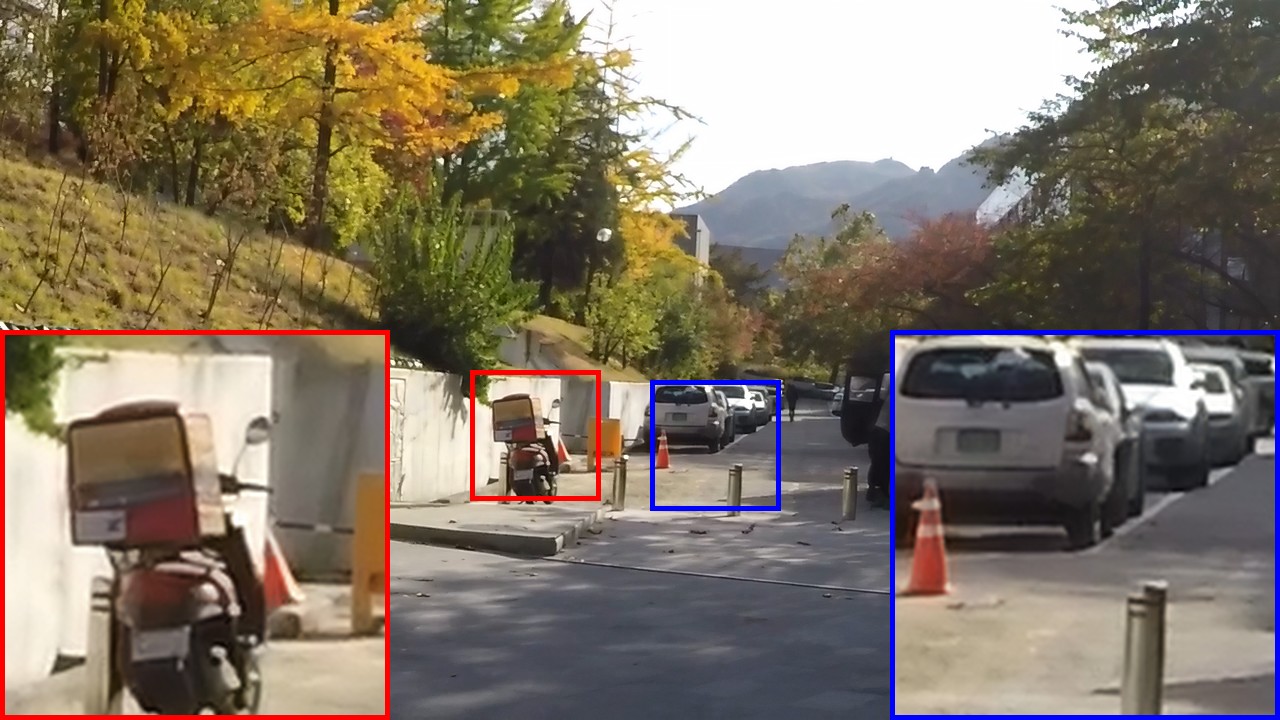}
        \centering Enhanced GT
        \vspace{-3mm}
    \end{subfigure}
       \vspace{1mm}
    \caption{
        Visualization of enhanced ground truth.
        Our enhanced GT not only exhibits sharper text and superior perceptual quality but also maintains semantic consistency.
        Zoom in for better visualization.
    }
    \label{fig:enhnaced_gt_sup}
    \vspace{-4mm}
\end{figure*}
\begin{figure*}[t]
    \captionsetup[subfigure]{labelformat=empty}
    \centering
    \begin{subfigure}[t]{0.33\linewidth}
        \centering
        \includegraphics[width=\linewidth]{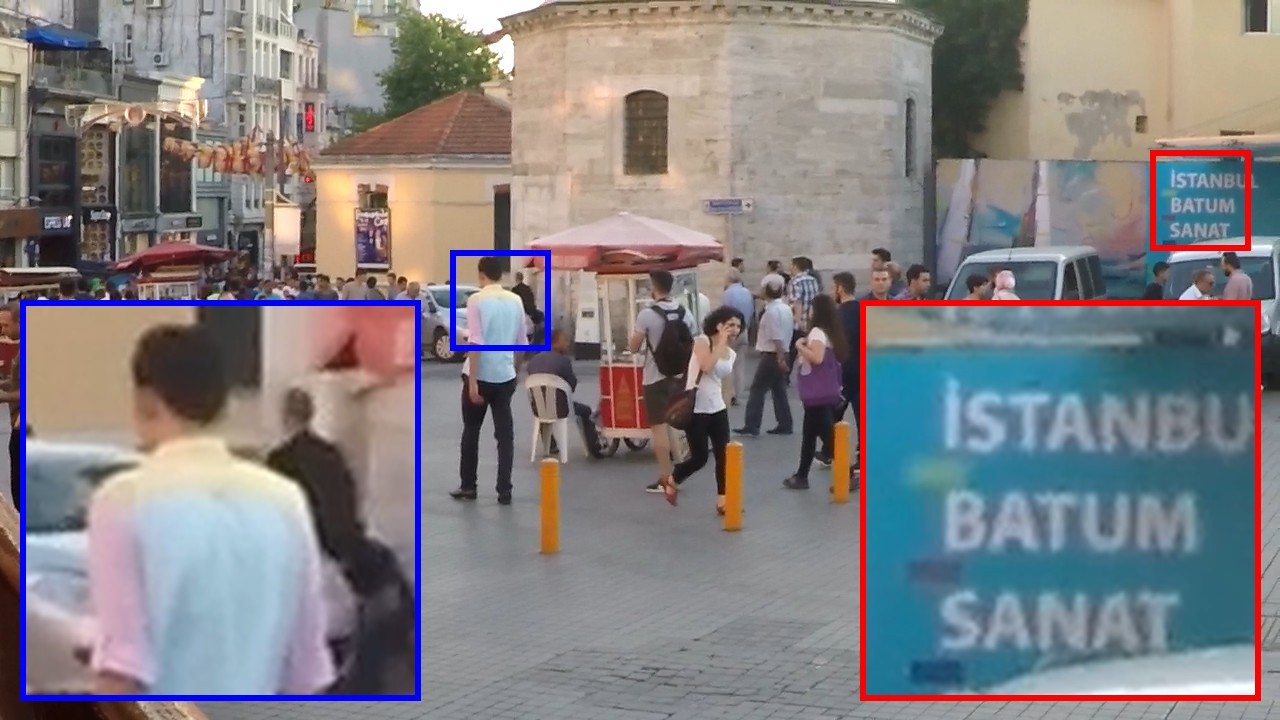}
        \caption{Original GT}
    \end{subfigure}\hfill
    \begin{subfigure}[t]{0.33\linewidth}
        \centering
        \includegraphics[width=\linewidth]{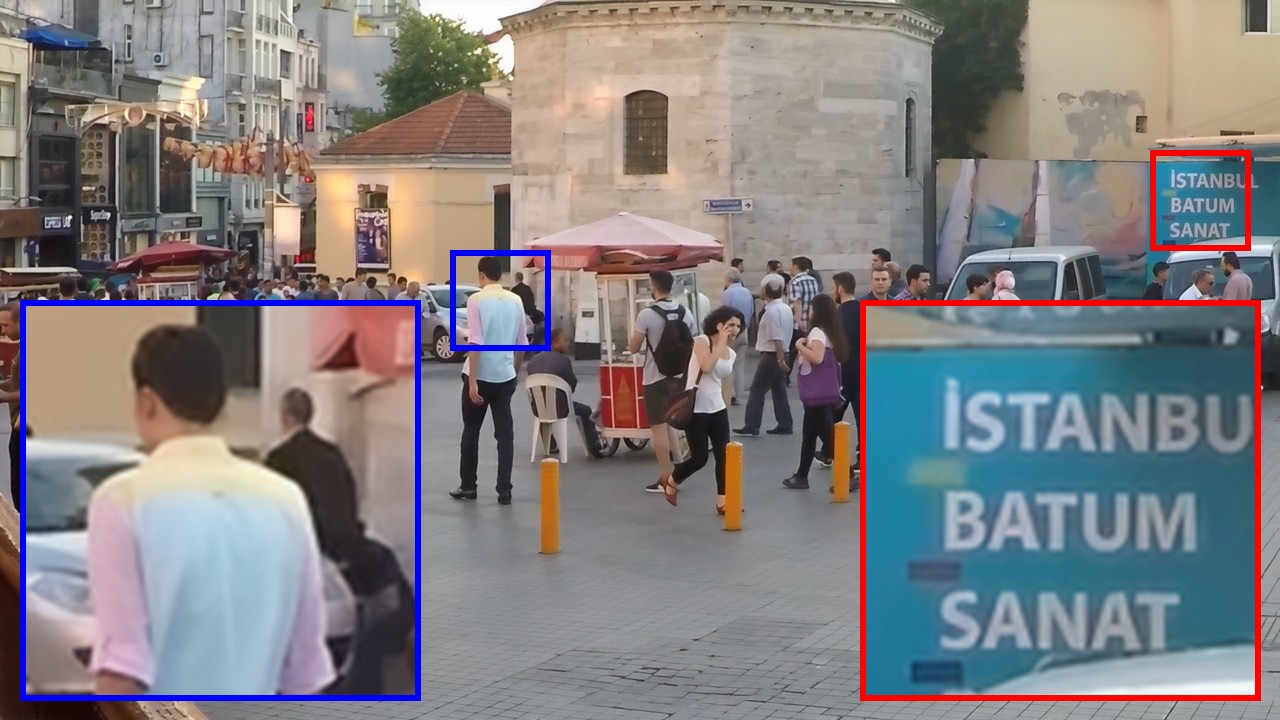}
        \caption{Enhanced GT (Ours)}
    \end{subfigure}\hfill
    \begin{subfigure}[t]{0.33\linewidth}
        \centering
        \includegraphics[width=\linewidth]{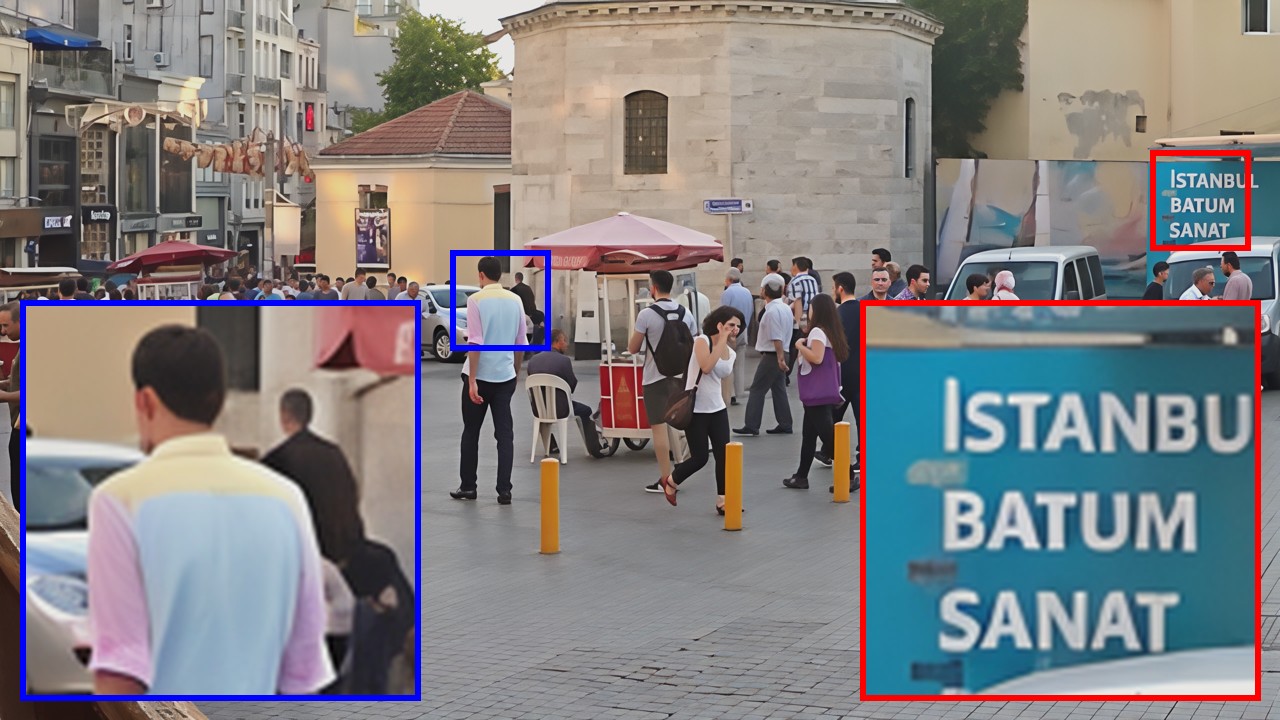}
        \caption{SR Variant ($I^\text{GT}_i$)}
    \end{subfigure}
    \vspace{-2mm}
    \caption{Comparison of supervision enhancement with SR variant. Zoom in for better visualization of the semantic details and color tone.}
    \label{fig:gt_comparison}
\end{figure*}
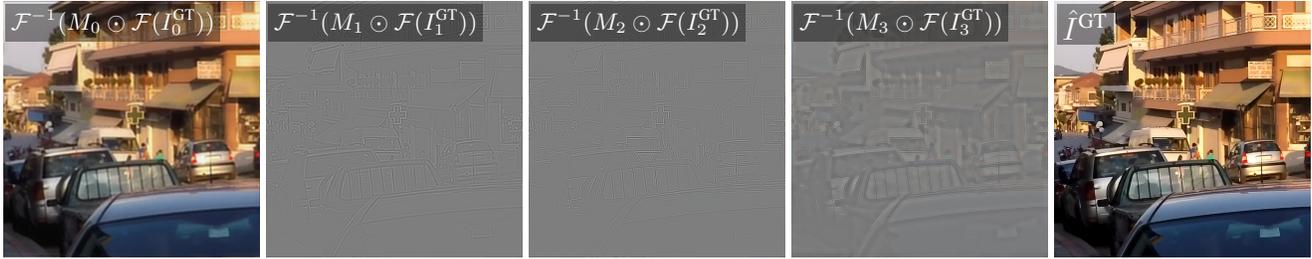
\begin{figure*}[t]
    \centering
    \scalebox{1.0}{%
        \begin{subfigure}[c]{0.195\textwidth}\centering
            \begin{tikzpicture}
                \node[anchor=south west,inner sep=0] (image) at (0,0) {
                    \includegraphics[width=\linewidth]{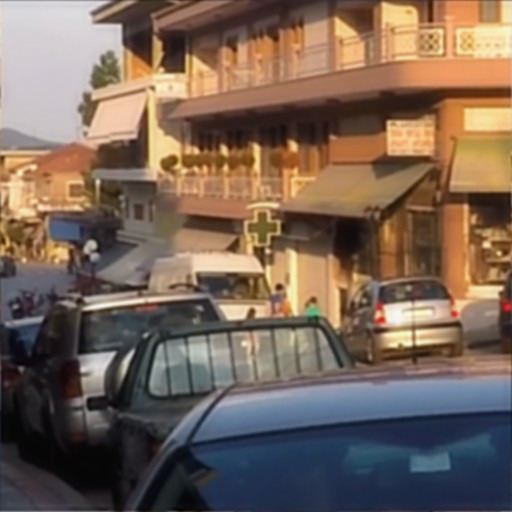}
                };
                \node[anchor=north west, fill=black, fill opacity=0.4, text=white, text opacity=1, inner sep=2pt, xshift=1pt, yshift=-1pt] at (image.north west) {\small $\mathcal{F}^{-1}( M_0 \odot \mathcal{F}(I_0^{\text{GT}}))$};
            \end{tikzpicture}
        \end{subfigure}
        \begin{subfigure}[c]{0.195\textwidth}\centering
            \begin{tikzpicture}
                \node[anchor=south west,inner sep=0] (image) at (0,0) {
                    \includegraphics[width=\linewidth]{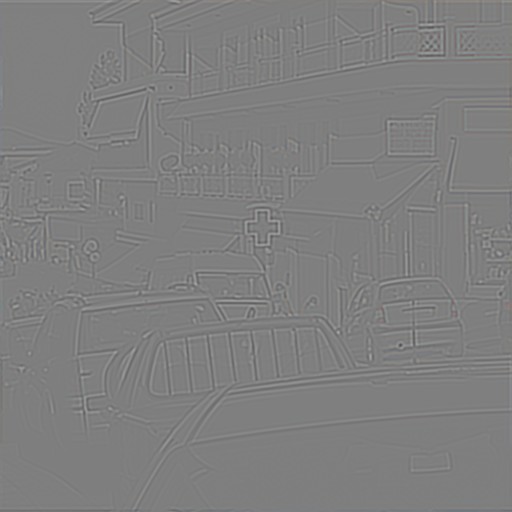}
                };
                \node[anchor=north west, fill=black, fill opacity=0.4, text=white, text opacity=1, inner sep=2pt, xshift=1pt, yshift=-1pt] at (image.north west) {\small $\mathcal{F}^{-1}( M_1 \odot \mathcal{F}(I_1^{\text{GT}}))$};
            \end{tikzpicture}
        \end{subfigure}
        \begin{subfigure}[c]{0.195\textwidth}\centering
            \begin{tikzpicture}
                \node[anchor=south west,inner sep=0] (image) at (0,0) {
                    \includegraphics[width=\linewidth]{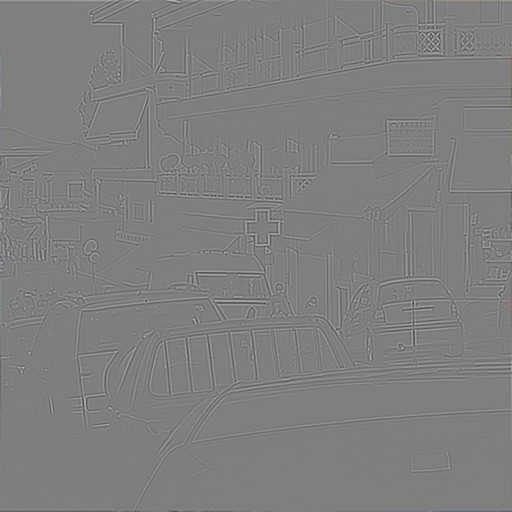}
                };
                \node[anchor=north west, fill=black, fill opacity=0.4, text=white, text opacity=1, inner sep=2pt, xshift=1pt, yshift=-1pt] at (image.north west) {\small $\mathcal{F}^{-1}( M_2 \odot \mathcal{F}(I_2^{\text{GT}}))$};
            \end{tikzpicture}
        \end{subfigure}
        \begin{subfigure}[c]{0.195\textwidth}\centering
            \begin{tikzpicture}
                \node[anchor=south west,inner sep=0] (image) at (0,0) {
                    \includegraphics[width=\linewidth]{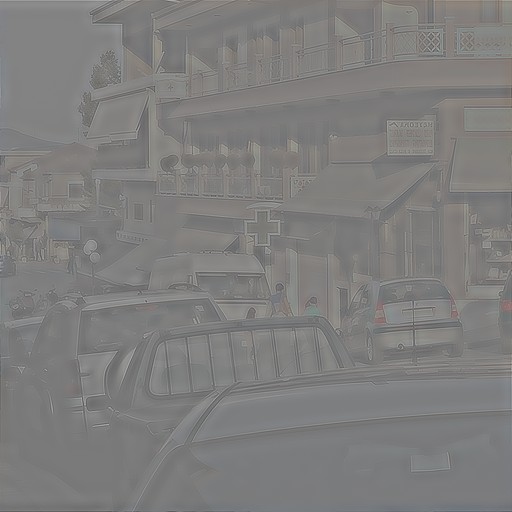}
                };
                \node[anchor=north west, fill=black, fill opacity=0.4, text=white, text opacity=1, inner sep=2pt, xshift=1pt, yshift=-1pt] at (image.north west) {\small $\mathcal{F}^{-1}( M_3 \odot \mathcal{F}(I_3^{\text{GT}}))$};

            \end{tikzpicture}
        \end{subfigure}
        \begin{subfigure}[c]{0.195\textwidth}\centering
            \begin{tikzpicture}
                \node[anchor=south west,inner sep=0] (image) at (0,0) {
                    \includegraphics[width=\linewidth]{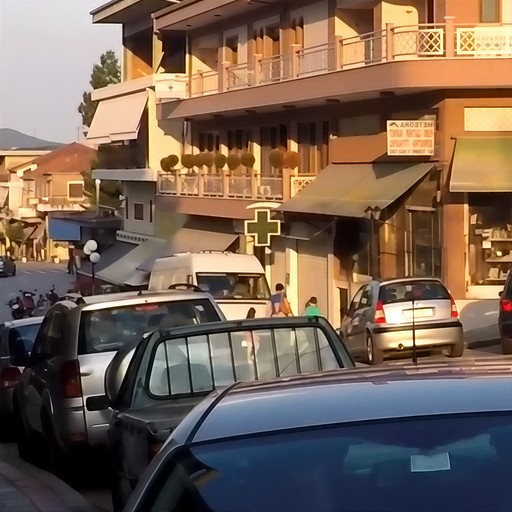}
                };
                \node[anchor=north west, fill=black, fill opacity=0.4, text=white, text opacity=1, inner sep=2pt, xshift=1pt, yshift=-1pt] at (image.north west) {\large $\hat{I}^{\text{GT}}$};
            \end{tikzpicture}
        \end{subfigure}
    }

    \vspace{-1mm}
    \caption{
    Visualization of composition of our enhanced ground truth image.
    Super-resolved variants and the original ground truth image are blended in the frequency domain using the generated masks.
    Each image visualized is the inverse Fourier transform of the masked frequency components of the corresponding original or super-resolved variant ground truth image.
    Zoom in for better visualization.
    }
    \label{fig:mixup_spatial_visual}
\end{figure*}

In Figure~\ref{fig:enhnaced_gt_sup}, we provide additional qualitative examples of our enhanced ground truth (GT) images.
Our method consistently shows superior perceptual quality by effectively removing residual degradations—such as noise and blur—from the original GTs, while preserving semantic fidelity.
In Figure~\ref{fig:gt_comparison}, we compare our supervision enhancement with a simple super-resolved variant.
While the super-resolved version mainly sharpens edges and increases overall brightness, our approach yields richer perceptual improvements without distorting semantic structure or altering the original color tone.
Lastly, Figure~\ref{fig:mixup_spatial_visual} illustrates how the enhanced GT is composed.
Our enhanced GT selectively integrates semantic details from the original GT and fine high-frequency components from the super-resolved variants, resulting in a sharper yet semantically consistent target image.

\subsection{Generalization on out-of-distribution dataset and unseen task}
\label{sec:generalization}
\begin{table*}[t]
    \centering
    \caption{
        The results are evaluated on the HIDE dataset~\citep{shen2019human} with the same settings as in the main paper.
    }
    \renewcommand{\arraystretch}{1.1}
    \scalebox{0.8}{
        \begin{tabular}{l  | cccc | cc | c}
            \toprule
            \rowcolor{gray!15}
                                                & \multicolumn{4}{c|}{\textit{No Ref.}} & \multicolumn{2}{c}{\textit{VLM-based.}} &                                                                                                                \\
            \rowcolor{gray!15}
            Method                              & MUSIQ$\uparrow$                       & MANIQA$\uparrow$                        & TOPIQ$\uparrow$ & LIQE$\uparrow$ & {VisualQuality-R1}$\uparrow$ & {Q-Insight}$\uparrow$ & {A-FINE}$\downarrow$ \\
            \midrule
            AdaRevD~\citep{mao2024adarevd}      & 55.91                                 & 0.5882                                  & 0.4064          & 2.0823         & 4.3013                       & 3.4613                & 33.18                \\
            {+ ORNet ($\lambda$=0.3)}           & 68.48                                 & 0.6282                                  & 0.5375          & 2.6209         & 4.3341                       & 3.5697                & 10.86                \\
            \midrule
            FFTFormer~\citep{kong2023efficient} & 54.42                                 & 0.5768                                  & 0.3978          & 2.0440         & 4.3176                       & 3.4733                & 36.35                \\
            {+ ORNet ($\lambda$=0.3)}           & 68.19                                 & 0.6256                                  & 0.5295          & 2.5832         & {4.3351}                     & 3.5691                & 13.57                \\
            \bottomrule
        \end{tabular}
    }
    \label{tab:hide_full}
\end{table*}

\begin{table*}[t]
    \centering
    \caption{
        The results are evaluated on the LOL dataset~\citep{wei2018deep} with the same settings as in the main paper.}
    \renewcommand{\arraystretch}{1.1}
    \scalebox{0.8}{
        \begin{tabular}{l  | cccc | cc | c}
            \toprule
            \rowcolor{gray!15}
                                                              & \multicolumn{4}{c|}{\textit{No Ref.}} & \multicolumn{2}{c|}{\textit{VLM-based.}} &                                                                                                                \\
            \rowcolor{gray!15}
            Method                                            & MUSIQ$\uparrow$                       & MANIQA$\uparrow$                         & TOPIQ$\uparrow$ & LIQE$\uparrow$ & {VisualQuality-R1}$\uparrow$ & {Q-Insight}$\uparrow$ & {A-FINE}$\downarrow$ \\
            \midrule
            \text{Retinexformer~\citep{cai2023retinexformer}} & 63.15                                 & 0.5870                                   & 0.5419          & 2.8354         & 3.4400                       & 3.3153                & 49.50                \\
            + ORNet ($\lambda$=0.3)                           & 72.80                                 & 0.6652                                   & 0.6435          & 3.9872         & 3.5200                       & 3.5047                & 26.96                \\

            \midrule
            \text{CIDNet~\citep{yan2025hvi}}                  & 69.51                                 & 0.6256                                   & 0.6288          & 3.8336         & 3.9133                       & 3.6967                & 46.52                \\
            {+ ORNet ($\lambda$=0.3)}                         & 74.88                                 & 0.7035                                   & 0.7108          & 4.7373         & 3.9000                       & 3.7613                & 20.05                \\
            \bottomrule
        \end{tabular}
    }
    \label{tab:lol_full}
\end{table*}

\paragraph{Quantitative results}
Furthermore, to evaluate generalization performance, we test our method on HIDE~\citep{shen2019human} as an out-of-distribution (OOD) deblurring benchmark.
We additionally evaluate it on LOL~\citep{wei2018deep}, a low-light enhancement benchmark, as an unseen restoration task.
For evaluating low light-enhancement, we adapt our refinement network to Retinexformer~\citep{cai2023retinexformer} and CIDNet~\citep{yan2025hvi}.
As shown in Table~\ref{tab:hide_full} and Table~\ref{tab:lol_full}, our approach consistently enhances perceptual quality even for the unknown dataset and task, demonstrating its strong generalization capabilities.

\begin{table}[t]
    \centering
    \caption{
        The results evaluated on the GyroBlur dataset.
    }
    \renewcommand{\arraystretch}{1.1}
    \scalebox{0.8}{
        \begin{tabular}{l | cc | cc}
            \toprule
            \rowcolor{gray!15}
                             & \multicolumn{2}{c|}{\textit{GyroBlur-Synth}}   & \multicolumn{2}{c}{\textit{GyroBlur-Real}} \\
            \rowcolor{gray!15}
            Method          & LPIPS$\downarrow$ & MUSIQ$\uparrow$          & MUSIQ$\uparrow$  & LIQE$\uparrow$  \\
            \midrule
            GyroDeblurNet     & 0.2408           & 41.58                    & 49.75          & 1.3407 \\
            \text{+ ORNet \textbf{(Ours)}}  & \textbf{0.2310}  & \textbf{56.88}           & \textbf{69.75} & \textbf{3.3835} \\
            \bottomrule
        \end{tabular}
    }
    \label{tab:gyroblur}
\end{table}
Additionally, we evaluate our method on the GyroBlur dataset~\citep{yang2025gyro}, which consists of both synthetic (GyroBlur-Synth) and real-world (GyroBlur-Real) motion blur sequences captured with gyroscope data.
As shown in Table~\ref{tab:gyroblur}, applying ORNet to GyroDeblurNet~\citep{yang2025gyro} yields significant perceptual improvements on both subsets, demonstrating that our framework generalizes effectively beyond standard deblurring benchmarks.

\paragraph{Qualitative results}
\begin{figure*}[t]
    \centering
    \captionsetup[subfigure]{labelformat=empty,justification=centering}
    \vspace{1mm}
    \begin{subfigure}[b]{0.24\textwidth}
        \includegraphics[width=\linewidth]{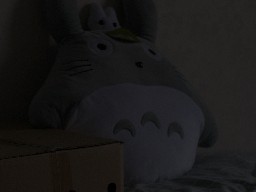}
        \vspace{-5mm}
    \end{subfigure}
    \hfill
    \begin{subfigure}[b]{0.24\textwidth}
        \includegraphics[width=\linewidth]{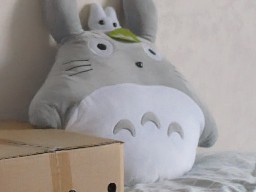}
        \vspace{-5mm}
    \end{subfigure}
    \hfill
    \begin{subfigure}[b]{0.24\textwidth}
        \includegraphics[width=\linewidth]{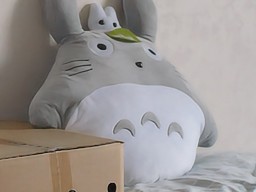}
        \vspace{-5mm}
    \end{subfigure}
    \hfill
    \begin{subfigure}[b]{0.24\textwidth}
        \includegraphics[width=\linewidth]{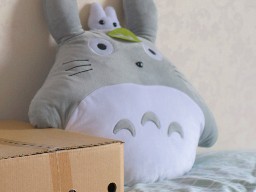}
        \vspace{-5mm}
    \end{subfigure}

    \vspace{2mm} 

    \begin{subfigure}[b]{0.24\textwidth}
        \includegraphics[width=\linewidth]{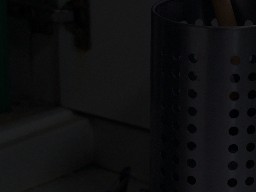}
        \vspace{-5mm}
        \caption{Low Light Input Image}
    \end{subfigure}
    \hfill
    \begin{subfigure}[b]{0.24\textwidth}
        \includegraphics[width=\linewidth]{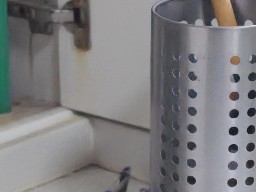}
        \vspace{-5mm}
        \caption{CIDNet~\citep{yan2025hvi}}
    \end{subfigure}
    \hfill
    \begin{subfigure}[b]{0.24\textwidth}
        \includegraphics[width=\linewidth]{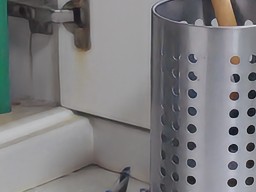}
        \vspace{-5mm}
        \caption{+ ORNet}
    \end{subfigure}
    \hfill
    \begin{subfigure}[b]{0.24\textwidth}
        \includegraphics[width=\linewidth]{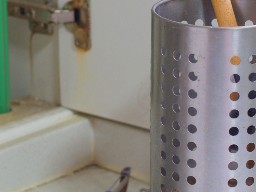}
        \vspace{-5mm}
        \caption{Ground Truth}
    \end{subfigure}
    \vspace{-1mm}
    \caption{Qualitative results of our ORNet when applied to the output of CIDNet~\citep{yan2025hvi} on the LOL low light enhancement dataset~\citep{wei2018deep}.
    Zoom in for better visualization.
    }
    \vspace{-2mm}
    \label{fig:lol}
\end{figure*}
Figure~\ref{fig:lol} presents the qualitative results of our ORNet ($\lambda$=0.3) when applied to the output of CIDNet~\citep{yan2025hvi} on the LOL low light enhancement dataset~\citep{wei2018deep}.
Our ORNet effectively enhances the overall quality of the output, resulting in a more visually appealing image.



\subsection{Comparison with using single GT variant}
\begin{table}[t]
    \center
    \caption{Comparison of our ORNet with only using supervision enhancement trained with only ground truth variant with upscale factor 4.}
    \renewcommand{\arraystretch}{1.1}
    \vspace{-1mm}
    \scalebox{0.85}{
    \begin{tabular}{l | cccc}
        \toprule
        \rowcolor{gray!15}
        Method & MUSIQ$\uparrow$ & MANIQA$\uparrow$ & TOPIQ$\uparrow$ & LIQE$\uparrow$ \\
        \midrule
        FFTformer~\citep{kong2023efficient} & 46.47 & 0.5420 & 0.3456 & 1.6131 \\
        \midrule
        + ORNet ($\lambda$=0.1)  & 48.07 & 0.5484 & 0.3537 & 1.6619 \\
        {+ ORNet ($\lambda$=0.3)} & 64.57 & 0.5949 & 0.4924 & 2.4664 \\
        + ORNet ($\lambda$=0.5)  & 69.18 & 0.6189 & 0.5905 & 2.9944 \\
        + ORNet ($\lambda$=0.7)  & 69.76 & 0.6352 & 0.6104 & 3.2444 \\
        + ORNet ($\lambda$=0.9)  & 69.76 & 0.6440 & 0.6198 & 3.3953 \\
        \midrule
        + ORNet\_4x ($\lambda$=0.1) & 47.86 & 0.5469 & 0.3534 & 1.6511 \\
        + ORNet\_4x ($\lambda$=0.3) & 62.81 & 0.5509 & 0.5069 & 2.1908 \\
        + ORNet\_4x ($\lambda$=0.5) & 67.53 & 0.5792 & 0.5738 & 2.6217 \\
        + ORNet\_4x ($\lambda$=0.7) & 68.15 & 0.5970 & 0.5775 & 2.7956 \\
        + ORNet\_4x ($\lambda$=0.9) & 68.20 & 0.6051 & 0.5773 & 2.8558 \\
        \bottomrule
    \end{tabular}
    }
    \label{tab:gt_variant}
    \vspace{-1mm}
\end{table}

To generate an enhanced ground truth, we first employ a super-resolution model to create ground truth variants.
For this purpose, we utilize upscale factors of 2, 3, and 4.
By applying a frequency mixup strategy to these diverse ground truth variants, we successfully construct an enhanced ground truth.
As an ablation study, we conduct an experiment with different ground truth variant setting, solely using upscale factor of 4.
Using these 4x variants, we follow our supervision enhancement framework and output refinement network.
The results are presented in Table~\ref{tab:gt_variant}.
As shown, ORNet\_4x, representing the model trained exclusively with the 4 upscale factor, achieved perceptual scores that are marginally lower than those obtained by the model trained with a mixture of variants.
This indicates that the incorporation of diverse GT variants is beneficial for achieving optimal final enhancement.

\subsection{Comparison with using ISR network instead of ORNet}
\begin{figure*}[t]
    \centering
    \captionsetup[subfigure]{labelformat=empty,justification=centering}

    \begin{subfigure}[b]{0.24\textwidth}
        \includegraphics[width=\linewidth]{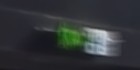}
        \vspace{-5mm}
        \caption{Blurry Input Image}
        \label{fig:sub1}
    \end{subfigure}
    \hfill
    \begin{subfigure}[b]{0.24\textwidth}
        \includegraphics[width=\linewidth]{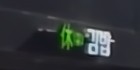}
        \vspace{-5mm}
        \caption{AdaRevD}
        \label{fig:sub2}
    \end{subfigure}
    \hfill
    \begin{subfigure}[b]{0.24\textwidth}
        \includegraphics[width=\linewidth]{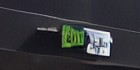}
        \vspace{-5mm}
        \caption{+ OSEDiff}
        \label{fig:sub3}
    \end{subfigure}
    \hfill
    \begin{subfigure}[b]{0.24\textwidth}
        \includegraphics[width=\linewidth]{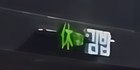}
        \vspace{-5mm}
        \caption{+ ORNet}
        \label{fig:sub4}
    \end{subfigure}
    \vspace{-1.5mm}
    \caption{
    Comparison of applying the ISR model (OSEDiff) directly and using our ORNet to refine the output of the restoration model (AdaRevD).
    The ISR model generates details that are not present in the input image, leading to unrealistic results.
    }
    \vspace{-2mm}
    \label{fig:direct_isr}
\end{figure*}

Our output refinement network (ORNet) is trained to refine the output of any existing image restoration model, which is trained with enhanced supervision.
To enhance the output of the restoration model, we could also directly apply the diffusion based image super-resolution (ISR) model used in our framework.
However, ISR model, which is designed to enhance the perceptual quality, often hallucinates details that are not present in the input image when applied directly.
Figure~\ref{fig:direct_isr} shows the results of applying the ISR model directly to the output of the restoration model, where the refined output destroys the textual detail.

\subsection{Ablation study on number of Gaussian ring-shaped basis}

\begin{figure*}[t]
    \centering
    \begin{subfigure}[c]{0.04\textwidth}\centering
        \rotatebox{90}{\parbox{3cm}{\centering B=5}}
    \end{subfigure}\hspace{2mm}%
    \scalebox{.88}{%
        \begin{subfigure}[c]{0.19\textwidth}\centering
            \begin{tikzpicture}
                \node[anchor=south west,inner sep=0] (image) at (0,0) {
                    \includegraphics[width=\linewidth]{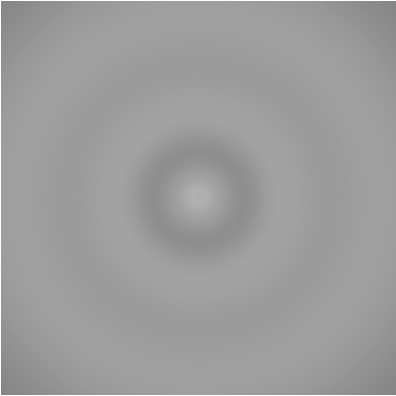}
                };
                \node[anchor=north west, fill=black, fill opacity=0.4, text=white, text opacity=1, inner sep=2pt, xshift=1pt, yshift=-1pt] at (image.north west) {$M_0$};
            \end{tikzpicture}
        \end{subfigure}
        \begin{subfigure}[c]{0.19\textwidth}\centering
            \begin{tikzpicture}
                \node[anchor=south west,inner sep=0] (image) at (0,0) {
                    \includegraphics[width=\linewidth]{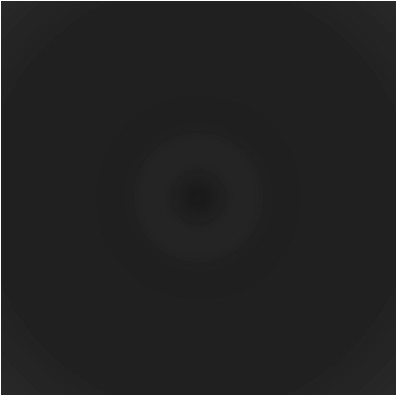}
                };
                \node[anchor=north west, fill=black, fill opacity=0.4, text=white, text opacity=1, inner sep=2pt, xshift=1pt, yshift=-1pt] at (image.north west) {$M_1$};
            \end{tikzpicture}
        \end{subfigure}
        \begin{subfigure}[c]{0.19\textwidth}\centering
            \begin{tikzpicture}
                \node[anchor=south west,inner sep=0] (image) at (0,0) {
                    \includegraphics[width=\linewidth]{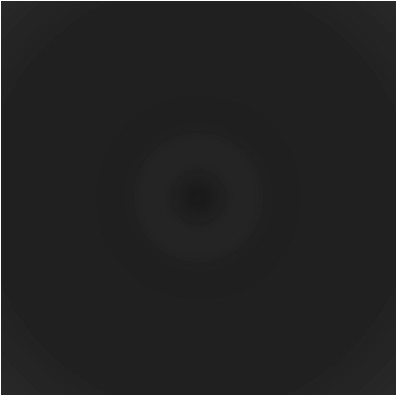}
                };
                \node[anchor=north west, fill=black, fill opacity=0.4, text=white, text opacity=1, inner sep=2pt, xshift=1pt, yshift=-1pt] at (image.north west) {$M_2$};
            \end{tikzpicture}
        \end{subfigure}
        \begin{subfigure}[c]{0.19\textwidth}\centering
            \begin{tikzpicture}
                \node[anchor=south west,inner sep=0] (image) at (0,0) {
                    \includegraphics[width=\linewidth]{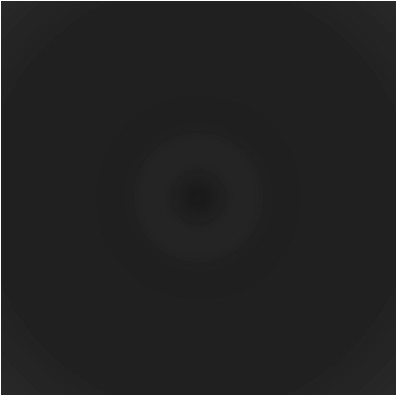}
                };
                \node[anchor=north west, fill=black, fill opacity=0.4, text=white, text opacity=1, inner sep=2pt, xshift=1pt, yshift=-1pt] at (image.north west) {$M_3$};
            \end{tikzpicture}
        \end{subfigure}
        \begin{subfigure}[c]{0.19\textwidth}\centering
            \begin{tikzpicture}
                \node[anchor=south west,inner sep=0] (image) at (0,0) {
                    \includegraphics[width=\linewidth]{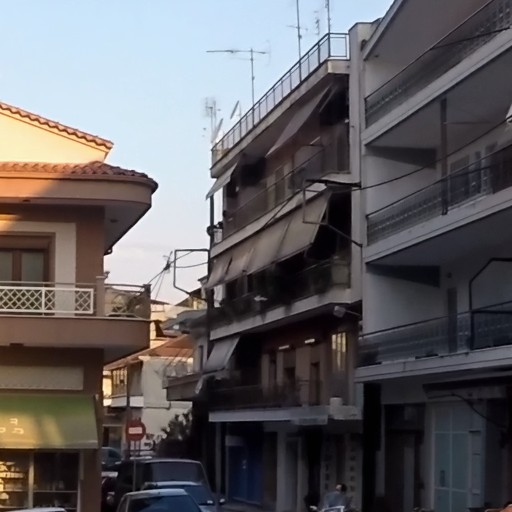}
                };
                \node[anchor=north west, fill=black, fill opacity=0.4, text=white, text opacity=1, inner sep=2pt, xshift=1pt, yshift=-1pt] at (image.north west) {$\hat{I}^{\text{GT}}$};
            \end{tikzpicture}
        \end{subfigure}
    }\\[0pt]
    \vspace{-1mm}
    \begin{subfigure}[c]{0.04\textwidth}\centering
        \rotatebox{90}{\parbox{3cm}{\centering B=25}}
    \end{subfigure}\hspace{2mm}%
    \scalebox{.88}{%
        \begin{subfigure}[c]{0.19\textwidth}\centering
            \begin{tikzpicture}
                \node[anchor=south west,inner sep=0] (image) at (0,0) {
                    \includegraphics[width=\linewidth]{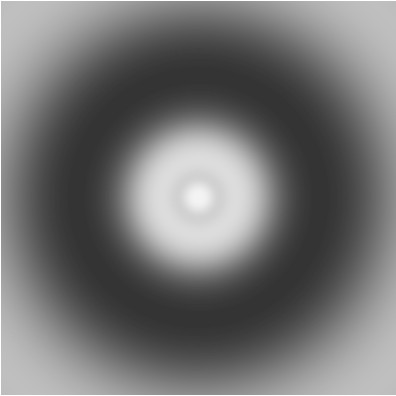}
                };
                \node[anchor=north west, fill=black, fill opacity=0.4, text=white, text opacity=1, inner sep=2pt, xshift=1pt, yshift=-1pt] at (image.north west) {$M_0$};
            \end{tikzpicture}
        \end{subfigure}
        \begin{subfigure}[c]{0.19\textwidth}\centering
            \begin{tikzpicture}
                \node[anchor=south west,inner sep=0] (image) at (0,0) {
                    \includegraphics[width=\linewidth]{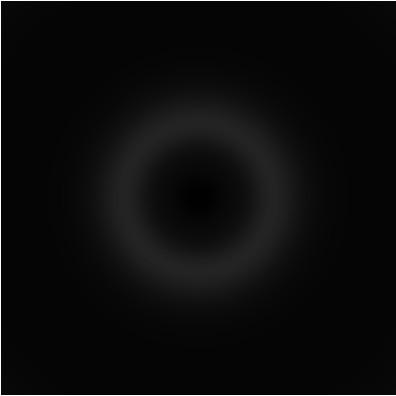}
                };
                \node[anchor=north west, fill=black, fill opacity=0.4, text=white, text opacity=1, inner sep=2pt, xshift=1pt, yshift=-1pt] at (image.north west) {$M_1$};
            \end{tikzpicture}
        \end{subfigure}
        \begin{subfigure}[c]{0.19\textwidth}\centering
            \begin{tikzpicture}
                \node[anchor=south west,inner sep=0] (image) at (0,0) {
                    \includegraphics[width=\linewidth]{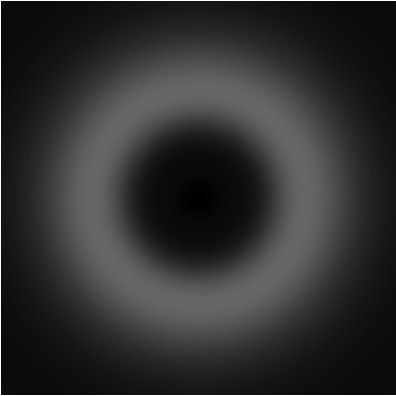}
                };
                \node[anchor=north west, fill=black, fill opacity=0.4, text=white, text opacity=1, inner sep=2pt, xshift=1pt, yshift=-1pt] at (image.north west) {$M_2$};
            \end{tikzpicture}
        \end{subfigure}
        \begin{subfigure}[c]{0.19\textwidth}\centering
            \begin{tikzpicture}
                \node[anchor=south west,inner sep=0] (image) at (0,0) {
                    \includegraphics[width=\linewidth]{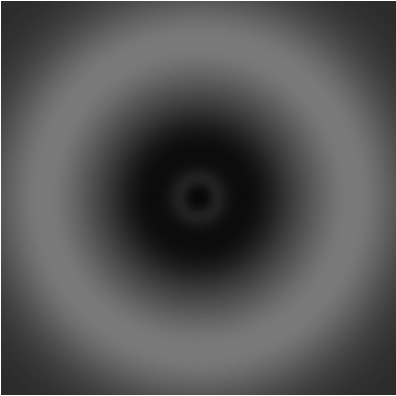}
                };
                \node[anchor=north west, fill=black, fill opacity=0.4, text=white, text opacity=1, inner sep=2pt, xshift=1pt, yshift=-1pt] at (image.north west) {$M_3$};
            \end{tikzpicture}
        \end{subfigure}
        \begin{subfigure}[c]{0.19\textwidth}\centering
            \begin{tikzpicture}
                \node[anchor=south west,inner sep=0] (image) at (0,0) {
                    \includegraphics[width=\linewidth]{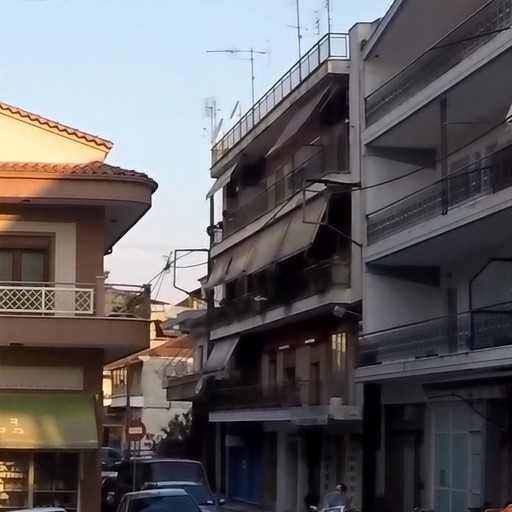}
                };
                \node[anchor=north west, fill=black, fill opacity=0.4, text=white, text opacity=1, inner sep=2pt, xshift=1pt, yshift=-1pt] at (image.north west) {$\hat{I}^{\text{GT}}$};
            \end{tikzpicture}
        \end{subfigure}
    }\\[0pt]
    \begin{subfigure}[c]{0.04\textwidth}\centering
        \rotatebox{90}{\parbox{3cm}{\centering B=100}}
    \end{subfigure}\hspace{2mm}%
    \scalebox{.88}{%
        \begin{subfigure}[c]{0.19\textwidth}\centering
            \begin{tikzpicture}
                \node[anchor=south west,inner sep=0] (image) at (0,0) {
                    \includegraphics[width=\linewidth]{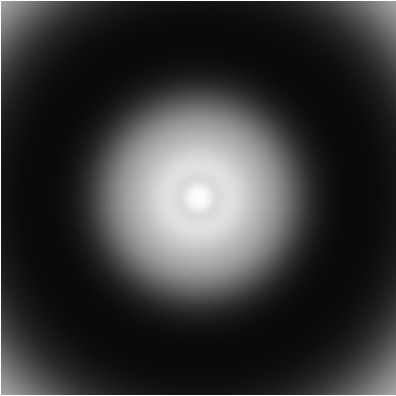}
                };
                \node[anchor=north west, fill=black, fill opacity=0.4, text=white, text opacity=1, inner sep=2pt, xshift=1pt, yshift=-1pt] at (image.north west) {$M_0$};
            \end{tikzpicture}
        \end{subfigure}
        \begin{subfigure}[c]{0.19\textwidth}\centering
            \begin{tikzpicture}
                \node[anchor=south west,inner sep=0] (image) at (0,0) {
                    \includegraphics[width=\linewidth]{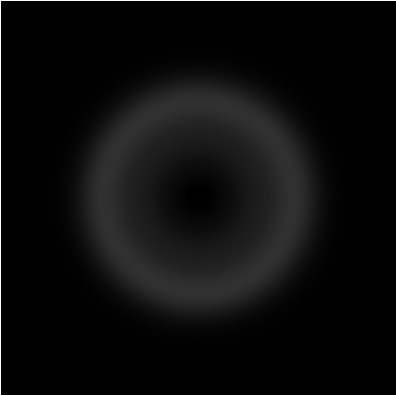}
                };
                \node[anchor=north west, fill=black, fill opacity=0.4, text=white, text opacity=1, inner sep=2pt, xshift=1pt, yshift=-1pt] at (image.north west) {$M_1$};
            \end{tikzpicture}
        \end{subfigure}
        \begin{subfigure}[c]{0.19\textwidth}\centering
            \begin{tikzpicture}
                \node[anchor=south west,inner sep=0] (image) at (0,0) {
                    \includegraphics[width=\linewidth]{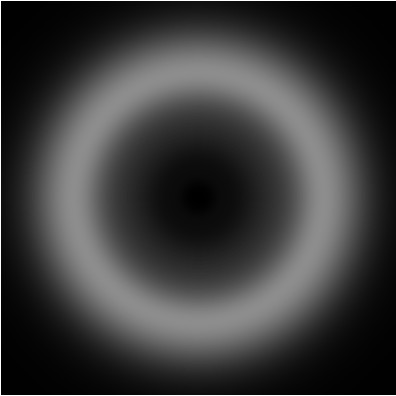}
                };
                \node[anchor=north west, fill=black, fill opacity=0.4, text=white, text opacity=1, inner sep=2pt, xshift=1pt, yshift=-1pt] at (image.north west) {$M_2$};
            \end{tikzpicture}
        \end{subfigure}
        \begin{subfigure}[c]{0.19\textwidth}\centering
            \begin{tikzpicture}
                \node[anchor=south west,inner sep=0] (image) at (0,0) {
                    \includegraphics[width=\linewidth]{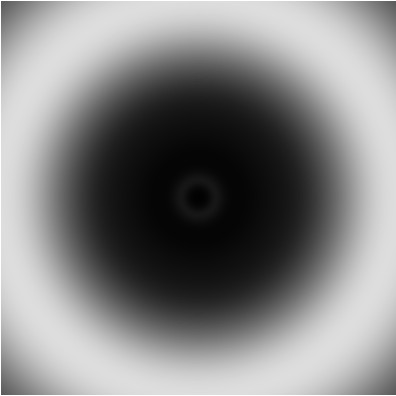}
                };
                \node[anchor=north west, fill=black, fill opacity=0.4, text=white, text opacity=1, inner sep=2pt, xshift=1pt, yshift=-1pt] at (image.north west) {$M_3$};
            \end{tikzpicture}
        \end{subfigure}
        \begin{subfigure}[c]{0.19\textwidth}\centering
            \begin{tikzpicture}
                \node[anchor=south west,inner sep=0] (image) at (0,0) {
                    \includegraphics[width=\linewidth]{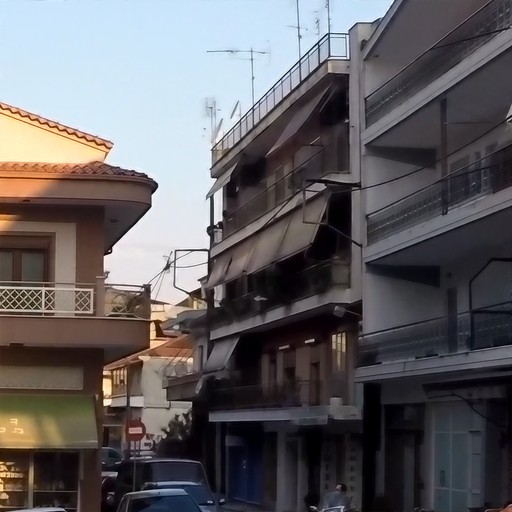}
                };
                \node[anchor=north west, fill=black, fill opacity=0.4, text=white, text opacity=1, inner sep=2pt, xshift=1pt, yshift=-1pt] at (image.north west) {$\hat{I}^{\text{GT}}$};
            \end{tikzpicture}
        \end{subfigure}
    }

    \vspace{-1mm}
    \caption{
    Visualization of generated masks and enhanced results with different numbers of frequency bands ($B$).
    The rows correspond to $B=5$, $B=25$ (Ours), and $B=100$, from top to bottom.
    $M_i$ denotes the generated masks, and $\hat{I}^{\text{GT}}$ represents the enhanced ground truth generated using these mask.
    }
    \vspace{-4mm}
    \label{fig:mask_analysis}
\end{figure*}
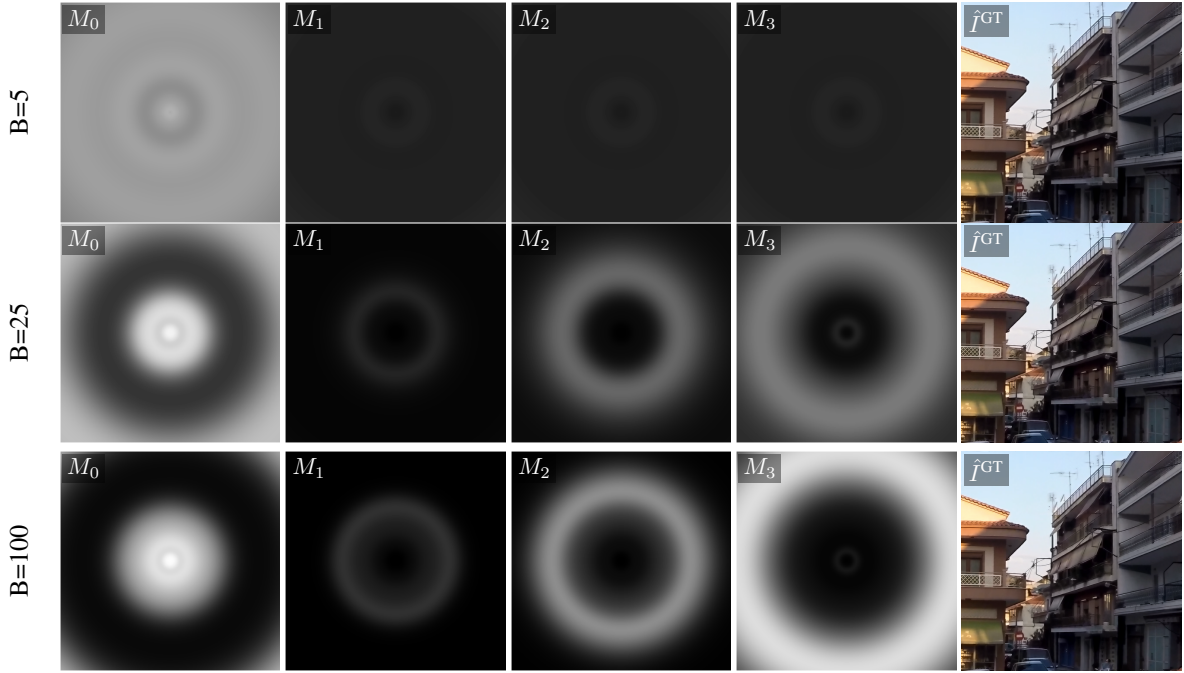

We conduct an ablation study on the number of predefined Gaussian ring-shaped bases, denoted as $B$, to analyze its impact on frequency control.
As shown in Fig.~\ref{fig:mask_analysis}, we compare the generated masks and results with $B \in \{5, 25, 100\}$.
When $B$ is too small (e.g., $B=5$), we observe that the model fails to achieve fine-grained control over the frequency components due to the limited resolution of the basis functions.
However, once $B$ exceeds a certain threshold, the model exhibits a consistent tendency in frequency selectivity.
Since we observed that $B=25$ already provides sufficient capacity for precise frequency control, we adopt $B=25$ as the default setting in our experiments.

\subsection{Ablation study on enhancement level $\lambda$}

\begin{table*}[!t]
    \centering
    \caption{
        Full table with diverse enhancement levels.
        The results are evaluated on the GoPro deblurring~\citep{nah2017deep} and SIDD denoising~\citep{abdelhamed2018high} test sets.
    }
    \renewcommand{\arraystretch}{0.9}
    \setlength{\tabcolsep}{15pt}
    \vspace{-1mm}
    \scalebox{0.75}{
        \begin{tabular}{l l | cccc | cc| c}
            \toprule
            \rowcolor{gray!15}
            {}                                                                          &                                            & \multicolumn{4}{c|}{\textit{Perceptual Quality Metrics}} & \multicolumn{2}{c}{\textit{VLM-based Metrics}} &                                                                                                                    \\
            \rowcolor{gray!15}
            {}                                                                          & {Method}                                   & {MUSIQ}$\uparrow$                                        & {MANIQA}$\uparrow$                             & {TOPIQ}$\uparrow$ & {LIQE}$\uparrow$ & {VisualQuality-R1}$\uparrow$ & {Q-Insight}$\uparrow$ & {A-FINE}$\downarrow$ \\
            \midrule
            \multirow{12}{*}{\rotatebox[origin=c]{90}{\large\textit{GoPro Deblurring}}} & \text{AdaRevD~\citep{mao2024adarevd}}      & 45.49                                                    & 0.5363                                         & 0.3393            & 1.5656           & 4.1737                       & 3.4386                & 36.19                \\
                                                                                        & \text{+ ORNet ($\lambda$=0.1)}             & 47.42                                                    & 0.5451                                         & 0.3490            & 1.6256           & 4.1939                       & 3.4605                & 33.33                \\
                                                                                        & \text{+ ORNet ($\lambda$=0.3)}             & 64.25                                                    & 0.5916                                         & 0.4880            & 2.4291           & 4.1952                       & 3.5206                & {14.38}              \\
                                                                                        & \text{+ ORNet ($\lambda$=0.5)}             & 69.11                                                    & 0.6161                                         & 0.5893            & 2.9711           & 4.0686                       & 3.5163                & 6.80                 \\
                                                                                        & \text{+ ORNet ($\lambda$=0.7)}             & 69.72                                                    & 0.6330                                         & 0.6098            & 3.2308           & 4.0025                       & 3.5063                & 4.77                 \\
                                                                                        & \text{+ ORNet ($\lambda$=0.9)}             & 69.73                                                    & 0.6420                                         & 0.6195            & 3.3870           & 3.9146                       & 3.4838                & 4.24                 \\
            \cmidrule{2-9}
                                                                                        & \text{FFTformer~\citep{kong2023efficient}} & 46.47                                                    & 0.5420                                         & 0.3456            & 1.6131           & 4.0942                       & 3.4569                & 36.91                \\
                                                                                        & \text{+ ORNet ($\lambda$=0.1)}             & 48.07                                                    & 0.5484                                         & 0.3537            & 1.6619           & 4.2334                       & 3.4764                & 34.24                \\
                                                                                        & \text{+ ORNet ($\lambda$=0.3)}             & 64.57                                                    & 0.5949                                         & 0.4924            & 2.4664           & 4.1995                       & 3.5278                & {15.69}              \\
                                                                                        & \text{+ ORNet ($\lambda$=0.5)}             & 69.18                                                    & 0.6189                                         & 0.5905            & 2.9944           & 4.0918                       & 3.5234                & 7.68                 \\
                                                                                        & \text{+ ORNet ($\lambda$=0.7)}             & 69.76                                                    & 0.6352                                         & 0.6104            & 3.2444           & 4.0262                       & 3.5162                & 5.49                 \\
                                                                                        & \text{+ ORNet ($\lambda$=0.9)}             & 69.76                                                    & 0.6440                                         & 0.6198            & 3.3953           & 3.9600                       & 3.4957                & 4.91                 \\
            \midrule
            \multirow{12}{*}{\rotatebox[origin=c]{90}{\large\textit{SIDD Denoising}}}   & \text{Xformer~\citep{zhang2023xformer}}    & 22.57                                                    & 0.3828                                         & 0.2472            & 1.2040           & 1.0759                       & 1.6710                & 52.33                \\
                                                                                        & \text{+ ORNet ($\lambda$=0.1)}             & 26.23                                                    & 0.3819                                         & 0.2738            & 1.3238           & 1.1029                       & 1.7972                & 51.25                \\
                                                                                        & \text{+ ORNet ($\lambda$=0.3)}             & 35.68                                                    & 0.4310                                         & 0.3710            & 1.9510           & 1.3228                       & 2.1227                & {40.25}              \\
                                                                                        & \text{+ ORNet ($\lambda$=0.5)}             & 37.53                                                    & 0.4517                                         & 0.3908            & 2.1195           & 1.3835                       & 2.1827                & 35.46                \\
                                                                                        & \text{+ ORNet ($\lambda$=0.7)}             & 37.99                                                    & 0.4615                                         & 0.3968            & 2.1711           & 1.3975                       & 2.2157                & 34.39                \\
                                                                                        & \text{+ ORNet ($\lambda$=0.9)}             & 38.05                                                    & 0.4661                                         & 0.3989            & 2.1867           & 1.4262                       & 2.2176                & 33.99                \\
            \cmidrule{2-9}
                                                                                        & \text{NAFNet~\citep{chen2022simple}}       & 22.73                                                    & 0.3937                                         & 0.2458            & 1.2189           & 1.0826                       & 1.7060                & 51.86                \\
                                                                                        & \text{+ ORNet ($\lambda$=0.1)}             & 26.39                                                    & 0.3917                                         & 0.2776            & 1.3228           & 1.1224                       & 1.8217                & 50.96                \\
                                                                                        & \text{+ ORNet ($\lambda$=0.3)}             & 35.87                                                    & 0.4380                                         & 0.3776            & 1.9591           & 1.3513                       & 2.1584                & {40.98}              \\
                                                                                        & \text{+ ORNet ($\lambda$=0.5)}             & 37.89                                                    & 0.4605                                         & 0.3977            & 2.1394           & 1.4030                       & 2.2269                & 35.39                \\
                                                                                        & \text{+ ORNet ($\lambda$=0.7)}             & 38.40                                                    & 0.4709                                         & 0.4039            & 2.1934           & 1.4321                       & 2.2498                & 34.14                \\
                                                                                        & \text{+ ORNet ($\lambda$=0.9)}             & 38.46                                                    & 0.4758                                         & 0.4057            & 2.2088           & 1.4492                       & 2.2501                & 33.77                \\
            \bottomrule
        \end{tabular}
    }
    \label{tab:supp_exp}
    \vspace{-0mm}
\end{table*}

\begin{table*}[!t]
    \centering
    \caption{
        Evaluation on an OOD environment, where an additional Gaussian blur ($\sigma$ = 2.5) is applied to the blurry GoPro test set images.
    }
    \renewcommand{\arraystretch}{1.1}
    \setlength{\tabcolsep}{7pt}
    \vspace{-1mm}
    \scalebox{0.8}{
        \begin{tabular}{l | ccc | ccc | cccc}
            \toprule
            \rowcolor{gray!15}
                                          & \multicolumn{3}{c|}{\textit{Original GT}} & \multicolumn{3}{c|}{\textit{Enhanced GT}} & \multicolumn{4}{c}{\textit{Perceptual Quality Metrics}}                                                                                                                                             \\
            \rowcolor{gray!15}
            {Method}                      & {PSNR}$\uparrow$                          & {SSIM}$\uparrow$                          & {LPIPS}$\downarrow$                                     & {PSNR}$\uparrow$ & {SSIM}$\uparrow$ & {LPIPS}$\downarrow$ & {MUSIQ}$\uparrow$ & {MANIQA}$\uparrow$ & {TOPIQ}$\uparrow$ & {LIQE}$\uparrow$ \\
            \midrule
            \text{FFTFormer}              & 24.5688                                   & 0.7532                                    & 0.4714                                                  & 23.6891          & 0.7224           & 0.5441              & 22.3812           & 0.2284            & 0.1832            & 1.0108           \\
            \text{+ORNet ($\lambda$=0.1)} & {24.5898}                                 & {0.7550}                                  & {0.4592}                                                & {23.7141}        & {0.7245}         & {0.5310}            & {23.1592}         & {0.2481}           & {0.1843}          & {1.0111}         \\
            \text{+ORNet ($\lambda$=0.3)} & {24.5774}                                 & {0.7670}                                  & {0.3429}                                                & {23.8124}        & {0.7405}         & {0.3777}            & {42.9131}         & {0.2638}           & {0.2646}          & {1.0656}         \\
            \text{+ORNet ($\lambda$=0.5)} & {24.5789}                                 & {0.7742}                                  & {0.3107}                                                & {23.8835}        & {0.7500}         & {0.3334}            & {49.6099}         & {0.2689}           & {0.3324}          & {1.2576}         \\
            \text{+ORNet ($\lambda$=0.7)} & {24.4442}                                 & {0.7759}                                  & {0.3042}                                                & {23.8007}        & {0.7524}         & {0.3253}            & {50.9512}         & {0.2950}           & {0.3487}          & {1.3977}         \\
            \text{+ORNet ($\lambda$=0.9)} & {24.2407}                                 & {0.7754}                                  & {0.3036}                                                & {23.6454}        & {0.7524}         & {0.3243}            & {51.5943}         & {0.3174}           & {0.3174}          & {1.0656}         \\
            \bottomrule
        \end{tabular}
    }
    \label{tab:gopro_blur_full}
    \vspace{--2mm}
\end{table*}
\begin{table*}[!t]
    \centering
    \caption{
        Evaluation on an OOD environment, where an additional white noise ($\sigma$ = 9) is applied to the blurry GoPro test set images.
    }
    \renewcommand{\arraystretch}{1.1}
    \setlength{\tabcolsep}{7pt}
    \vspace{-1mm}
    \scalebox{0.8}{
        \begin{tabular}{l | ccc | ccc | cccc}
            \toprule
            \rowcolor{gray!15}
                                          & \multicolumn{3}{c|}{\textit{Original GT}} & \multicolumn{3}{c|}{\textit{Enhanced GT}} & \multicolumn{4}{c}{\textit{Perceptual Quality Metrics}}                                                                                                                                             \\
            \rowcolor{gray!15}
            {Method}                      & {PSNR}$\uparrow$                          & {SSIM}$\uparrow$                          & {LPIPS}$\downarrow$                                     & {PSNR}$\uparrow$ & {SSIM}$\uparrow$ & {LPIPS}$\downarrow$ & {MUSIQ}$\uparrow$ & {MANIQA}$\uparrow$ & {TOPIQ}$\uparrow$ & {LIQE}$\uparrow$ \\
            \midrule
            \text{FFTFormer}              & 24.3574                                   & 0.5867                                    & 0.4463                                                  & {23.8352}        & {0.5713}         & {0.4751}            & 30.1430           & 0.4517             & 0.2655            & 1.1819           \\
            \text{+ORNet ($\lambda$=0.1)} & {24.3804}                                 & {0.5886}                                  & {0.4438}                                                & {23.8616}        & {0.5734}         & {0.4720}            & {30.5873}         & {0.4528}           & {0.2664}          & {1.1909}         \\
            \text{+ORNet ($\lambda$=0.3)} & {24.4088}                                 & {0.6179}                                  & {0.4070}                                                & {23.9693}        & {0.6057}         & {0.4233}            & {41.8760}         & {0.4699}           & {0.3188}          & {1.4321}         \\
            \text{+ORNet ($\lambda$=0.5)} & {24.4819}                                 & {0.6771}                                  & {0.3549}                                                & {24.1483}        & {0.6671}         & {0.3602}            & {55.2554}         & {0.5204}           & {0.4034}          & {1.8986}         \\
            \text{+ORNet ($\lambda$=0.7)} & {24.3252}                                 & {0.6943}                                  & {0.3372}                                                & {24.0463}        & {0.6852}         & {0.3397}            & {58.9384}         & {0.5473}           & {0.4212}          & {2.1779}         \\
            \text{+ORNet ($\lambda$=0.9)} & {24.0224}                                 & {0.7018}                                  & {0.3313}                                                & {23.7722}        & {0.6928}         & {0.3335}            & {60.1486}         & {0.5618}           & {0.4295}          & {2.2923}         \\
            \bottomrule
        \end{tabular}
    }
    \label{tab:gopro_noise_full}
    \vspace{--2mm}
\end{table*}

\paragraph{Quantitative result}
    Tables~\ref{tab:supp_exp}, \ref{tab:gopro_blur_full}, and \ref{tab:gopro_noise_full} present the extended results across different $\lambda$ values.
We consistently observe that increasing $\lambda$ leads to better scores in perceptual quality metrics.
However, an excessively high $\lambda$ exacerbates the risks discussed in the main paper, such as hallucination issues.
Furthermore, in the out-of-distribution (OOD) settings shown in Tables~\ref{tab:gopro_blur_full} and~\ref{tab:gopro_noise_full}, we observe that an excessively high $\lambda$ value can degrade reference-based performance against both the original and enhanced GTs.
This appears to be because when $\lambda$ is too large, the ORNet applies changes in color tone and further enhancements that go beyond removing the remaining degradations (blur, noise), resulting in a reference based performance drop. 
This highlights the importance of selecting an appropriate $\lambda$ to achieve a balance between enhancement and fidelity.

\paragraph{Qualitative results}

\begin{figure*}[t]
  \centering
  \scalebox{1.0}{
    \begin{tabular}{@{}c@{\hspace{0.3em}}c@{\hspace{0.3em}}c@{\hspace{0.3em}}c@{\hspace{0.3em}}c@{}}

      \parbox[t]{0.24\textwidth}{
        \centering
        \includegraphics[width=\linewidth]{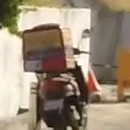}
        \vspace{-7mm}
        \vspace{0mm}    
      } &
    \parbox[t]{0.24\textwidth}{
        \centering
        \includegraphics[width=\linewidth]{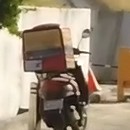}
        \vspace{-7mm}
        \vspace{0mm}    
      } &
    \parbox[t]{0.24\textwidth}{
        \centering
        \includegraphics[width=\linewidth]{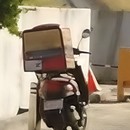}
        \vspace{-7mm}
      } &
    \parbox[t]{0.24\textwidth}{
        \centering
        \includegraphics[width=\linewidth]{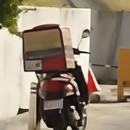}
        \vspace{-7mm}
      }  \\
      \vspace{2mm}

      \parbox[t]{0.24\textwidth}{
        \centering
        \includegraphics[width=\linewidth]{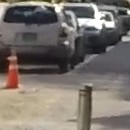}
        \vspace{-7mm}
        \caption*{ \fontsize{8pt}{8pt}\selectfont Original GT}
        \vspace{0mm}    
      } &
    \parbox[t]{0.24\textwidth}{
        \centering
        \includegraphics[width=\linewidth]{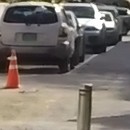}
        \vspace{-7mm}
        \caption*{ \fontsize{8pt}{8pt}\selectfont $\lambda$ = 0.3}
        \vspace{0mm}    
      } &
    \parbox[t]{0.24\textwidth}{
        \centering
        \includegraphics[width=\linewidth]{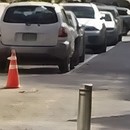}
        \vspace{-7mm}
        \caption*{ \fontsize{8pt}{8pt}\selectfont  $\lambda$ = 0.5}
      } &
    \parbox[t]{0.24\textwidth}{
        \centering
        \includegraphics[width=\linewidth]{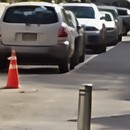}
        \vspace{-7mm}
        \caption*{ \fontsize{8pt}{8pt}\selectfont $\lambda$ = 1.0}
      }  \\
    \vspace{2mm}
        \vspace{-5mm}
    \end{tabular}
  }
  \vspace{-5mm}
  \centering
  \caption{
  Qualitative comparison of GT enhancement with varying $\lambda$ values. 
  Excessively large $\lambda$ values increase the risk of hallucinations, such as color shifts and semantic deviations from the original GT.
  Zoom in for better visualization.
  }
  \vspace{-2mm}
  \label{fig:enhanced_gt_cond}
\end{figure*}

Figure~\ref{fig:enhanced_gt_cond} visualizes the results of our Ground Truth (GT) enhancement with varying values of the hyperparameter $\lambda$. 
As $\lambda$ increases, the perceptual quality may be enhanced, but this can introduce undesirable artifacts such as altered color tones and semantic changes that deviate from the original GT. 
In contrast, an optimally chosen $\lambda$ effectively removes residual noise and blur, leading to a perceptually improved image while preserving the color and semantic integrity of the original.

\paragraph{Mask visualization on diverse enhancement level $\lambda$}

\begin{figure*}[h]
    \centering
    \hspace{-14mm}
        \begin{subfigure}[b]{0.13\textwidth}\centering
            $M_0$
            \vspace{11mm}
        \end{subfigure}\hspace{-9mm}
        \begin{subfigure}[b]{0.13\textwidth}\centering
            \includegraphics[width=\linewidth]{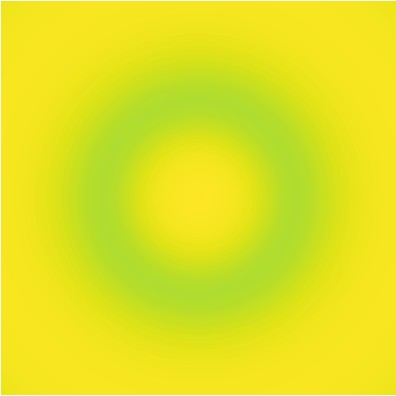}
        \end{subfigure}
        \begin{subfigure}[b]{0.13\textwidth}\centering
            \includegraphics[width=\linewidth]{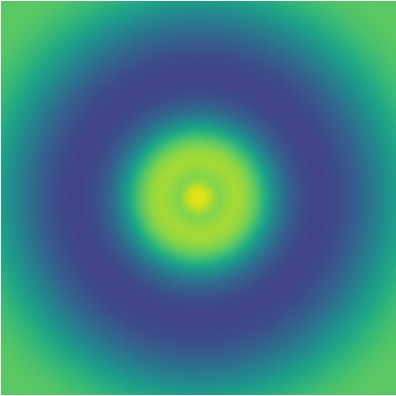}
        \end{subfigure}
        \begin{subfigure}[b]{0.13\textwidth}\centering
            \includegraphics[width=\linewidth]{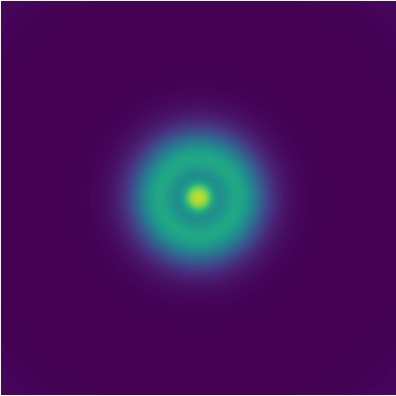}
        \end{subfigure}
        \begin{subfigure}[b]{0.13\textwidth}\centering
            \includegraphics[width=\linewidth]{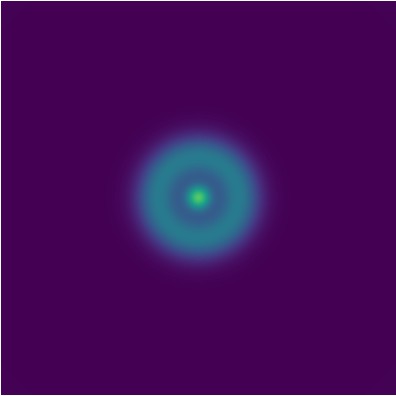}
        \end{subfigure}
        \begin{subfigure}[b]{0.13\textwidth}\centering
            \includegraphics[width=\linewidth]{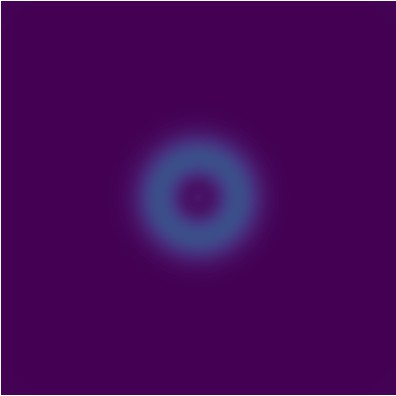}
        \end{subfigure}\hfill
        \vfill
        \hspace{-14mm}
        \begin{subfigure}[b]{0.13\textwidth}\centering
            $M_1$
            \vspace{11mm}
        \end{subfigure}\hspace{-9mm}
        \begin{subfigure}[b]{0.13\textwidth}\centering
            \includegraphics[width=\linewidth]{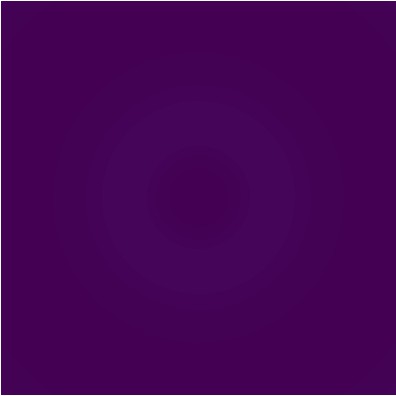}
        \end{subfigure}
        \begin{subfigure}[b]{0.13\textwidth}\centering
            \includegraphics[width=\linewidth]{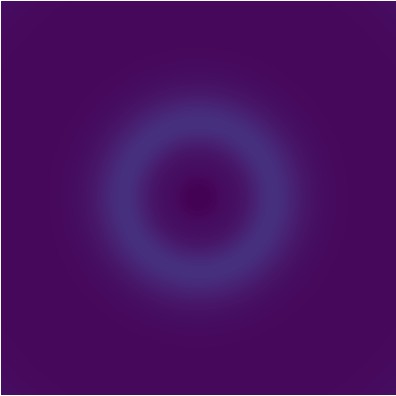}
        \end{subfigure}
        \begin{subfigure}[b]{0.13\textwidth}\centering
            \includegraphics[width=\linewidth]{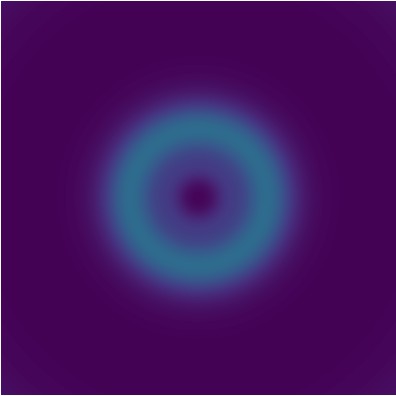}
        \end{subfigure}
        \begin{subfigure}[b]{0.13\textwidth}\centering
            \includegraphics[width=\linewidth]{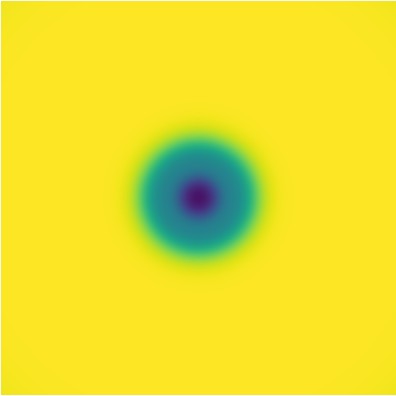}
        \end{subfigure}
        \begin{subfigure}[b]{0.13\textwidth}\centering
            \includegraphics[width=\linewidth]{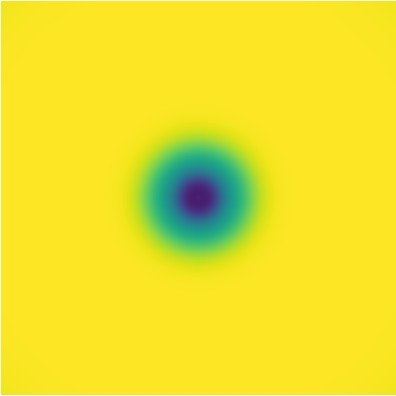}
        \end{subfigure}\hfill
        \vfill
        \hspace{-14mm}
        \begin{subfigure}[b]{0.13\textwidth}\centering
            $M_2$
            \vspace{11mm}
        \end{subfigure}\hspace{-9mm}
        \begin{subfigure}[b]{0.13\textwidth}\centering
            \includegraphics[width=\linewidth]{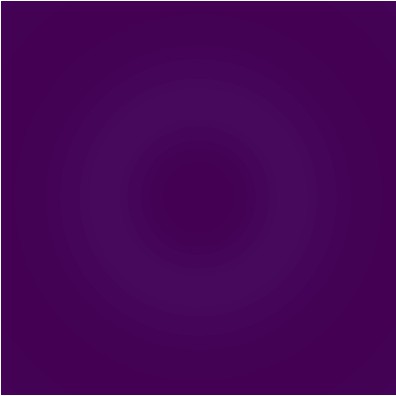}
        \end{subfigure}
        \begin{subfigure}[b]{0.13\textwidth}\centering
            \includegraphics[width=\linewidth]{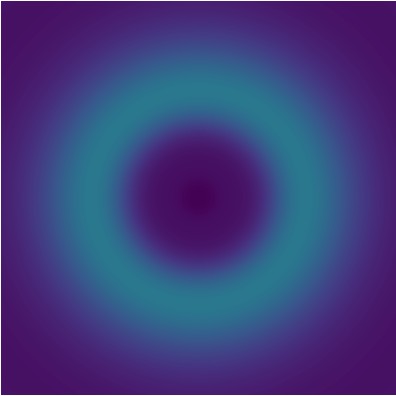}
        \end{subfigure}
        \begin{subfigure}[b]{0.13\textwidth}\centering
            \includegraphics[width=\linewidth]{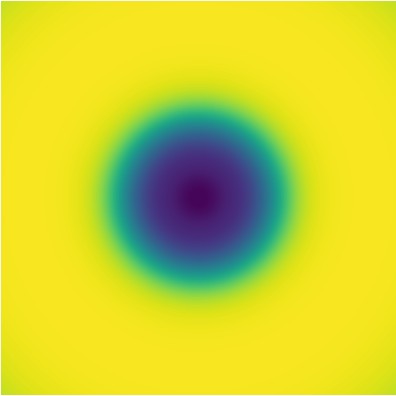}
        \end{subfigure}
        \begin{subfigure}[b]{0.13\textwidth}\centering
            \includegraphics[width=\linewidth]{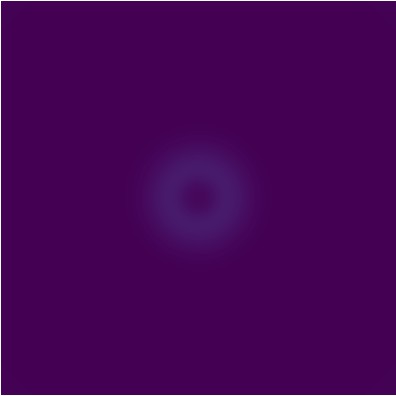}
        \end{subfigure}
        \begin{subfigure}[b]{0.13\textwidth}\centering
            \includegraphics[width=\linewidth]{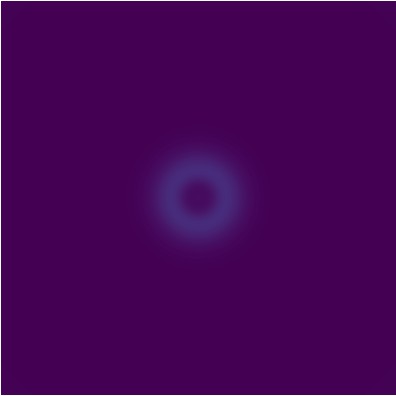}
        \end{subfigure}\hfill
        \vfill
        \hspace{-14mm}
        \begin{subfigure}[b]{0.13\textwidth}\centering
            $M_3$
            \vspace{11mm}

        \end{subfigure}\hspace{-9mm}
        \begin{subfigure}[b]{0.13\textwidth}\centering
            \includegraphics[width=\linewidth]{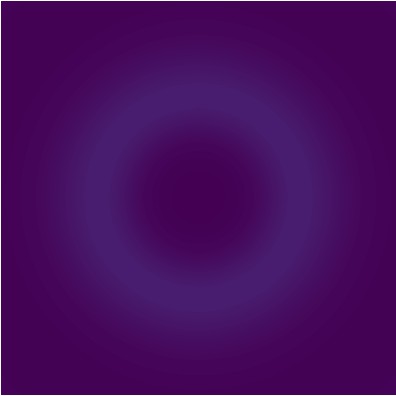}
        \end{subfigure}
        \begin{subfigure}[b]{0.13\textwidth}\centering
            \includegraphics[width=\linewidth]{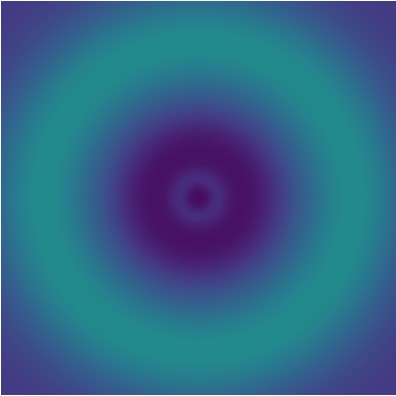}
        \end{subfigure}
        \begin{subfigure}[b]{0.13\textwidth}\centering
            \includegraphics[width=\linewidth]{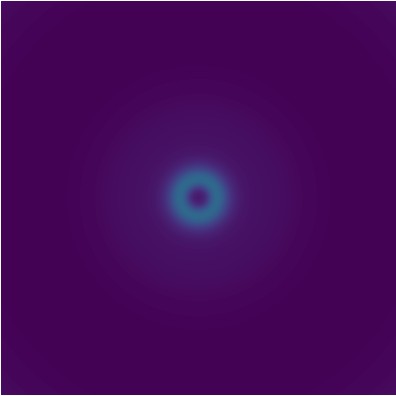}
        \end{subfigure}
        \begin{subfigure}[b]{0.13\textwidth}\centering
            \includegraphics[width=\linewidth]{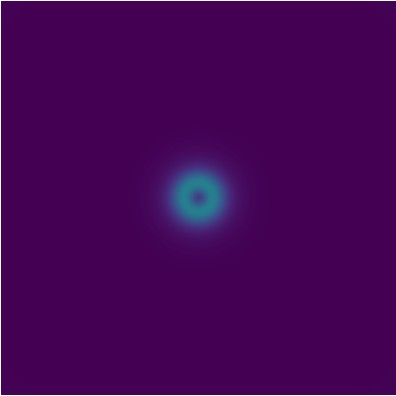}
        \end{subfigure}
        \begin{subfigure}[b]{0.13\textwidth}\centering
            \includegraphics[width=\linewidth]{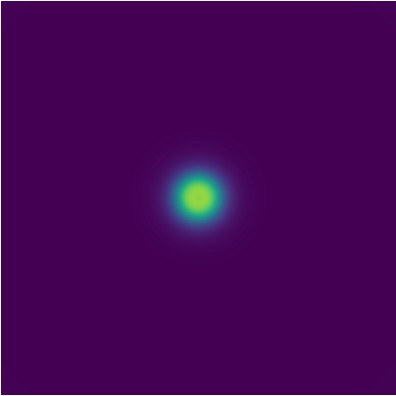}
        \end{subfigure}\hfill
        \vfill

        \hspace{-14mm}
        \begin{subfigure}[b]{0.13\textwidth}\centering
           
            \hspace{10mm}
            
        \end{subfigure}\hspace{-9mm}
        \begin{subfigure}[b]{0.13\textwidth}\centering
            $\lambda=0.1$
        \end{subfigure}
        \begin{subfigure}[b]{0.13\textwidth}\centering
           $\lambda=0.3$
        \end{subfigure}
        \begin{subfigure}[b]{0.13\textwidth}\centering
            $\lambda=0.5$
        \end{subfigure}
        \begin{subfigure}[b]{0.13\textwidth}\centering
            $\lambda=0.7$
        \end{subfigure}
        \begin{subfigure}[b]{0.13\textwidth}\centering
           $\lambda=0.9$
        \end{subfigure}\hfill
        \vfill

    \caption{
        Frequency masks generated by the proposed controllable frequency mask generator. The enhancement level is controlled by the input parameter $\lambda$.
    }
    \label{fig:mask_analysis_full}
\end{figure*}

Figure~\ref{fig:mask_analysis_full} shows the visualization of the generated masks with different $\lambda$ values.
When $\lambda$ is small, the generated masks are mostly focused on $M_0$, dedicated for original ground truth image.
As $\lambda$ increases, generated masks cover a diverse other ground truth variants.

\paragraph{Additional user study}
\begin{figure*}[t]
    \centering
    \includegraphics[width=0.5\linewidth]{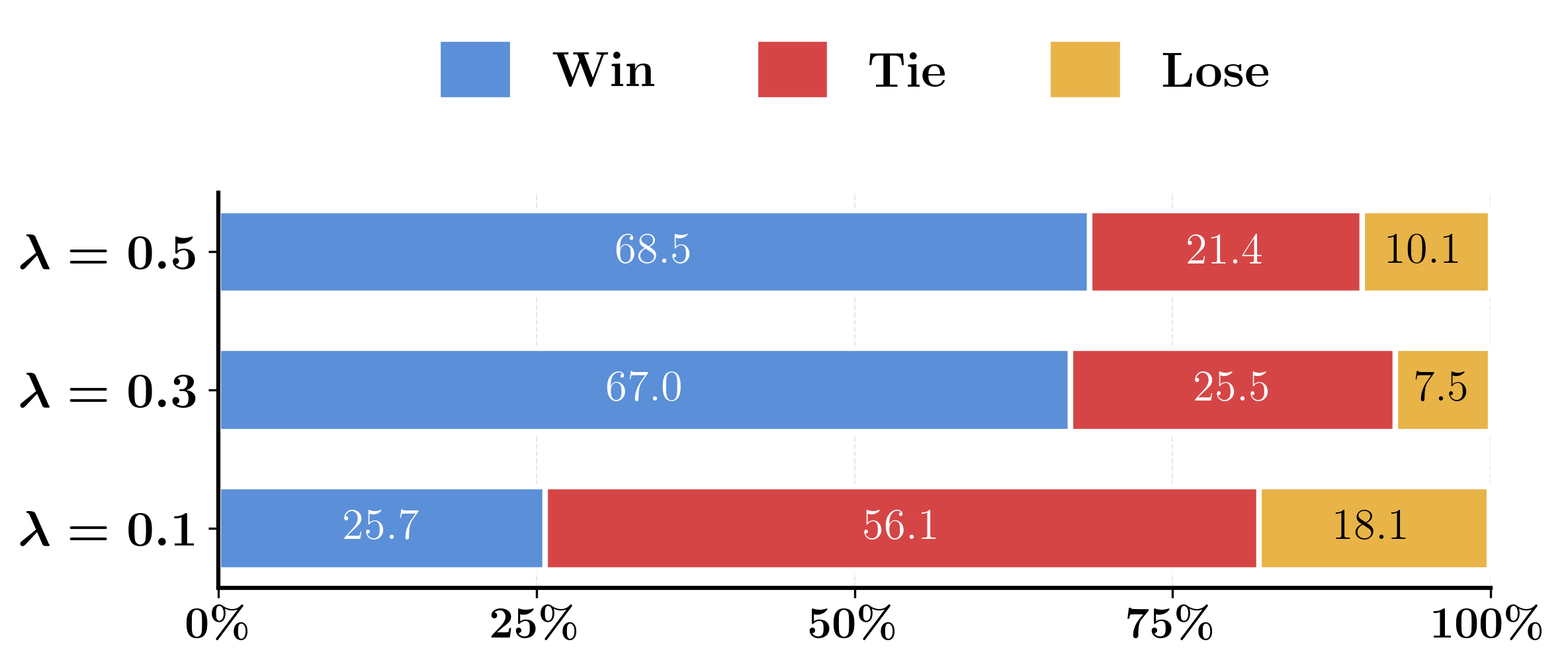}
    \vspace{-2mm}
    \caption{
        User study with various $\lambda$ values. Participants consistently preferred our ORNet outputs to the baselines.
    }
    \label{fig:user_study_full}
\end{figure*}
We conducted a user study to validate the perceptual quality of the output images of ORNet with different $\lambda$ values.
Following the same protocol as described in the main paper, we evaluated three levels of the refinement weight $\lambda \in \{0.1, 0.3, 0.5\}$.
As shown in Figure~\ref{fig:user_study_full}, the results indicate that while both $\lambda=0.3$ and $\lambda=0.5$ achieved similarly high preference rates, $\lambda=0.3$ yielded the lowest loss rate.

\subsection{Training Stability}
To assess the training stability of our modular output refinement network (ORNet), we trained the ORNet with five independent times using distinct random seeds.
Following training, each refinement network was applied to the outputs of a pretrained AdaRevD~\citep{mao2024adarevd} on the GoPro test dataset.
Then, the standard deviation is calculated for each metric with $\lambda=0.3$.
The resulting standard deviations were as follows: PSNR (0.023), SSIM (0.0002), LPIPS (0.0005), DISTS (0.0005), MUSIQ (0.097), MANIQA (0.0007), TOPIQ (0.001), and LIQE (0.01).
The observed standard deviations for each metric are notably low.
This outcome indicates a high degree of stability in our training procedure for the output refinement network.


\section{Limitations}
Although our enhanced supervision framework achieves notable perceptual gains, selecting the enhancement strength $\lambda$ remains inherently challenging. Existing no-reference IQA metrics capture perceptual naturalness but provide limited insight into semantic fidelity, making it difficult to determine $\lambda$ in a fully principled manner.

To partially mitigate this issue, we evaluate fidelity at $\lambda$ = 0.3 using the available surrogate metrics—CLIP embedding similarity together with object-level consistency assessments based on detection alignment with the original ground truth (see Figure~\ref{fig:mixed_sample} and Table~\ref{tab:ocr_face} of the main paper).
These analyses indicate that $\lambda$ = 0.3 preserves semantic structures while providing noticeable perceptual enhancement. Nevertheless, this evaluation remains constrained by the lack of a unified metric that jointly reflects fidelity and naturalness.
Developing such a metric, or an adaptive mechanism that balances the two aspects automatically, is an important direction for future work.


\end{document}